\documentclass[runningheads]{llncs}

\usepackage{eccv}

\usepackage{eccvabbrv}

\usepackage{graphicx}
\usepackage{booktabs}

\usepackage[accsupp]{axessibility}  %

\usepackage[pagebackref,breaklinks,colorlinks,citecolor=eccvblue]{hyperref}

\usepackage{orcidlink}

\usepackage{float}
\usepackage[noend,ruled,linesnumbered]{algorithm2e}
\SetKwInput{KwInput}{Input}
\SetKwInput{KwOutput}{Output}
\usepackage{amsfonts}
\usepackage{amsmath}
\usepackage[capitalize]{cleveref}
\crefname{section}{Sec.}{Secs.}
\Crefname{section}{Section}{Sections}
\Crefname{table}{Table}{Tables}
\crefname{table}{Tab.}{Tabs.}
\usepackage{cuted}
\usepackage{booktabs}
\usepackage{tabularx}
\usepackage{multicol}
\setkeys{Gin}{keepaspectratio}
\usepackage{xcolor}
\usepackage{pgfplots}
\usepgfplotslibrary{statistics}
\usepackage{pgf,tikz,pgfplotstable}
\pgfplotsset{compat=newest}
\usetikzlibrary{shadows,backgrounds,shapes, shapes.geometric, arrows.meta,decorations.text,decorations.pathreplacing,arrows,patterns
}
\usepackage{adjustbox}
\usepackage[percentage]{overpic}
\usepackage{calc}
\usepackage{mathtools}
\usepackage{standalone}
\usepackage[shortlabels]{enumitem}
\usepackage{chngcntr}

\newcommand{\rebuttalfix}[1]{\textcolor{black}{#1}}

\newcommand{\smcomb}[0]{\texttt{Sm-Comb}}
\newcommand{\fastdog}[0]{\texttt{FastDOG}}
\newcommand{\lprelax}[0]{\texttt{LP-Relax}}
\newcommand{\caoetal}[0]{\texttt{URSSM}}
\newcommand{\kernelm}[0]{\texttt{KrnlM}}
\newcommand{\ours}[0]{\texttt{DiscoMatch}}
\newcommand{\rotatedCentering}[3]{\rotatebox{#1}{\hspace{(#2-\widthof{#3})/2}#3}}

\newcommand{\scircled}[1]{{\scriptsize\textcircled{\tiny #1}}}
\newcommand{\var}{\text{var}}
\newcommand{\idx}{\text{idx}}

\definecolor{cPLOT0}{RGB}{214,113,176}
\definecolor{cPLOT1}{RGB}{80,150,80}
\definecolor{cPLOT2}{RGB}{98,148,96}
\definecolor{cPLOT3}{RGB}{69,214,214}
\definecolor{cPLOT4}{RGB}{141, 90, 203}
\definecolor{cPLOT5}{RGB}{255,200,87}
\definecolor{cPLOT6}{RGB}{108, 117, 125}
\definecolor{cBLUE}{RGB}{116, 74, 222}
\definecolor{cPINK}{RGB}{214,113,176}
\definecolor{cGREEN}{RGB}{80,150,80}
\definecolor{cRED}{RGB}{201, 5, 10}
\definecolor{cPINK2}{RGB}{231,71,111}

\def\myparagraph#1{\vspace*{3.5pt}\noindent{\bf #1~~}}
\usepackage{enumitem} %
\newenvironment{packed_description}{
\begin{description}[leftmargin=12pt]
  \setlength{\itemsep}{3pt}
  \setlength{\parskip}{1pt}
  \setlength{\parsep}{1pt}
}{\end{description}}

\def\yellowcirclemarker{
\tikz[baseline=-0.9ex]
\node[color=cPLOT5, scale=1.5, line width=1pt]{\pgfuseplotmark{o}};
\xspace
}

\begin{document}

\title{DiscoMatch: Fast Discrete Optimisation for Geometrically Consistent 3D Shape Matching}

\titlerunning{Fast Discrete Optimisation for Geometrically Consistent 3D Shape Matching}

\author{
Paul~Roetzer$^*$\inst{1}
\and
Ahmed~Abbas$^*$\inst{2}%
\and
Dongliang~Cao\inst{1}
\and
Florian~Bernard\inst{1}
\and
Paul~Swoboda\inst{3}
}

\authorrunning{P.~Roetzer \& A.~Abbas et al.}

\institute{University of Bonn \and
MPI for Informatics, Saarland Informatics Campus \and
Heinrich-Heine University D\"usseldorf
}

\maketitle

\newcommand{\teaserheight}[0]{3.5cm}%
\def\interheight{2.4cm}%
\def\interheightsmall{1.9cm}%
\def\redarrow{\rotatedCentering{90}{\interheight}{\textbf{{$\downarrow$}}}}
\def\greenarrow{\rotatedCentering{90}{\interheight}{{$\downarrow$}}}
\def\spacerBetweenFigures{$\qquad\quad$}
\vspace{-0.8cm}
\begin{figure}[H]
   \centerline{%
   \footnotesize%
   \setlength{\tabcolsep}{1pt}
   \begin{tabular}{c}%
     \begin{tabular}{cccccccccccc}
         \rotatedCentering{90}{\interheight}{\caoetal~\cite{cao2023unsupervised}}&
         \hspace{0.1cm}
         \includegraphics[height=\interheight]{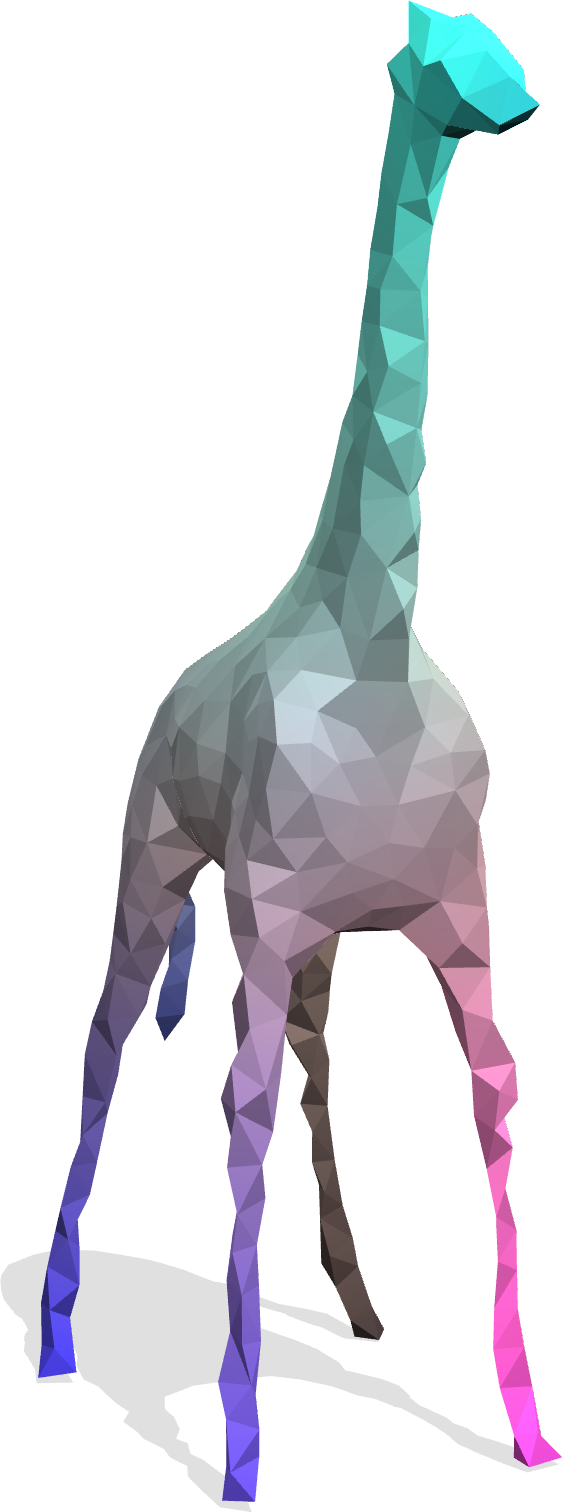}&
         \redarrow&
         \includegraphics[height=\interheight]{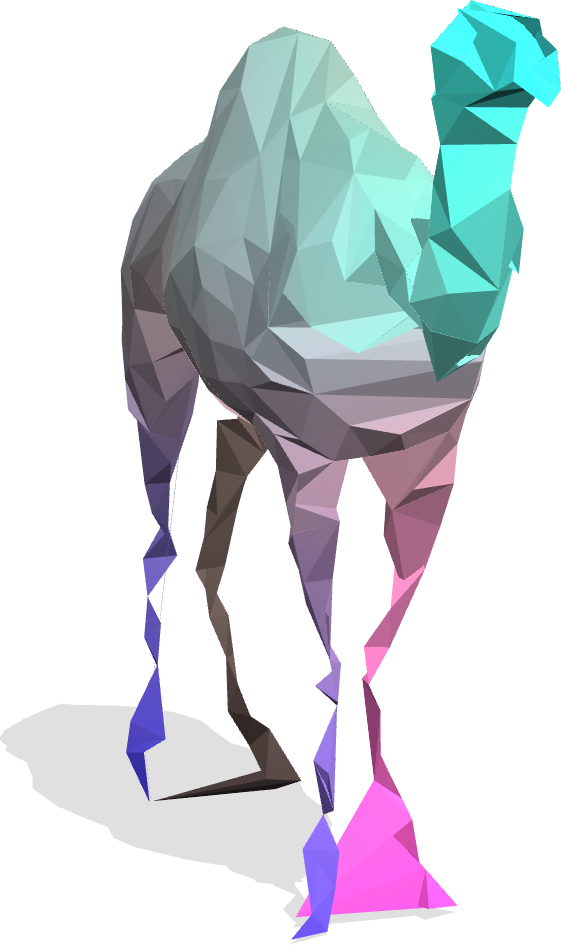}
         &\spacerBetweenFigures
         &
         \includegraphics[height=\interheight]{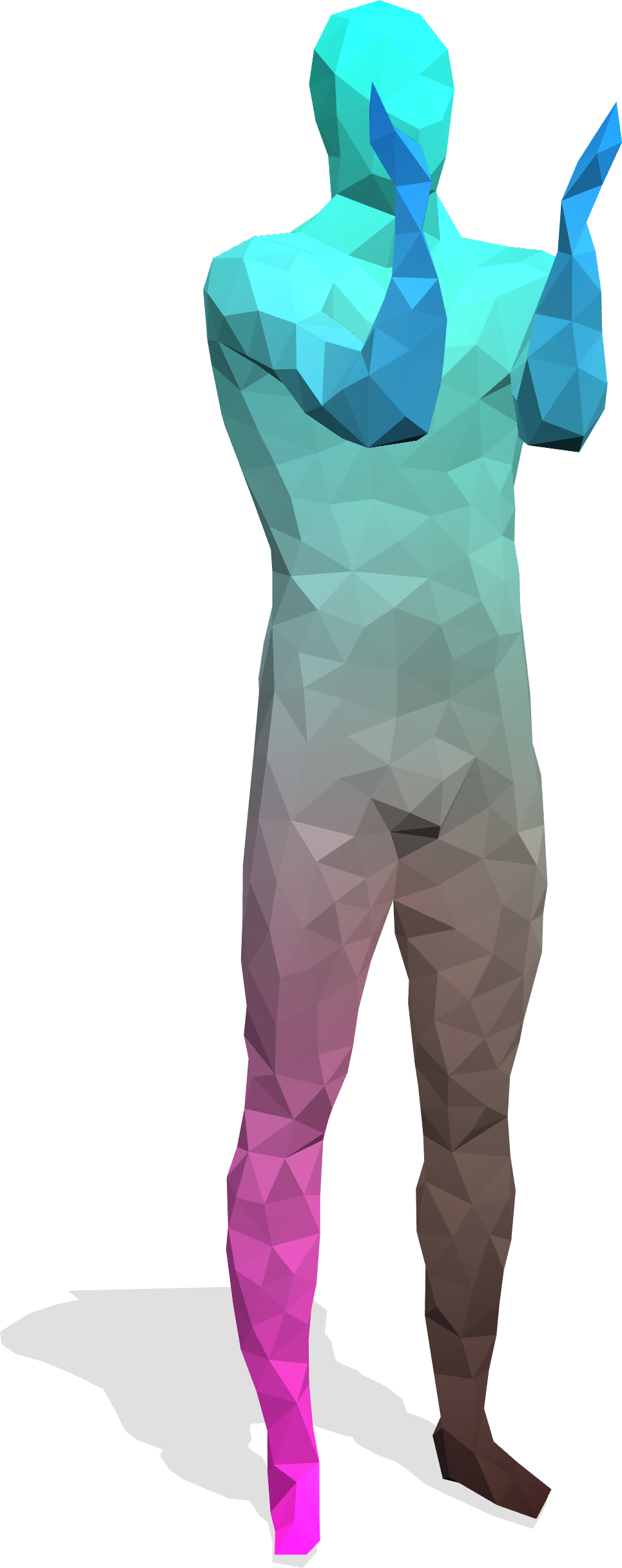}&
         \redarrow&
         \includegraphics[height=\interheight]{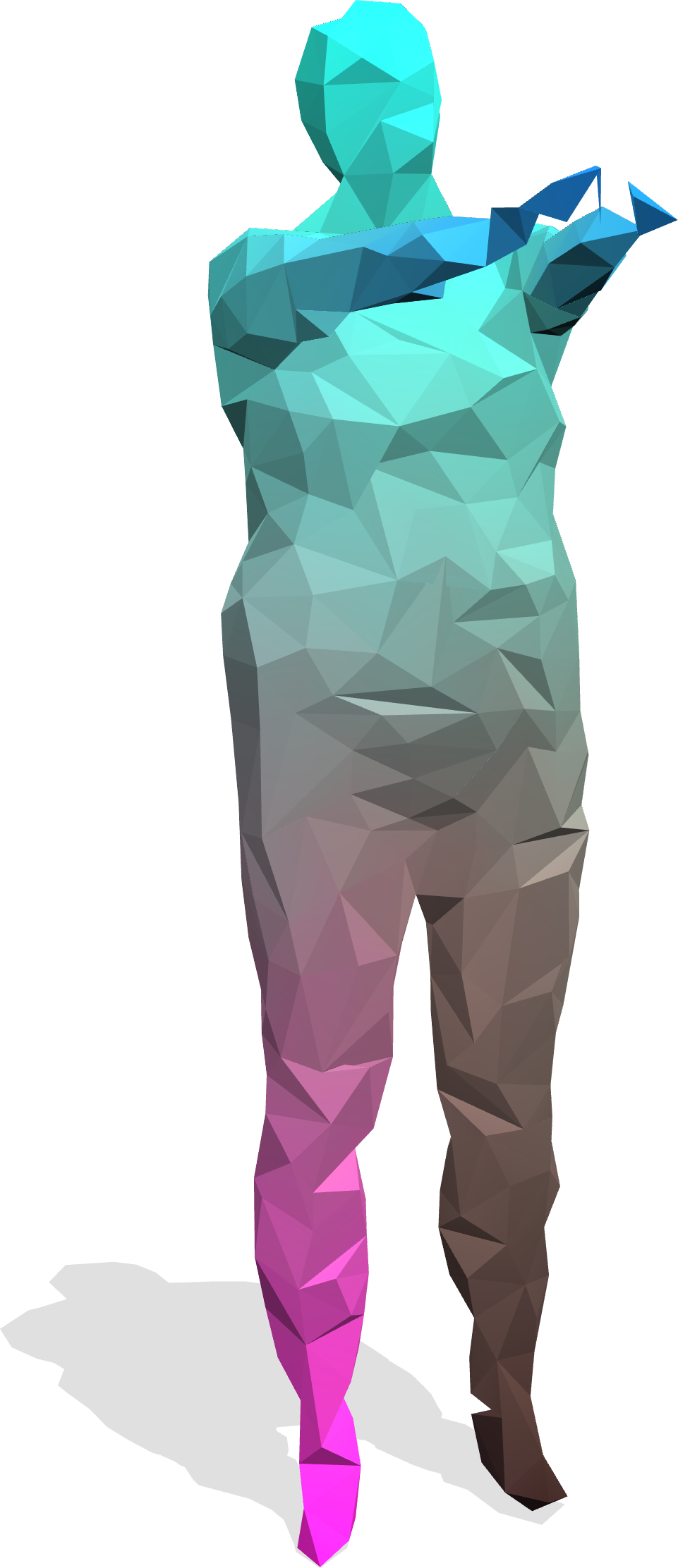}
         &\spacerBetweenFigures
         &
         \includegraphics[height=\interheightsmall]{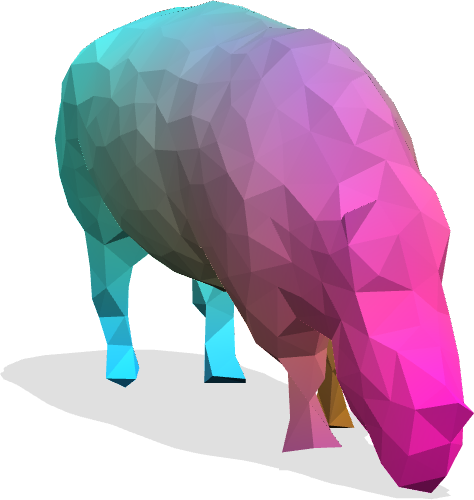}&
         \redarrow&
         \includegraphics[height=\interheightsmall]{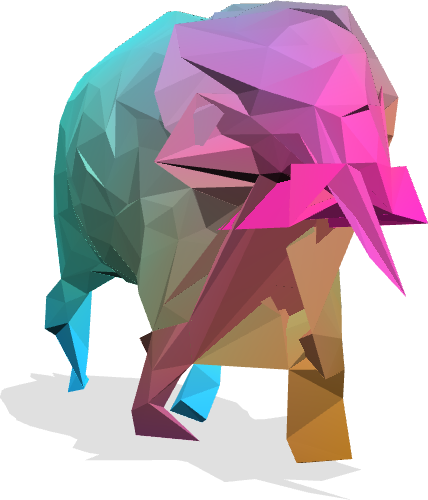}\\
         \rotatedCentering{90}{\interheight}{\ours}&
         \hspace{0.1cm}
         \includegraphics[height=\interheight]{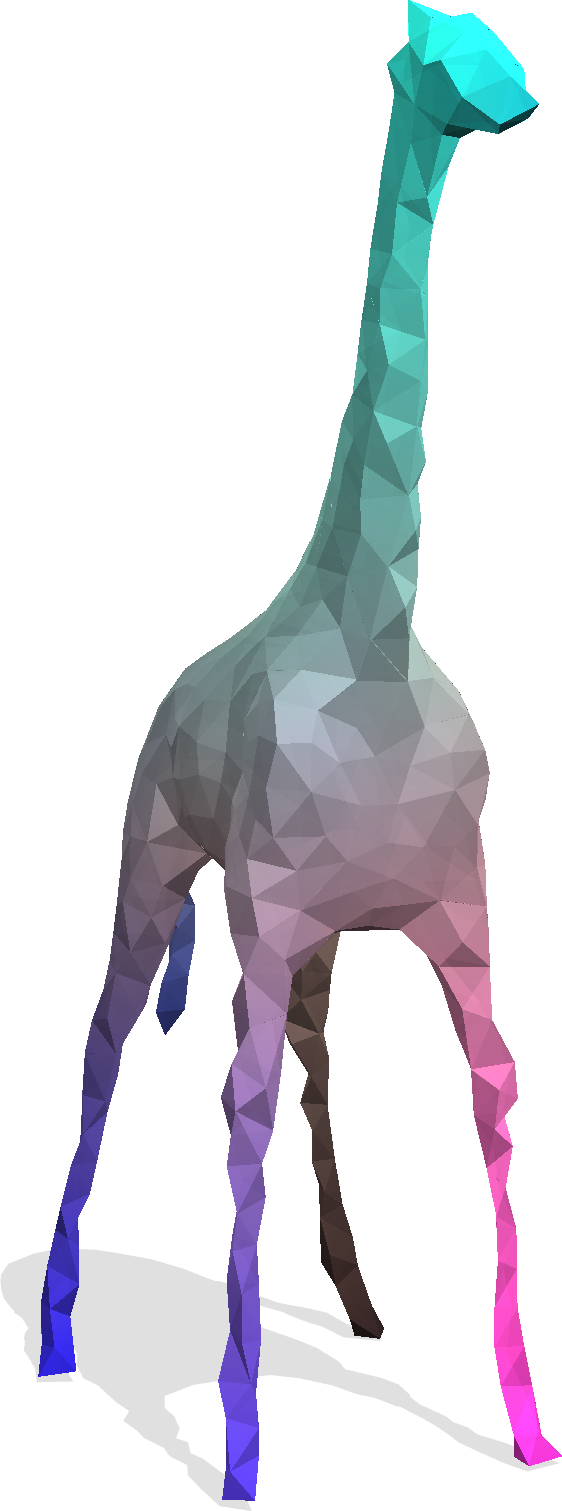}&
         \greenarrow&
         \includegraphics[height=\interheight]{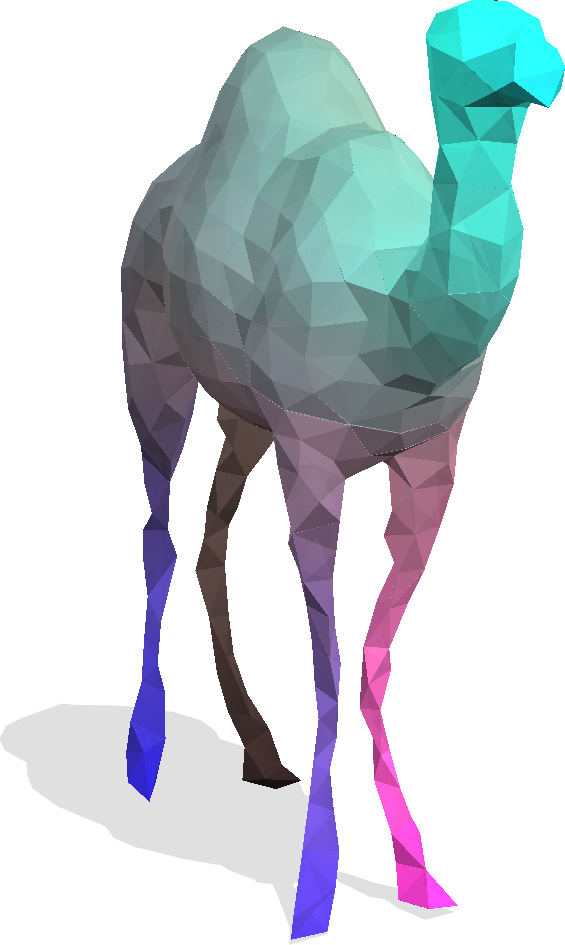}
         &\spacerBetweenFigures
         &
         \includegraphics[height=\interheight]{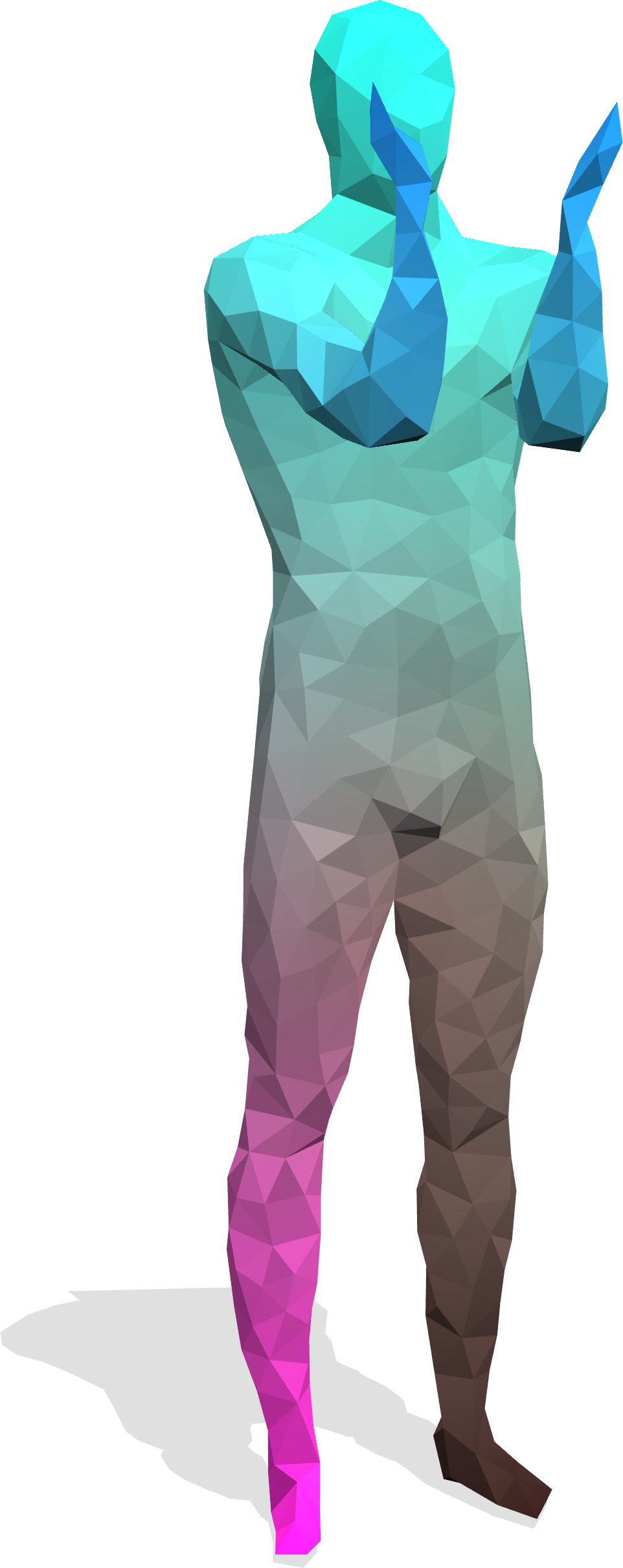}&
         \greenarrow&
         \includegraphics[height=\interheight]{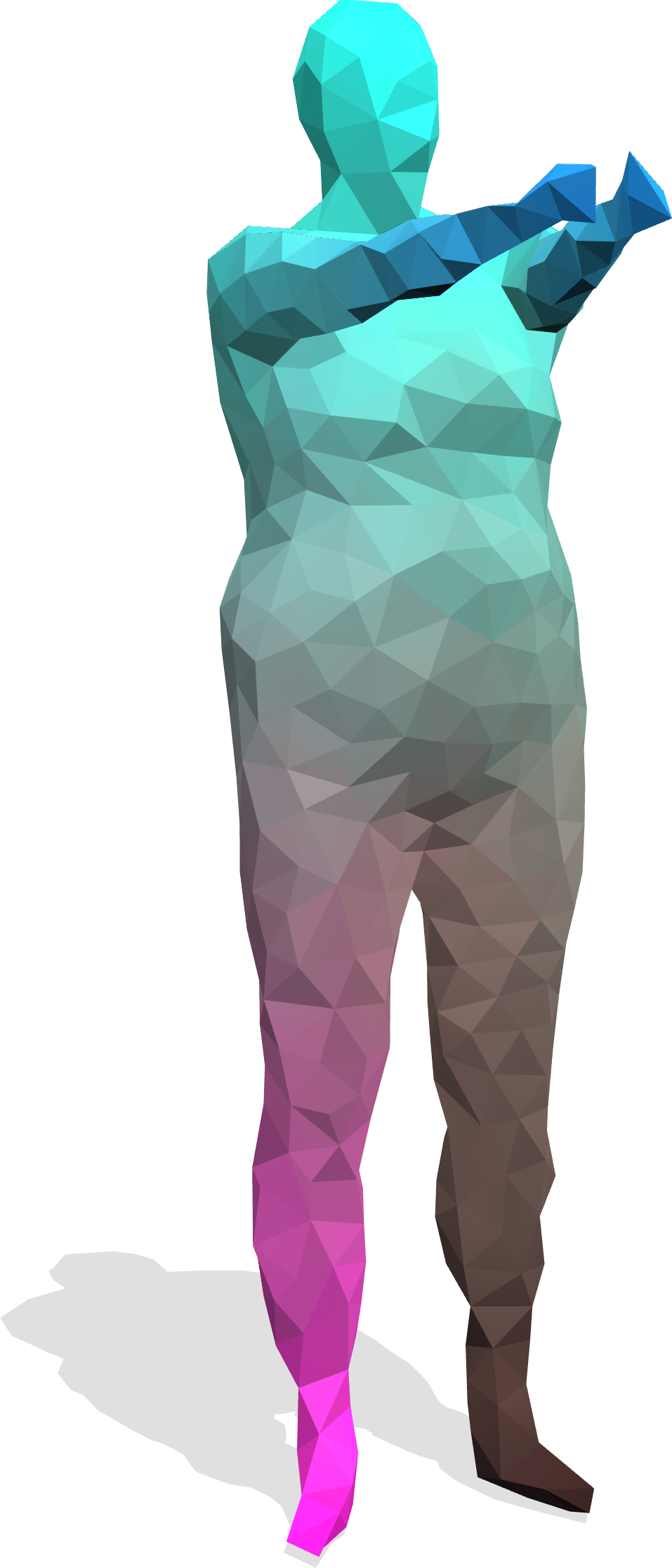}
         &\spacerBetweenFigures
         &
         \includegraphics[height=\interheightsmall]{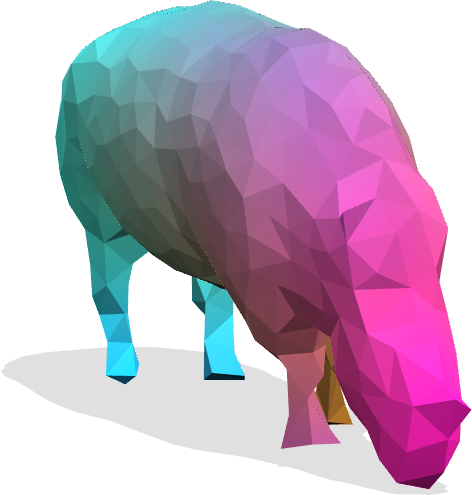}&
         \greenarrow&
         \includegraphics[height=\interheightsmall]{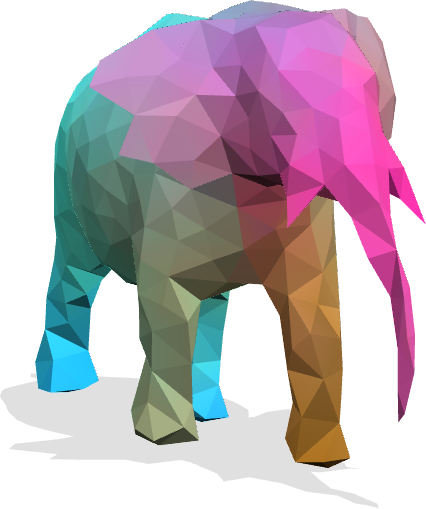}
     \end{tabular}\\
     {\scriptsize Triangulation Transfer via Computed Matching ($\rightarrow$)}
   \end{tabular}
   }
    \captionof{figure}{
    We efficiently solve a large integer linear program (ILP) to {guarantee geometric consistency of matchings between pairs of 3D shapes}.
    Our matchings are smooth and more accurate than the recent state-of-the-art shape matching approach \caoetal~\cite{cao2023unsupervised}. Improved solution quality can be seen when transferring triangulation between shapes. 
    }
\label{fig:teaser}
\end{figure}
\vspace{-0.8cm}

\begin{abstract}
In this work we propose to combine the advantages of learning-based and combinatorial formalisms for 3D shape matching. While learning-based methods lead to state-of-the-art matching performance, they do not ensure geometric consistency, so that obtained matchings are locally non-smooth. On the contrary, axiomatic, optimisation-based methods allow to take geometric consistency into account by explicitly constraining the space of valid matchings. However, existing axiomatic formalisms do not scale to practically relevant problem sizes, and require user input for the initialisation of non-convex optimisation problems. 
We work towards closing this gap
by proposing a novel combinatorial solver that combines a unique set of favourable properties: our approach (i)~is initialisation free, (ii)~is massively parallelisable and powered by a quasi-Newton method, (iii)~provides optimality gaps, and (iv)~delivers improved matching quality with decreased runtime and globally optimal results for many instances.
\def\thefootnote{*}\footnotetext{These authors contributed equally to this work.}\def\thefootnote{\arabic{footnote}}

\keywords{Shape matching \and Discrete optimisation}

\end{abstract}

\section{Introduction}\label{sec:intro}

The shape matching problem, i.e.,~finding correspondences between two non-rigidly deformed 3D shapes, lies at the heart of many visual computing tasks.
Resulting correspondences are relevant for many downstream applications in computer vision (3D reconstruction, shape space modelling), graphics (texture transfer, shape interpolation) or medical image analysis (shape-based segmentation, navigation).
Thus, shape matching has received a lot of attention in recent years and has been tackled by numerous variants of both axiomatic and learning-based methods.

Learning-based methods usually provide the best matching performance on the most challenging benchmarks~\cite{cao2023unsupervised}. However, a major downside is that they do not provide any guarantees about crucial properties of the final correspondences, such as \emph{geometric consistency} of matchings: for triangular surface meshes, geometric consistency means that neighbouring triangles of one shape are consistently matched to neighbouring triangles of the other shape (analogous to diffeomorphisms in the continuous domain), cf. \cref{fig:geo-cons} left.%

While axiomatic methods can -- in principle -- take geometric consistency into account, most existing shape matching approaches disregard it~\cite{van2011survey,ovsjanikov2012functional,tam2012registration,huang2013consistent,solomon2016entropic,sahilliouglu2018genetic,bernard2020mina}, mainly due to the resulting non-convexity of the optimisation problem or hard-to-solve constraints. Although there are some exceptions, respective formulations can typically be solved only for small instances, such as~\cite{windheuser2011geometrically}, or they rely on manual user initialisations to avoid poor local optima and to better constrain the problem~\cite{schreiner2004inter, takayama2022compatible, schmidt2023surface}.

We argue that geometric consistency is of crucial importance and it should be considered as an essential aspect for any shape matching pipeline.
Hence, in this work we propose to combine expressive and powerful shape features obtained from a learning-based method
with a novel combinatorial solver for the geometrically consistent shape matching formalism~\cite{windheuser2011geometrically}.
We summarise our main contributions as follows:\footnote{Our code is publicly available at \url{https://github.com/paul0noah/disco-match}.} 
\setlist{nolistsep}
\begin{enumerate}[a.,noitemsep]
    \item We propose a theoretically well-grounded solver for global shape matching that is able to quantify optimality gaps. We are empirically able to certify global optimality in $90\%$ of the considered shape matching instances.
    
    \item Our solver extends the dual decomposition approach~\cite{abbas2022fastdog} and also addresses specific challenges posed by large shape matching instances.
    By doing so we demonstrate a $9$ to $11$ times faster convergence compared to~\cite{abbas2022fastdog} on shape matching problems. 
    Additionally some components of our solver are generally applicable and yield $5$ to $10$ times runtime improvements over~\cite{abbas2022fastdog}. 
    
    \item We outperform the current state-of-the-art learning-based shape matching approach on most considered datasets w.r.t. matching accuracy and smoothness, while at the same time guaranteeing geometric consistency.
\end{enumerate}

\section{Related Work}\label{sec:related-work}
In the following we summarise works that are most relevant to our approach. For an extensive review of shape matching methods we refer to the surveys~\cite{van2011survey, sahilliouglu2020recent}.

\myparagraph{Deep Shape Matching.}
With the rapid development of deep learning, many learning-based approaches are proposed for 3D shape matching, which can be categorised into supervised and unsupervised methods. In the context of supervised learning, some methods~\cite{wiersma2020cnns,li2020shape} formulate shape matching as a classification problem, while others~\cite{groueix20183d,trappolini2021shape} solve the problem based on non-rigid shape alignment. Nevertheless, supervised methods require ground-truth correspondences, which in practice are hard to obtain. Therefore, many unsupervised methods were introduced to get rid of the requirement of ground-truth correspondences. Notably, the functional map framework~\cite{ovsjanikov2012functional} is one of the most dominant ingredients to enable unsupervised learning. %

Earlier works~\cite{halimi2019unsupervised,roufosse2019unsupervised} impose isometry regularisation in the spatial and spectral domain, respectively. Follow-up works~\cite{donati2020deep,sharma2020weakly} use point-based networks~\cite{qi2017pointnet++,thomas2019kpconv} and lead to better matching performance. Later, learnable implicit diffusion processes pushed the state of the art in shape matching for a broad range settings~\cite{sharp2020diffusionnet}, including near-isometry~\cite{attaiki2023understanding}, non-isometry~\cite{donati2022deep,li2022learning,cao2024spectral}, partiality~\cite{attaiki2021dpfm}, multi-shape matching~\cite{cao2022unsupervised} as well as multi-modal matching~\cite{cao2023self}. Meanwhile, instead of solely learning features, several works~\cite{marin2020correspondence,jiang2023neural} also attempted to learn the basis functions at the same time. 

One important ingredient of our work is the utilisation of recent learned feature embeddings. While existing shape matching approaches that are purely learning-based do not provide any guarantees about geometric consistency (i.e.,~the smoothness of maps), we propose to combine learned embeddings with an axiomatic optimisation approach in order to guarantee geometric conistency via hard constraints.

\myparagraph{Geometric Consistency.}
Several 3D shape matching methods provide geometrically consistent matchings by performing local optimisation on initial maps between two shapes.
Continuous maps starting from user-defined correspondences are computed in~\cite{schreiner2004inter}.
Refinement of an initial matching in a geometrically consistent manner using heat diffusion in~\cite{sharma2011topologically}.
3D shape matching is casted as image matching problem to obtain smooth maps via a smooth surface-to-image parameterisation in\cite{litke2005image}.
In~\cite{vestner2017product} the so-called product manifold filter is used to find smooth maps via a linearisation of the quadratic assignment problem.
Local map refinement of initial maps is done in~\cite{ezuz2017deblurring} via denoising, or in~\cite{ezuz2019elastic} via minimising an elastic energy.
Common triangulations between two shapes have been utilised in~\cite{schmidt2020inter} or via intrinsic triangulation in~\cite{takayama2022compatible}.
Mapping of shapes to spheres and then optimising for a homeomorphism is done in~\cite{schmidt2023surface}. 

While all of the above methods provide high quality and smooth maps they rely on initial correspondences which are non-trivial to find.
Thus these methods can only be considered as \emph{local} shape matching methods.
Our approach allows to infer smooth maps (i.e.,~a discrete analogue to a diffeomorphism) between 3D shapes based on an efficient relaxation solver for integer programs.

\textbf{Algorithms for Global Shape Matching.}
Practically relevant and globally optimal 3D shape matching is considered extremely challenging.
Nevertheless, in certain cases respective problems are solvable efficiently: for example 2D to 2D matching via dynamic programming~\cite{coughlan2000efficient} analogous to
dynamic time warping~\cite{sankoff1983time}, or via graph cuts~\cite{schmidt2009planar}; another example is 
contour-to-image matching via dynamic programming~\cite{felzenszwalb2005representation} or shortest paths~\cite{schoenemann2009combinatorial}.
Matching a 2D contour to a 3D mesh can also be phrased as shortest path problem~\cite{lahner2016efficient, roetzer2023conjugate}.

For the 3D to 3D matching scenario there exist several sparse matching methods~\cite{maron2016point,gasparetto2017spatial,bernard2020mina},
or the popular functional map \cite{ovsjanikov2012functional} approach that operates in the spectral domain.
Relaxations to bi-stochastic matrices for the QAP are used in~\cite{vestner2017efficient} to map nearby points on one shape to nearby points on the other shape.
Furthermore, an optimisation problem to find a dense discrete diffeomorphism between two 3D shapes is introduced in~\cite{windheuser2011geometrically}.
While this optimisation problem purely relies on linear constraints to ensure geometric consistency, it is nevertheless hard to solve since it optimises over a large number of binary variables.
Thus, various strategies to address this problem have been proposed: a linear programming relaxation~\cite{windheuser2011geometrically} combined with a coarse-to-fine scheme~\cite{windheuser2011large}, %
as well as an approach using a combinatorial solver with a problem-specific primal heuristic~\cite{roetzer2022scalable}.
Still, the mentioned methods
do not scale to practically relevant resolutions and oftentimes cannot find good lower-bounds (which as a consequence results in bad primal solutions in many cases).%
We show that with our combinatorial optimiser we can find solutions much faster and often times find a globally optimal solution due to faster and better lower bound computation.

\section{Background}\label{sec:background}
We first introduce the integer linear programming (ILP) formulation\rebuttalfix{\cite{windheuser2011geometrically}} for \rebuttalfix{geometrically consistent} shape matching \rebuttalfix{by ensuring neighbourhood consistency of matched elements.} Further, we touch upon the method \rebuttalfix{\cite{lange2021efficient} (which solves large-scale ILPs based on smaller subproblems), since we build on it.}
Our notation is summarised in \cref{table:notation}.

\begin{table}[htb!]
\centering
{\footnotesize
        \begin{tabularx}{\columnwidth}{lp{2.2cm}p{0.5cm}lp{4.3cm}}
        \toprule
        \textbf{Symbol} & \textbf{Description}& & \textbf{Symbol} & \textbf{Description}\\
        \toprule
        $\mathcal{M}=(V_\mathcal{M}, T_\mathcal{M})$ & Triangle mesh  
                &&  $x\in\{0,1\}^{|P|}$ & Indicator representation of $P$ \\
        $V_\mathcal{M}\in \mathbb{R}^{|V_\mathcal{M}|\times 3}$ & Vertices of $\mathcal{M}$  
                &&    $c\in\mathbb{R}_+^{|P|}$ &Cost vector\\
        $T_\mathcal{M}\in \mathbb{N}^{|T_\mathcal{M}|\times 3}$ & Triangles of $\mathcal{M}$  
                &&  $A^\partial x = \textbf{0}$ & Geo.\ consistency constraints \\
        $F_\mathcal{M}\in \mathbb{R}^{|V_\mathcal{M}|\times 128}$ & Features of $\mathcal{M}$  
                &&  $A^{\mathcal{M}} x = \textbf{1}$ & Projection on $\mathcal{M}$ constraints  \\
        $\mathcal{N}=(V_\mathcal{N}, T_\mathcal{N})$ & Triangle mesh  
                &&  $A^{\mathcal{N}} x = \textbf{1}$ & Projection on $\mathcal{N}$ constraints \\
        $\dots$ & (analog. to $\mathcal{M}$)  
                &&  $\mathcal{I}$, $\mathcal{J}$ & Set of vars., constraints resp. \\
        $P$ & Product space
                &&  $\mathcal{I}_j$, ($\mathcal{J}_i$) & Vars.\ of $j$ in $\mathcal{J}$ (viceversa) \\
        \bottomrule
        \end{tabularx}
}
	\caption{Summary of the notation used in this paper.
	}
	\label{table:notation}
\end{table}

\subsection{3D Shape Matching as ILP}
\begin{figure}
    \centering
    {\scriptsize
    \begin{tabular}{c|c}
         \includegraphics[width=0.44\columnwidth, trim={0.2cm, 0cm, 0cm, 0.7cm}]{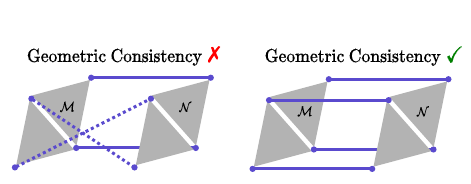}
         &
         \includegraphics[width=0.44\columnwidth]{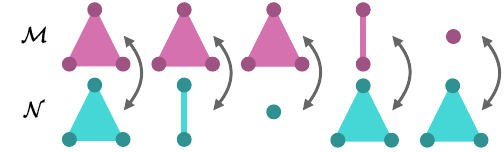}\\
         Maps Between Vertices ({\color{cBLUE} \rule[0.4mm]{0.3cm}{0.9mm}})
         &
         Matching Elements
    \end{tabular}
    }
    \caption{\textit{Left:} Geometric consistency is an essential property of correspondences between shapes, meaning that neighbouring elements in $\mathcal{M}$ are consistently matched to neighbouring elements in $\mathcal{N}$. \textit{Right:} {Matching elements} of the approach by \cite{windheuser2011geometrically}: triangle-triangle, triangle-edge and triangle-vertex between shapes $\mathcal{M}$ and $\mathcal{N}$. At least one triangle is involved in every matching element.
    }
    \label{fig:corres-types}
    \label{fig:geo-cons}
\end{figure}

The matching formalism in~\cite{windheuser2011geometrically} aims to find a \emph{discrete diffeomorphism} between two triangle meshes $\mathcal{M}$ and $\mathcal{N}$ by solving a constrained binary integer linear program.
The matching is found within the search space $P$ (i.e.~the {product space}) which consists of three types of potential correspondences between $\mathcal{M}$ and $\mathcal{N}$: triangle-triangle, triangle-edge and triangle-vertex, see \cref{fig:corres-types} right.
The latter two are needed to account for different triangulation as well as stretching and compression of shapes.
Employing geometric consistency constraints $A^\partial x = \textbf{0}$ ensures that neighbouring elements in $\mathcal{M}$ are matched only to neighbouring elements in $\mathcal{N}$, see \cref{fig:geo-cons} left.
Additional projection constraints $A^{\mathcal{M}} x = \textbf{1}$ and $A^{\mathcal{N}} x = \textbf{1}$ ensure that each triangle in $T_\mathcal{M}$ and $T_\mathcal{N}$ respectively are matched exactly once and thus ensure surjectivity.
A solution $x$ fulfilling all constraints yields a {discrete diffeomorphism} between shapes $\mathcal{M}$ and $\mathcal{N}$ \cite{windheuser2011geometrically}. 
The full optimisation problem reads

\begin{equation}\label{eq:ilp-sm}
    \underset{x\in\{0,1\}^{|P|}}{\min} c^\top x \quad\text{s.t.}\; 
    \underbrace{
    \begin{bmatrix}
    A^\partial\\
    A^{\mathcal{M}}\\
    A^{\mathcal{N}}
    \end{bmatrix}}_{\coloneqq A} x = 
    \underbrace{
    \begin{bmatrix}
    \textbf{0}\\
    \textbf{1}\\
    \textbf{1}
    \end{bmatrix}}_{\coloneqq b}.
    \tag{P}
\end{equation}
For a detailed derivation of~\eqref{eq:ilp-sm} and more insights we refer readers to~\cite{windheuser2011geometrically, windheuser2011large, roetzer2022scalable}.
As mentioned, the solution of this optimisation problem yields a discrete diffeomorphism between two triangle meshes. This ensures geometric consistency.
Since the problem is a difficult and large combinatorial optimisation problem, only heuristic solvers (cf.\ \cite{windheuser2011large, roetzer2022scalable}) exists to solve the problem for reasonable shape resolutions.
We show that by combining deep feature embeddings and adapting a novel combinatorial solver, we can solve the problem significantly faster than before, while at the same time certifying global optimality in most cases.

\subsection{Dual Decomposition for ILPs}

In order to solve large shape matching integer linear programs (ILPs)~\eqref{eq:ilp-sm}, we take~\cite{abbas2022fastdog} as a starting point. Its massive parallelism and generic applicability make it a suitable choice. We give an overview of their approach below. 

Since ILPs are NP-hard to solve, a relaxation based on Lagrangian dual is proposed in~\cite{abbas2022fastdog}.
Formally, denote $\mathcal{I}$ and $\mathcal{J}$ as the set of variables and constraints, respectively, of the ILP~\eqref{eq:ilp-sm}. Moreover, denote the variables in constraint $j$ as $\mathcal{I}_j$, and rows containing variable $i$ as $\mathcal{J}_i$. Then represent each constraint $j$ as an independent subproblem $\mathcal{S}_j \coloneqq \{ s \in \{0,1\}^{\mathcal{I}_j} : \sum_{i \in \mathcal{I}_j} a_{ji} x_{i} = b_j \}$. The relaxed (dual) problem is written as
\begin{align}
\max_\lambda \quad {\sum_{j \in \mathcal{J}} \min_{s \in \mathcal{S}_j} s^\top \lambda^{j}} \quad
\text{s.t.} \quad \sum_{j \in \mathcal{J}_i} \lambda^j_i = c_i \quad \forall i \in \mathcal{I}, \label{eq:dual-problem}
\tag{D}
\end{align}
where $\lambda^j_i$ is the dual variable for primal variable $i$ and row (constraint) $j$ and $\lambda^j$ is the vector containing all variables of $j$, i.e., $\lambda^j = (\lambda_i^j)_{i \in \mathcal{I}_j}$. If optima of individual subproblems $\mathcal{S}_j$ agree, then the dual problem  solves the original problem~\eqref{eq:ilp-sm}. 
In general it provides a lower bound to the original problem. For more details and derivations we refer to~\cite{lange2021efficient, abbas2022fastdog}. 

To optimise the dual problem~\eqref{eq:dual-problem} a parallel block coordinate ascent algorithm is proposed in~\cite{abbas2022fastdog}. Their update scheme also guarantees that dual objective improves or stays the same during dual optimisation. To find a feasible solution to~\eqref{eq:ilp-sm} (i.e., a matching in our case) it critically relies on optimising the dual problem. Thus, an improved scheme for solving the dual not only helps in finding better lower bounds but also in recovering better solutions of the original problem. 

    \section{Method}\label{sec:method}
\begin{figure*}[h]
    \centering
    \includegraphics[width=0.95\linewidth]{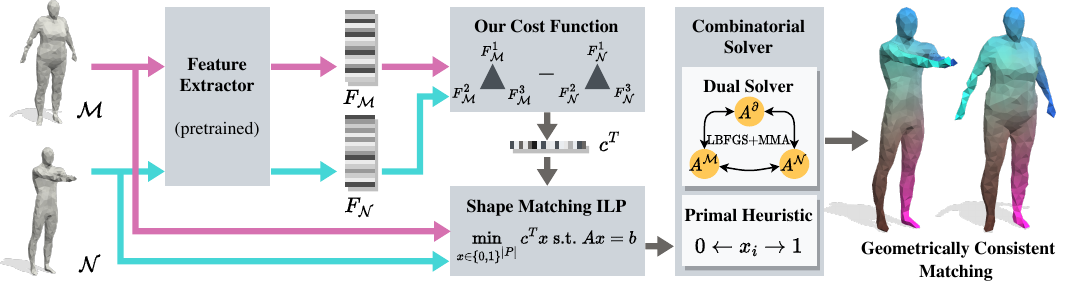}
    \caption{Illustration of our {shape matching pipeline}. 
    We use a pretrained feature extractor to define the cost function of ILP~\eqref{eq:ilp-sm}, which we solve using our combinatorial solver to find a geometrically consistent matching between shapes. The dual solver performs second-order quasi-Newton updates on top of the first-order min-marginal averaging scheme (MMA) of~\cite{abbas2022fastdog}. For primal recovery we adapt the approach of~\cite{abbas2022fastdog} by replacing their dual optimisation procedure by ours leading to faster and better-quality solutions. %
    }
    \label{fig:pipeline}
\end{figure*}

We now present our shape matching framework. First, we will discuss improvements to~\cite{abbas2022fastdog} for efficiently solving the ILP~\eqref{eq:ilp-sm}.
Next we propose our improved cost function which helps to obtain more accurate results by injecting more information into the optimisation problem. The dataflow of our proposed matching pipeline is shown in~\cref{fig:pipeline}.
\rebuttalfix{We note that our approach is initialisation-free in two ways: (i) we maximise a concave optimisation problem (i.e.~Problem~\eqref{eq:dual-problem}), so that any local optimum is a global optimum and initialisation of the solver does not play a significant role.
(ii) Our approach does not require landmark correspondence~\cite{schmidt2023surface} or heuristically chosen initial matchings~\cite{roetzer2022scalable}.}

\subsection{Faster Optimisation}
\label{sec:faster_optimisation}
Although the method of~\cite{abbas2022fastdog} delivers good performance on large scale ILPs, it exhibits two weaknesses. First, it can be quite slow to converge as it only relies on first-order updates for optimising the dual~\eqref{eq:dual-problem} and indirectly also the associated primal~\eqref{eq:ilp-sm}. 
We aim for better optimisation of the dual problem by leveraging second-order information.
Second, the equalities $A^{\mathcal{M}} x = \textbf{1}$ and $A^{\mathcal{N}} x = \textbf{1}$ involve a large number of variables (cf. \cref{fig:sparse-pattern}), thus creating large subproblems, each of which must be processed sequentially, causing bottlenecks when computing on GPUs. 
To tackle this issue we propose a technique which splits a subproblem into multiple smaller ones that can be solved in parallel.

\myparagraph{Quasi-Newton Updates.}
We incorporate second-order information in optimising the dual problem~\eqref{eq:dual-problem} through quasi-Newton updates. To this end we interleave L-BFGS~\cite{nocedal1980updating_lbfgs, liu1989lbfgs} updates with the first-order min-marginal averaging (MMA) scheme of~\cite{abbas2022fastdog}. An iteration of our scheme is given in Algorithm~\ref{alg:lbfgs}.

\begin{algorithm}
\footnotesize
\newcommand\mycommfont[1]{\footnotesize\ttfamily\textcolor{cBLUE}{#1}}
\SetCommentSty{mycommfont}
\DontPrintSemicolon
\KwInput{
Dual variables $\lambda$,
Hessian inverse estimate $\hat{H}$,
Previous step size $\gamma$
}
$g^j(\lambda) \coloneqq \arg\min_{s \in \mathcal{S}_j} s^\top \lambda^{j}$, \, $\forall j \in \mathcal{J}$
\label{alg:lbfgs_subgradient}
\tcp*[r]{Subgradient of dual obj.~\eqref{eq:dual-problem}}
$\hat{d} = \hat{H}g(\lambda)$\label{alg:lbfgs_update_direction}
\tcp*[r]{Compute update direction}
$d^j_i = \hat{d}^j_i - \frac{1}{\lvert \mathcal{J}_i \rvert}\sum_{k \in \mathcal{J}_i} \hat{d}^k_i, \, \forall i \in \mathcal{I}, j \in \mathcal{J}_i$
\label{alg:dual_feasible_update}
\tcp*[r]{Ensure dual feasibility~\eqref{eq:dual-problem}}
$\gamma \leftarrow$ \texttt{FindStepSize}$(\lambda, d, \gamma)$
\label{alg:find_step_size}\; 
$\lambda \leftarrow \lambda + \gamma d $
\label{alg:lbfgs_step}
\tcp*[r]{Update Dual variables}
$\lambda \leftarrow $ \texttt{MMA}$(\lambda)$
\label{alg:lbfgs_pmma}
\tcp*[r]{First-order update of~\cite{abbas2022fastdog}}
$\hat{H} \leftarrow $ \texttt{L-BFGS}$(\hat{H}, g(\lambda))$
\label{alg:lbfgs_update_hessian}
\tcp*[r]{Update inverse Hessian estimate}
\Return{$\lambda, \hat{H}$, $\gamma$}
\caption{Quasi-Newton powered MMA~\eqref{eq:dual-problem}}
\label{alg:lbfgs}
\end{algorithm}

In detail, we first compute a subgradient of the dual objective by finding a minimising assignment for each subproblem. Next, we calculate the search direction $\hat{d}$ by multiplying the subgradient with an estimate of the inverse Hessian. Naively updating the dual variables in this search direction however, can violate dual feasibility~\eqref{eq:dual-problem}. To address this, we project $\hat{d}$ to a feasible search direction $d$ such that $\sum_j d_i^j$ is zero for all $i$. Subsequently, a suitable step size is determined that provides sufficient increase in the objective~\eqref{eq:dual-problem}. Using this step size, we perform the quasi-Newton update, followed by a min-marginal averaging (MMA) update of~\cite{abbas2022fastdog}. Finally, we update the estimate of the inverse Hessian for use in subsequent iterations. More detailed information about step size selection and inverse Hessian updates can be found in the Appendix. Note that due to its general nature, our second-order update scheme is applicable to all problems which fall under the framework of~\cite{abbas2022fastdog}.

In essence, our algorithm combines two distinct update schemes for dual optimisation. It employs a first-order update through min-marginal averaging (MMA), which is fast and guarantees non-decreasing objective in its updates. Our L-BFGS scheme leverages second-order information across MMA iterations and facilitate faster convergence towards the optimum. Empirically, we observe this hybrid scheme to be significantly faster than either L-BFGS or MMA alone. 

\myparagraph{Constraint Splitting.}
Similar to~\cite{abbas2022fastdog} we represent each constraint as a binary decision diagram (BDD)~\cite{bryant1986graph}, a general purpose data structure for encoding functions of binary variables.
Essentially, BDDs are directed acyclic graphs and optimal solutions can be found by computing a shortest path.
For an example of a BDD see Figure~\ref{fig:BDD} in the Appendix.
Shortest paths and other needed computations require sequentially going through nodes in each of the BDDs.
These computations can be parallelized across BDDs on GPUs.
However, when a few long BDDs are present, the full parallelism of GPUs can often not be exploited.
For this reason we propose a BDD splitting technique that takes a BDD as input, together with the desired smaller chunk size, and outputs a number of sub-BDDs of chunk size length that together are equivalent to the original BDD.

The high level idea is as follows:
Let us assume we have a constraint of the form $\sum_{i=1}^k x_i = 2$ and chunk size $k/2$.
By introducing auxiliary variables $y_1,y_2 \in \{0,1\}$ we can split the constraint into two equal-sized sub-constraints
$\sum_{i=1}^{\lfloor k/2 \rfloor} x_i - y_1 - y_2 = 0$ and
$y_1 + y_2 + \sum_{i =\lceil k/2 \rceil}^{k} x_i = 2$.
The technical procedure to do this for BDDs representing arbitrary (also non-arithmetic) constraints is detailed in the Appendix, Section~\ref{sec:bdd-splitting}.

\subsection{Energy Adaption}

We use a pretrained feature extractor to predict per-vertex features $F_\mathcal{M}$ and $F_\mathcal{N}$ for both shapes.
These features are used in our adaption of the costs in~\eqref{eq:ilp-sm} as
\begin{equation}\label{eq:cost}
    c_p = \sum_{v=1}^{3} (A_{m_v}^\mathcal{M} + A_{n_v}^\mathcal{N}) ||F_{m_v}^\mathcal{X} - F_{n_v}^\mathcal{Y}||.
\end{equation}
Here $m_v \in V_\mathcal{M}$, $n_v \in V_\mathcal{N}$ and $\big((m_1, n_1)$, $(m_2, n_2)$, $(m_3, n_3)\big)$ forms the $p$-th product triangle in $P$.
$A_{m_v}^\mathcal{M}$ and $A_{n_v}^\mathcal{N}$ are the mixed areas according to~\cite{meyer2003discrete} at vertices $m_v$ and $n_v$ on shapes $\mathcal{M}$ and $\mathcal{N}$, respectively.
Using our learned cost function allows to infuse more descriptive information into the problem compared to the previously used hand designed costs based on elastic energy  from~\cite{windheuser2011geometrically}.
Empirically this yields not only in improved matching accuracy but also reduces ILP solver runtimes due to easier optimisation. 

\section{Experiments}\label{sec:experiments}
In this section we experimentally analyse on four shape matching datasets the quality of shape matching as well as runtime of our method and finally show how shapes with higher resolutions can be tackled.

\myparagraph{Datasets.}
We consider the challenging remeshed version FAUST\_r~\cite{ren2018continuous,donati2020deep} of FAUST~\cite{bogo2014faust}, consisting of 100 human shapes (190 pairs in test set).
Furthermore, we choose the dataset SMAL~\cite{zuffi20173d} with 49 animal shapes of eight species (136 pairs in test set). 
We also consider the SHREC'20 dataset~\cite{dyke2020track} which consists of $10$ highly non-isometric deformed animal shapes.
Lastly, we consider DT4D-H~\cite{magnet2022smooth}, which contains nine classes of humanoid shapes sampled from the DeformingThings 4D~\cite{li20214dcomplete} dataset.
We randomly pick 100 intra-class (DT4d-Intra) and 100 inter-class (DT4d-Inter) pairs for evaluation.
We split the datasets for training and testing as done in~\cite{cao2023unsupervised} and consequently evaluate only on unseen shapes (except for SHREC'20). We evaluate all shape matching methods by downsampling shapes to $450$ faces using~\cite{libigl}, unless stated otherwise. We compute features on high-resolution shapes and transfer them to the low-resolution shapes using nearest-neighbour search.
Furthermore, we evaluate all methods only on shape pairs which have the same genus, since otherwise there does not exist a solution to \eqref{eq:ilp-sm}, i.e.,~the constraints of the optimisation problem cannot be fulfilled.
We repair any defects~(non-manifoldness, open boundaries) by~\cite{attene2010lightweight, sullivan2019pyvista}.

\myparagraph{Methods.}
\vspace{0.5em}

\begin{packed_description}
\item[\textnormal{\lprelax~\cite{windheuser2011geometrically}:}]
The relaxation based approach for solving the ILP~\eqref{eq:ilp-sm}. Here, binary constraints are relaxed continuously and the largest element in $x$ is iteratively set to one until the constraints are fulfilled. Since this approach does not scale to larger problems we consider it only in some of the experiments. 
\item[\textnormal{\smcomb~\cite{roetzer2022scalable}:}]
We compare to the shape matching solver for solving the ILP formulation~\eqref{eq:ilp-sm}. Here, the dual is optimised to obtain improved costs on the primal which drive a custom primal heuristic.
\item[\textnormal{\fastdog~\cite{abbas2022fastdog}:}]
A general purpose solver for binary optimisation problems including the ILP~\eqref{eq:ilp-sm}. It shows good scalability on problems from  vision and machine learning due to massively parallel dual and primal optimisation routines.
\item[\textnormal{\caoetal~\cite{cao2023unsupervised}:}]
The current state-of-the-art for 3D shape matching based on an unsupervised learning-based scheme which, however, does not enforce geometric consistency. 
We compare against this method to also evaluate if imposing geometric consistency can further improve shape matching performance.
\item[\textnormal{\kernelm~\cite{vestner2017efficient}:}]
Relaxes a permutation matrix to the set of bi-stochastic matrices while using so-called kernel matrices to efficiently compute solutions of a quadratic assignment problem. For a fair comparison we use the features computed with the pretrained feature extractor of~\cite{cao2023unsupervised}.
\item[\textnormal{\ours:}]
The ILP formulation with our improved cost function~\eqref{eq:cost} based on the pretrained feature extractor of~\cite{cao2023unsupervised}. For solving the dual~\eqref{eq:dual-problem} we use our quasi-Newton scheme along with the BDD splitting approach from Sec.~\ref{sec:faster_optimisation}. For recovering a primal solution we modify the approach of~\cite{abbas2022fastdog} by replacing their dual solver by ours. 
\end{packed_description}

We will also utilize our improved cost function~\eqref{eq:cost} in evaluation of all existing ILP based solvers for a fair comparison unless state otherwise. 
For CPU-based solvers (\lprelax\, \smcomb) we use an Intel Core i9-12900K CPU and 64GB RAM. For GPU-based solvers we use an NVIDIA A40 GPU with 48GB DRAM.

\myparagraph{Metrics.}
We evaluate shape matching methods w.r.t.\ the following metrics 
\label{sec:metrics}
\begin{packed_description}
\item[\textnormal{Geodesic Distances:}]
We follow the Princeton protocol~\cite{kim2011blended} for evaluation matching quality. To do so, we compute geodesic distances on the meshes and normalise them by the square-root of the area of shape (cf.~\cite[Sec.~8.2]{kim2011blended}).
\item[\textnormal{Conformal Distortion:}]
We show conformal distortion errors~\cite{hormann2000mips} to evaluate geometric consistency and smoothness of the computed point map. Intuitively, the conformal distortion error evaluates local consistency of triangles after matching (cf.~\cite[Eq.~3]{hormann2000mips}).
\item[\textnormal{Optimality gaps:}]
To quantify the optimality of a primal solution with objective $p$, we report the primal dual gap $(p - d_{max}) / p$, where $d_{max}$ is the maximum dual objective by the respective solver. 
Note that the primal dual gap cannot be lower than the integrality gap, since~\eqref{eq:dual-problem} is a relaxation of the original ILP~\eqref{eq:ilp-sm}. Therefore the primal dual gap provides an upper bound to the optimality gap. 
We consider a solution to be globally optimal if the primal-dual gap is less than $10^{-2}$.
Moreover, we evaluate our dual optimisation scheme (Alg.~\ref{alg:lbfgs}) through relative dual gaps $(d^* - d_t) / d^*$, where $d_t$ is the dual objective at time $t$ and $d^*$ is the best known objective for the given instance. 
\end{packed_description}

\subsection{Results}

\begin{figure*}[!h]
    \centering
    \setlength{\tabcolsep}{-10pt}
    \renewcommand{\arraystretch}{0.9}
    \begin{tabular}{cccc}
    \hspace{-0.6cm}
         \newcommand{\pckLineWidth}{3pt}
\newcommand{\plotWidth}{0.55\columnwidth}
\newcommand{\plotHeight}{0.43\columnwidth}
\newcommand{\pckTitle}{}

\pgfplotsset{%
    label style = {font=\normalfont},
    tick label style = {font=\normalfont},
    title style =  {font=\large},
    legend style={  fill= gray!10,
                    fill opacity=0.6, 
                    font=\normalfont,
                    draw=gray!20, %
                    text opacity=1}
}
\begin{tikzpicture}[scale=0.5, transform shape]
	\begin{axis}[
		width=\plotWidth,
		height=\plotHeight,
		grid=major,
		title=\pckTitle,
		legend style={
			at={(0.97,0.03)},
			anchor=south east,
			legend columns=1},
		legend cell align={left},
        title style={yshift=-0.1cm},
	ylabel={{\normalfont$\%$ Correct Matchings ($\rightarrow$)}},
        xlabel={Geodesic Error Threshold},
        xmin=0,
        xmax=0.25,
        ylabel near ticks,
        xtick={0, 0.05, 0.1, 0.15, 0.2, 0.25},
        xticklabels={$0$, $0.05$, $0.1$, $0.15$, $0.2$, $0.25$},
        ymin=0,
        ymax=1,
        ytick={0, 0.20, 0.40, 0.60, 0.80, 1},
        yticklabels={$0$, $20$, $40$, $60$, $80$, $100$},
	]
    \addplot [color=cPLOT2, smooth, line width=\pckLineWidth]
    table[row sep=crcr]{%
0	0.0656071719641402\\
0.00862068965517241	0.0664221678891606\\
0.0172413793103448	0.126460201032328\\
0.0258620689655172	0.315946753599565\\
0.0344827586206897	0.455827220863896\\
0.0431034482758621	0.578429774517794\\
0.0517241379310345	0.697935343656615\\
0.0603448275862069	0.775522955718555\\
0.0689655172413793	0.825400706329802\\
0.0775862068965517	0.858000543330617\\
0.0862068965517241	0.878293941863624\\
0.0948275862068965	0.891795707688128\\
0.103448275862069	0.901358326541701\\
0.112068965517241	0.908312958435208\\
0.120689655172414	0.913501765824504\\
0.129310344827586	0.91841890790546\\
0.137931034482759	0.921977723444716\\
0.146551724137931	0.925808204292312\\
0.155172413793103	0.929530019016572\\
0.163793103448276	0.932980168432491\\
0.172413793103448	0.936294485194241\\
0.181034482758621	0.939853300733496\\
0.189655172413793	0.942542787286064\\
0.198275862068966	0.945340939961967\\
0.206896551724138	0.947921760391198\\
0.21551724137931	0.949877750611247\\
0.224137931034483	0.951779407769628\\
0.232758620689655	0.953816897582179\\
0.241379310344828	0.955609888617224\\
0.25	0.957294213528932\\
    };
    \addlegendentry{\textcolor{black}{\caoetal: 0.063}}
    \addplot [color=cPLOT4, smooth, line width=\pckLineWidth]
    table[row sep=crcr]{%
0	0.0447704428144526\\
0.00862068965517241	0.0453681064928009\\
0.0172413793103448	0.0809834284161913\\
0.0258620689655172	0.214805759304537\\
0.0344827586206897	0.320565063841347\\
0.0431034482758621	0.427546862265689\\
0.0517241379310345	0.556452051073078\\
0.0603448275862069	0.658299375169791\\
0.0689655172413793	0.742080956261885\\
0.0775862068965517	0.811301276826949\\
0.0862068965517241	0.862401521325727\\
0.0948275862068965	0.903069817984243\\
0.103448275862069	0.931268676989948\\
0.112068965517241	0.951073077967943\\
0.120689655172414	0.964303178484108\\
0.129310344827586	0.974191795707688\\
0.137931034482759	0.980603096984515\\
0.146551724137931	0.986416734582994\\
0.155172413793103	0.99051888073893\\
0.163793103448276	0.993778864439011\\
0.172413793103448	0.995381689758218\\
0.181034482758621	0.996604183645748\\
0.189655172413793	0.997555012224939\\
0.198275862068966	0.998234175495789\\
0.206896551724138	0.998614506927465\\
0.21551724137931	0.998886172235805\\
0.224137931034483	0.999157837544146\\
0.232758620689655	0.999293670198316\\
0.241379310344828	0.999429502852486\\
0.25	0.999483835914154\\
    };
    \addlegendentry{\textcolor{black}{\kernelm: 0.053}}
    \addplot [color=cPLOT0, smooth, line width=\pckLineWidth]
    table[row sep=crcr]{%
0	0.0552301424836336\\
0.00862068965517241	0.0558381807495085\\
0.0172413793103448	0.110257605545309\\
0.0258620689655172	0.282960740995967\\
0.0344827586206897	0.423559456008431\\
0.0431034482758621	0.563469060986238\\
0.0517241379310345	0.709823871582318\\
0.0603448275862069	0.814264577717424\\
0.0689655172413793	0.886580595472142\\
0.0775862068965517	0.932730699852044\\
0.0862068965517241	0.958268307018788\\
0.0948275862068965	0.972678813920023\\
0.103448275862069	0.980704919029571\\
0.112068965517241	0.985731368694136\\
0.120689655172414	0.988589148543748\\
0.129310344827586	0.990615942763331\\
0.137931034482759	0.992115770485823\\
0.146551724137931	0.993473722612943\\
0.155172413793103	0.994284440300776\\
0.163793103448276	0.995135693873001\\
0.172413793103448	0.995865339792051\\
0.181034482758621	0.99620989480938\\
0.189655172413793	0.996676057479884\\
0.198275862068966	0.997000344555017\\
0.206896551724138	0.997223291919171\\
0.21551724137931	0.997446239283326\\
0.224137931034483	0.997689454589676\\
0.232758620689655	0.997871866069438\\
0.241379310344828	0.997993473722613\\
0.25	0.998094813433592\\
    };
    \addlegendentry{\textcolor{black}{\smcomb: 0.042}}
    \addplot [color=cPLOT3, dotted, smooth, line width=\pckLineWidth]
    table[row sep=crcr]{%
0	0.0555818890010923\\
0.00862068965517241	0.0561795437215341\\
0.0172413793103448	0.110834037466768\\
0.0258620689655172	0.284524864497249\\
0.0344827586206897	0.425509552171135\\
0.0431034482758621	0.565937802691507\\
0.0517241379310345	0.712734167302104\\
0.0603448275862069	0.817942006883334\\
0.0689655172413793	0.890629186159141\\
0.0775862068965517	0.936504338148919\\
0.0862068965517241	0.961399748572842\\
0.0948275862068965	0.975372503761103\\
0.103448275862069	0.983554190796117\\
0.112068965517241	0.988232384642335\\
0.120689655172414	0.990952744059518\\
0.129310344827586	0.992889969705088\\
0.137931034482759	0.993982235228654\\
0.146551724137931	0.995239371019929\\
0.155172413793103	0.995919460874225\\
0.163793103448276	0.996620159511984\\
0.172413793103448	0.997217814232426\\
0.181034482758621	0.997547554767842\\
0.189655172413793	0.997815468952868\\
0.198275862068966	0.998165818271747\\
0.206896551724138	0.998330688539456\\
0.21551724137931	0.998454341240237\\
0.224137931034483	0.998639820291408\\
0.232758620689655	0.998742864208726\\
0.241379310344828	0.998845908126043\\
0.25	0.998845908126043\\
    };
    \addlegendentry{\textcolor{black}{\fastdog: \textbf{0.041}}}
    \addplot [color=cPLOT5, dashed, smooth, line width=\pckLineWidth]
    table[row sep=crcr]{%
0	0.0553075446924553\\
0.00862068965517241	0.0559135440864559\\
0.0172413793103448	0.110514089485911\\
0.0258620689655172	0.283587516412484\\
0.0344827586206897	0.424401575598424\\
0.0431034482758621	0.564609635390365\\
0.0517241379310345	0.711463488536511\\
0.0603448275862069	0.816361983638016\\
0.0689655172413793	0.889122310877689\\
0.0775862068965517	0.935198464801535\\
0.0862068965517241	0.96034743965256\\
0.0948275862068965	0.974608625391375\\
0.103448275862069	0.982910817089183\\
0.112068965517241	0.987819412180588\\
0.120689655172414	0.990526209473791\\
0.129310344827586	0.992505807494192\\
0.137931034482759	0.993657206342794\\
0.146551724137931	0.994889405110595\\
0.155172413793103	0.995677204322796\\
0.163793103448276	0.996465003534996\\
0.172413793103448	0.997071002928997\\
0.181034482758621	0.997333602666397\\
0.189655172413793	0.997636602363398\\
0.198275862068966	0.997939602060398\\
0.206896551724138	0.998141601858398\\
0.21551724137931	0.998343601656398\\
0.224137931034483	0.998586001413999\\
0.232758620689655	0.998727401272599\\
0.241379310344828	0.998808201191799\\
0.25	0.998828401171599\\
    };
    \addlegendentry{\textcolor{black}{\ours: \textbf{0.041}}}
        
	\end{axis}

\node[above,font=\large] at (current bounding box.north) {\textbf{FAUST}};
\end{tikzpicture}& 
         \input{figs/tikz/pck_smal}&
         \newcommand{\pckLineWidth}{3pt}
\newcommand{\plotWidth}{0.55\columnwidth}
\newcommand{\plotHeight}{0.43\columnwidth}
\newcommand{\pckTitle}{}

\pgfplotsset{%
    label style = {font=\normalfont},
    tick label style = {font=\normalfont},
    title style =  {font=\large},
    legend style={  fill= gray!10,
                    fill opacity=0.6, 
                    font=\normalfont,
                    draw=gray!20, %
                    text opacity=1}
}
\begin{tikzpicture}[scale=0.5, transform shape]
	\begin{axis}[
		width=\plotWidth,
		height=\plotHeight,
		grid=major,
		title=\pckTitle,
		legend style={
			at={(0.97,0.03)},
			anchor=south east,
			legend columns=1},
		legend cell align={left},
        title style={yshift=-0.1cm},
        xlabel={Geodesic Error Threshold},
        xmin=0,
        xmax=0.25,
        ylabel near ticks,
        xtick={0, 0.05, 0.1, 0.15, 0.2, 0.25},
        xticklabels={$0$, $0.05$, $0.1$, $0.15$, $0.2$, $0.25$},
        ymin=0,
        ymax=1,
        ytick={0, 0.20, 0.40, 0.60, 0.80, 1},
        yticklabels={$0$, $20$, $40$, $60$, $80$, $100$},
	]
    \addplot [color=cPLOT2, smooth, line width=\pckLineWidth]
    table[row sep=crcr]{%
0	0.0505941378399789\\
0.00862068965517241	0.0547663057829416\\
0.0172413793103448	0.141008714021653\\
0.0258620689655172	0.297755479271191\\
0.0344827586206897	0.440559809875891\\
0.0431034482758621	0.570055452865065\\
0.0517241379310345	0.664008449960391\\
0.0603448275862069	0.721837866385001\\
0.0689655172413793	0.762397676260892\\
0.0775862068965517	0.787430683918669\\
0.0862068965517241	0.807921837866385\\
0.0948275862068965	0.822709268550304\\
0.103448275862069	0.83580670715606\\
0.112068965517241	0.846844467916557\\
0.120689655172414	0.855505677317138\\
0.129310344827586	0.862899392659097\\
0.137931034482759	0.868497491418009\\
0.146551724137931	0.874306839186691\\
0.155172413793103	0.880168999207816\\
0.163793103448276	0.885714285714286\\
0.172413793103448	0.890784261948772\\
0.181034482758621	0.895431740163718\\
0.189655172413793	0.899339846844468\\
0.198275862068966	0.903617639292316\\
0.206896551724138	0.907314496963295\\
0.21551724137931	0.910060734090309\\
0.224137931034483	0.913335093741748\\
0.232758620689655	0.915922894111434\\
0.241379310344828	0.91872194349089\\
0.25	0.921362556113018\\
    };
    \addlegendentry{\textcolor{black}{\caoetal: 0.085}}
    \addplot [color=cPLOT4, smooth, line width=\pckLineWidth]
    table[row sep=crcr]{%
0	0.0373910747293372\\
0.00862068965517241	0.0401901241087932\\
0.0172413793103448	0.0983892263005017\\
0.0258620689655172	0.215157116451017\\
0.0344827586206897	0.325270662793768\\
0.0431034482758621	0.448692896752046\\
0.0517241379310345	0.562820174280433\\
0.0603448275862069	0.658568787958806\\
0.0689655172413793	0.737470293108001\\
0.0775862068965517	0.800580934776868\\
0.0862068965517241	0.847583839450753\\
0.0948275862068965	0.882651175072617\\
0.103448275862069	0.908423554264589\\
0.112068965517241	0.929442830736731\\
0.120689655172414	0.942276208080275\\
0.129310344827586	0.951993662529707\\
0.137931034482759	0.959651439133879\\
0.146551724137931	0.964932664378136\\
0.155172413793103	0.969474518088197\\
0.163793103448276	0.972590440982308\\
0.172413793103448	0.974861367837338\\
0.181034482758621	0.975970425138632\\
0.189655172413793	0.976921045682598\\
0.198275862068966	0.977501980459467\\
0.206896551724138	0.978452601003433\\
0.21551724137931	0.979191972537629\\
0.224137931034483	0.979772907314497\\
0.232758620689655	0.980406654343808\\
0.241379310344828	0.981040401373119\\
0.25	0.981462899392659\\
    };
    \addlegendentry{\textcolor{black}{\kernelm: 0.077}}
    \addplot [color=cPLOT0, smooth, line width=\pckLineWidth]
    table[row sep=crcr]{%
0	0.0447041292498333\\
0.00862068965517241	0.0488216148386338\\
0.0172413793103448	0.130936041723854\\
0.0258620689655172	0.284067291478766\\
0.0344827586206897	0.437669111015254\\
0.0431034482758621	0.598917689502372\\
0.0517241379310345	0.732363436727971\\
0.0603448275862069	0.82557546762872\\
0.0689655172413793	0.887063252421474\\
0.0775862068965517	0.923767695384495\\
0.0862068965517241	0.946002117564017\\
0.0948275862068965	0.95894278655739\\
0.103448275862069	0.96651111721109\\
0.112068965517241	0.972667738520058\\
0.120689655172414	0.976589153366535\\
0.129310344827586	0.979255715462139\\
0.137931034482759	0.981216422885377\\
0.146551724137931	0.982824202972432\\
0.155172413793103	0.98411826987177\\
0.163793103448276	0.985294694325713\\
0.172413793103448	0.986275048037332\\
0.181034482758621	0.987176973452021\\
0.189655172413793	0.987922042272852\\
0.198275862068966	0.988549468648288\\
0.206896551724138	0.98905925257833\\
0.21551724137931	0.989529822359907\\
0.224137931034483	0.98988274969609\\
0.232758620689655	0.990431747774597\\
0.241379310344828	0.990823889259245\\
0.25	0.991216030743892\\
    };
    \addlegendentry{\textcolor{black}{\smcomb: 0.044}}
    \addplot [color=cPLOT3, dotted, smooth, line width=\pckLineWidth]
    table[row sep=crcr]{%
0	0.0452647070522254\\
0.00862068965517241	0.0492422735770256\\
0.0172413793103448	0.132094984288612\\
0.0258620689655172	0.286464341116105\\
0.0344827586206897	0.44039616562587\\
0.0431034482758621	0.602442225846227\\
0.0517241379310345	0.736406666401496\\
0.0603448275862069	0.830555666043515\\
0.0689655172413793	0.892526152499901\\
0.0775862068965517	0.92896066186707\\
0.0862068965517241	0.95103615607971\\
0.0948275862068965	0.963446163637087\\
0.103448275862069	0.970963764368959\\
0.112068965517241	0.977049441151903\\
0.120689655172414	0.981066783341951\\
0.129310344827586	0.983612425917823\\
0.137931034482759	0.98548188218448\\
0.146551724137931	0.987033133129152\\
0.155172413793103	0.988385505747584\\
0.163793103448276	0.989419673044032\\
0.172413793103448	0.990533391670976\\
0.181034482758621	0.991408456306432\\
0.189655172413793	0.9920448669504\\
0.198275862068966	0.992561950598624\\
0.206896551724138	0.9930392585816\\
0.21551724137931	0.99343701523408\\
0.224137931034483	0.993755220556064\\
0.232758620689655	0.994152977208544\\
0.241379310344828	0.994312079869536\\
0.25	0.994709836522016\\
    };
    \addlegendentry{\textcolor{black}{\fastdog: \textbf{0.042}}}
    \addplot [color=cPLOT5, dashed, smooth, line width=\pckLineWidth]
    table[row sep=crcr]{%
0	0.0452613350125945\\
0.00862068965517241	0.0492758186397985\\
0.0172413793103448	0.13169080604534\\
0.0258620689655172	0.286445214105793\\
0.0344827586206897	0.44088476070529\\
0.0431034482758621	0.603156486146096\\
0.0517241379310345	0.737562972292191\\
0.0603448275862069	0.831194899244333\\
0.0689655172413793	0.892317380352645\\
0.0775862068965517	0.928723236775819\\
0.0862068965517241	0.950566750629723\\
0.0948275862068965	0.96316120906801\\
0.103448275862069	0.970599811083123\\
0.112068965517241	0.976778967254408\\
0.120689655172414	0.980714735516373\\
0.129310344827586	0.983312342569269\\
0.137931034482759	0.985083438287154\\
0.146551724137931	0.98665774559194\\
0.155172413793103	0.988035264483627\\
0.163793103448276	0.989097921914358\\
0.172413793103448	0.990160579345088\\
0.181034482758621	0.990987090680101\\
0.189655172413793	0.991695528967254\\
0.198275862068966	0.99216782115869\\
0.206896551724138	0.992679471032746\\
0.21551724137931	0.993033690176322\\
0.224137931034483	0.99334855163728\\
0.232758620689655	0.993702770780856\\
0.241379310344828	0.993899559193955\\
0.25	0.994293136020151\\
    };
    \addlegendentry{\textcolor{black}{\ours: \textbf{0.042}}}
        
	\end{axis}

\node[above,font=\large] at (current bounding box.north) {\textbf{DT4D-Intra}};
\end{tikzpicture}& 
         \newcommand{\pckLineWidth}{3pt}
\newcommand{\plotWidth}{0.55\columnwidth}
\newcommand{\plotHeight}{0.43\columnwidth}
\newcommand{\pckTitle}{}

\pgfplotsset{%
    label style = {font=\normalfont},
    tick label style = {font=\normalfont},
    title style =  {font=\large},
    legend style={  fill= gray!10,
                    fill opacity=0.6, 
                    font=\normalfont,
                    draw=gray!20, %
                    text opacity=1}
}
\begin{tikzpicture}[scale=0.5, transform shape]
	\begin{axis}[
		width=\plotWidth,
		height=\plotHeight,
		grid=major,
		title=\pckTitle,
		legend style={
			at={(0.97,0.03)},
			anchor=south east,
			legend columns=1},
		legend cell align={left},
        title style={yshift=-0.1cm},
        xlabel={Geodesic Error Threshold},
        xmin=0,
        xmax=0.25,
        ylabel near ticks,
        xtick={0, 0.05, 0.1, 0.15, 0.2, 0.25},
        xticklabels={$0$, $0.05$, $0.1$, $0.15$, $0.2$, $0.25$},
        ymin=0,
        ymax=1,
        ytick={0, 0.20, 0.40, 0.60, 0.80, 1},
        yticklabels={$0$, $20$, $40$, $60$, $80$, $100$},
	]
    \addplot [color=cPLOT2, smooth, line width=\pckLineWidth]
    table[row sep=crcr]{%
0	0.0330049675844068\\
0.00862068965517241	0.037888355645365\\
0.0172413793103448	0.110213016755073\\
0.0258620689655172	0.244758777469058\\
0.0344827586206897	0.368695798602341\\
0.0431034482758621	0.504672897196262\\
0.0517241379310345	0.629620274480088\\
0.0603448275862069	0.720131346299571\\
0.0689655172413793	0.792035025679885\\
0.0775862068965517	0.840363728214195\\
0.0862068965517241	0.872189946956302\\
0.0948275862068965	0.892649659004799\\
0.103448275862069	0.90511071819483\\
0.112068965517241	0.915130083354382\\
0.120689655172414	0.921192220257641\\
0.129310344827586	0.92540203755157\\
0.137931034482759	0.928264713311442\\
0.146551724137931	0.930706407341921\\
0.155172413793103	0.932979708680643\\
0.163793103448276	0.935421402711122\\
0.172413793103448	0.938199882125116\\
0.181034482758621	0.939968005388566\\
0.189655172413793	0.941651932306138\\
0.198275862068966	0.943251662877831\\
0.206896551724138	0.944935589795403\\
0.21551724137931	0.946451124021217\\
0.224137931034483	0.947461480171761\\
0.232758620689655	0.947966658247032\\
0.241379310344828	0.949145407089332\\
0.25	0.949987370548118\\
    };
    \addlegendentry{\textcolor{black}{\caoetal: 0.073}}
    \addplot [color=cPLOT4, smooth, line width=\pckLineWidth]
    table[row sep=crcr]{%
0	0.0153237349499032\\
0.00862068965517241	0.018018018018018\\
0.0172413793103448	0.0534646796329039\\
0.0258620689655172	0.127810053043698\\
0.0344827586206897	0.194998737054812\\
0.0431034482758621	0.28045802812158\\
0.0517241379310345	0.371390081670456\\
0.0603448275862069	0.453818304285594\\
0.0689655172413793	0.538014650164183\\
0.0775862068965517	0.61395975414667\\
0.0862068965517241	0.677948977014398\\
0.0948275862068965	0.73149785299318\\
0.103448275862069	0.780079144565126\\
0.112068965517241	0.817630714826976\\
0.120689655172414	0.851561842216048\\
0.129310344827586	0.875810389829081\\
0.137931034482759	0.898206617832786\\
0.146551724137931	0.918076955460133\\
0.155172413793103	0.932137745221857\\
0.163793103448276	0.943083270186074\\
0.172413793103448	0.952681653616233\\
0.181034482758621	0.959754146670035\\
0.189655172413793	0.966068872610929\\
0.198275862068966	0.971794224130673\\
0.206896551724138	0.974656899890545\\
0.21551724137931	0.97794055737981\\
0.224137931034483	0.98029805506441\\
0.232758620689655	0.982234571019618\\
0.241379310344828	0.983076534478404\\
0.25	0.983834301591311\\
    };
    \addlegendentry{\textcolor{black}{\kernelm: 0.081}}
    \addplot [color=cPLOT0, dotted, smooth, line width=\pckLineWidth]
    table[row sep=crcr]{%
0	0.027370417193426\\
0.00862068965517241	0.0308470290771176\\
0.0172413793103448	0.0924778761061947\\
0.0258620689655172	0.209102402022756\\
0.0344827586206897	0.319532237673831\\
0.0431034482758621	0.449178255372946\\
0.0517241379310345	0.572819216182048\\
0.0603448275862069	0.674778761061947\\
0.0689655172413793	0.762136536030341\\
0.0775862068965517	0.826864728192162\\
0.0862068965517241	0.872060682680152\\
0.0948275862068965	0.903223767383059\\
0.103448275862069	0.925284450063211\\
0.112068965517241	0.941150442477876\\
0.120689655172414	0.952907711757269\\
0.129310344827586	0.960113780025284\\
0.137931034482759	0.965613147914033\\
0.146551724137931	0.970796460176991\\
0.155172413793103	0.974209860935525\\
0.163793103448276	0.976991150442478\\
0.172413793103448	0.979393173198483\\
0.181034482758621	0.981542351453856\\
0.189655172413793	0.98299620733249\\
0.198275862068966	0.984323640960809\\
0.206896551724138	0.985461441213654\\
0.21551724137931	0.986472819216182\\
0.224137931034483	0.987041719342604\\
0.232758620689655	0.987800252844501\\
0.241379310344828	0.988495575221239\\
0.25	0.989190897597977\\
    };
    \addlegendentry{\textcolor{black}{\smcomb: 0.055}}
    \addplot [color=cPLOT3, dotted, smooth, line width=\pckLineWidth]
    table[row sep=crcr]{%
0	0.0280960927792351\\
0.00862068965517241	0.0318928620737264\\
0.0172413793103448	0.0949882645312716\\
0.0258620689655172	0.215863592434074\\
0.0344827586206897	0.331906668507524\\
0.0431034482758621	0.46589810851857\\
0.0517241379310345	0.593331492475494\\
0.0603448275862069	0.696396520778683\\
0.0689655172413793	0.784274471903907\\
0.0775862068965517	0.849993096783101\\
0.0862068965517241	0.894035620599199\\
0.0948275862068965	0.926480740024852\\
0.103448275862069	0.948640066270882\\
0.112068965517241	0.962101339224078\\
0.120689655172414	0.971696810713793\\
0.129310344827586	0.97790970592296\\
0.137931034482759	0.982189700400387\\
0.146551724137931	0.985779373187906\\
0.155172413793103	0.988747756454508\\
0.163793103448276	0.990680657186249\\
0.172413793103448	0.992268397073036\\
0.181034482758621	0.993856136959823\\
0.189655172413793	0.994891619494685\\
0.198275862068966	0.995858069860555\\
0.206896551724138	0.996479359381472\\
0.21551724137931	0.997031616733398\\
0.224137931034483	0.997514841916333\\
0.232758620689655	0.997790970592296\\
0.241379310344828	0.998067099268259\\
0.25	0.99813613143725\\
    };
    \addlegendentry{\textcolor{black}{\fastdog: \textbf{0.051}}}
    \addplot [color=cPLOT5, dashed, smooth, line width=\pckLineWidth]
    table[row sep=crcr]{%
0	0.0282316992161552\\
0.00862068965517241	0.0320844958150658\\
0.0172413793103448	0.0966520526106018\\
0.0258620689655172	0.217483725255746\\
0.0344827586206897	0.332801913112794\\
0.0431034482758621	0.465656968247642\\
0.0517241379310345	0.593131393649528\\
0.0603448275862069	0.697555466985519\\
0.0689655172413793	0.785040520791816\\
0.0775862068965517	0.850006642752757\\
0.0862068965517241	0.894380231167796\\
0.0948275862068965	0.926265444400159\\
0.103448275862069	0.948252956024977\\
0.112068965517241	0.961472034010894\\
0.120689655172414	0.970439750232496\\
0.129310344827586	0.976617510296267\\
0.137931034482759	0.980736017005447\\
0.146551724137931	0.984323103494088\\
0.155172413793103	0.98658163943138\\
0.163793103448276	0.988441610203268\\
0.172413793103448	0.990235153447589\\
0.181034482758621	0.991364421416235\\
0.189655172413793	0.992427261857314\\
0.198275862068966	0.993423674770825\\
0.206896551724138	0.993955094991364\\
0.21551724137931	0.994552942739471\\
0.224137931034483	0.994951507904876\\
0.232758620689655	0.995482928125415\\
0.241379310344828	0.996014348345955\\
0.25	0.996412913511359\\
    };
    \addlegendentry{\textcolor{black}{\ours: \textbf{0.051}}}
        
	\end{axis}
\node[above,font=\large, inner sep=1pt] at (current bounding box.north) {\textbf{DT4D-Inter}};
\end{tikzpicture}
         \vspace{-0.2em}
          \\%
        \hspace{-0.7cm}
         \newcommand{\pckLineWidth}{3pt}
\newcommand{\plotWidth}{0.55\columnwidth}
\newcommand{\plotHeight}{0.44\columnwidth}
\newcommand{\pckTitle}{}

\pgfplotsset{%
    label style = {font=\normalfont},
    tick label style = {font=\normalfont},
    title style =  {font=\large},
    legend style={  fill= gray!10,
                    fill opacity=0.6, 
                    font=\normalfont,
                    draw=gray!20, %
                    text opacity=1}
}
\begin{tikzpicture}[scale=0.5, transform shape]
	\begin{axis}[
		width=\plotWidth,
		height=\plotHeight,
		grid=major,
		title=\pckTitle,
		legend style={
			at={(0.97,0.03)},
			anchor=south east,
			legend columns=1},
		legend cell align={left},
        title style={yshift=-0.1cm},
        ylabel={{$\%$ of Triangles ($\rightarrow$)}},
        xlabel={Conformal Distortion Threshold},
        xmin=0,
        xmax=1,
        ylabel near ticks,
        xtick={0, 0.25, 0.5, 0.75, 1},
        ymin=0,
        ymax=0.65,
        ytick={0, 0.20, 0.40, 0.60, 0.65, 0.80},
        yticklabels={$0$, $20$, $40$, $60$, $\phantom{100}$, $80$, $100$},
	]
    \addplot [color=cPLOT2, smooth, line width=\pckLineWidth]
    table[row sep=crcr]{%
0	0\\
0.034375	0.0435273368606702\\
0.06875	0.0820693709582598\\
0.103125	0.117166372721928\\
0.1375	0.15015873015873\\
0.171875	0.180388007054674\\
0.20625	0.208042328042328\\
0.240625	0.23512051734274\\
0.275	0.259106407995297\\
0.309375	0.281622574955908\\
0.34375	0.303033509700176\\
0.378125	0.322657260435038\\
0.4125	0.340634920634921\\
0.446875	0.358118753674309\\
0.48125	0.374955908289242\\
0.515625	0.390605526161082\\
0.55	0.404103468547913\\
0.584375	0.418106995884774\\
0.61875	0.431440329218107\\
0.653125	0.442516166960611\\
0.6875	0.45358024691358\\
0.721875	0.463727219282775\\
0.75625	0.4736860670194\\
0.790625	0.482621987066432\\
0.825	0.491851851851852\\
0.859375	0.499623750734862\\
0.89375	0.507995296884186\\
0.928125	0.51564961787184\\
0.9625	0.52270429159318\\
0.996875	0.529641387419165\\
1.03125	0.536072898295121\\
    };
    \addlegendentry{\textcolor{black}{\caoetal}}
    \addplot [color=cPLOT4, smooth, line width=\pckLineWidth]
    table[row sep=crcr]{%
0	0\\
0.034375	0.0100350877192982\\
0.06875	0.0204561403508772\\
0.103125	0.030046783625731\\
0.1375	0.040093567251462\\
0.171875	0.0499766081871345\\
0.20625	0.0589707602339181\\
0.240625	0.0684561403508772\\
0.275	0.0774970760233918\\
0.309375	0.0867017543859649\\
0.34375	0.0961754385964912\\
0.378125	0.104760233918129\\
0.4125	0.113356725146199\\
0.446875	0.12187134502924\\
0.48125	0.13080701754386\\
0.515625	0.139730994152047\\
0.55	0.14859649122807\\
0.584375	0.157415204678363\\
0.61875	0.165824561403509\\
0.653125	0.17425730994152\\
0.6875	0.182397660818713\\
0.721875	0.190584795321637\\
0.75625	0.198643274853801\\
0.790625	0.206573099415205\\
0.825	0.214233918128655\\
0.859375	0.222374269005848\\
0.89375	0.230280701754386\\
0.928125	0.237707602339181\\
0.9625	0.245204678362573\\
0.996875	0.252175438596491\\
1.03125	0.259461988304094\\
    };
    \addlegendentry{\textcolor{black}{\kernelm}}
    \addplot [color=cPLOT0, smooth, line width=\pckLineWidth]
    table[row sep=crcr]{%
0	0\\
0.034375	0.0434269005847953\\
0.06875	0.0913918128654971\\
0.103125	0.134994152046784\\
0.1375	0.175473684210526\\
0.171875	0.211941520467836\\
0.20625	0.245953216374269\\
0.240625	0.278748538011696\\
0.275	0.30833918128655\\
0.309375	0.335122807017544\\
0.34375	0.360842105263158\\
0.378125	0.384573099415205\\
0.4125	0.405953216374269\\
0.446875	0.426444444444444\\
0.48125	0.445614035087719\\
0.515625	0.463426900584795\\
0.55	0.480760233918129\\
0.584375	0.496222222222222\\
0.61875	0.511181286549708\\
0.653125	0.524350877192982\\
0.6875	0.537719298245614\\
0.721875	0.548982456140351\\
0.75625	0.559415204678363\\
0.790625	0.569485380116959\\
0.825	0.578818713450292\\
0.859375	0.587684210526316\\
0.89375	0.59646783625731\\
0.928125	0.604584795321637\\
0.9625	0.611929824561404\\
0.996875	0.619146198830409\\
1.03125	0.625567251461988\\
1.065625	0.631707602339181\\
1.1	0.637391812865497\\
    };
    \addlegendentry{\textcolor{black}{\smcomb}}
    \addplot [color=cPLOT3, dotted, smooth, line width=\pckLineWidth]
    table[row sep=crcr]{%
0	0\\
0.034375	0.0508654970760234\\
0.06875	0.0979766081871345\\
0.103125	0.141356725146199\\
0.1375	0.181099415204678\\
0.171875	0.217017543859649\\
0.20625	0.250865497076023\\
0.240625	0.282818713450292\\
0.275	0.312350877192982\\
0.309375	0.33927485380117\\
0.34375	0.364362573099415\\
0.378125	0.38774269005848\\
0.4125	0.408900584795322\\
0.446875	0.429345029239766\\
0.48125	0.448210526315789\\
0.515625	0.466233918128655\\
0.55	0.482900584795322\\
0.584375	0.498421052631579\\
0.61875	0.513157894736842\\
0.653125	0.526269005847953\\
0.6875	0.539508771929825\\
0.721875	0.550631578947368\\
0.75625	0.56080701754386\\
0.790625	0.570912280701754\\
0.825	0.580023391812866\\
0.859375	0.588853801169591\\
0.89375	0.597614035087719\\
0.928125	0.605836257309942\\
0.9625	0.61306432748538\\
0.996875	0.620152046783626\\
1.03125	0.626374269005848\\
    };
    \addlegendentry{\textcolor{black}{\fastdog}}
    \addplot [color=cPLOT5, smooth, dashed, line width=\pckLineWidth]
    table[row sep=crcr]{%
0	0\\
0.034375	0.0501704879482657\\
0.06875	0.0966607877718989\\
0.103125	0.139753086419753\\
0.1375	0.179271017048795\\
0.171875	0.214944150499706\\
0.20625	0.248724279835391\\
0.240625	0.280411522633745\\
0.275	0.30962962962963\\
0.309375	0.336319811875367\\
0.34375	0.361552028218695\\
0.378125	0.38487948265726\\
0.4125	0.405690770135215\\
0.446875	0.425878894767784\\
0.48125	0.4445620223398\\
0.515625	0.462433862433862\\
0.55	0.478847736625514\\
0.584375	0.494074074074074\\
0.61875	0.508500881834215\\
0.653125	0.521540270429159\\
0.6875	0.534450323339212\\
0.721875	0.545549676660788\\
0.75625	0.555837742504409\\
0.790625	0.565867136978248\\
0.825	0.575061728395062\\
0.859375	0.583821281599059\\
0.89375	0.592451499118166\\
0.928125	0.600576131687243\\
0.9625	0.607466196355085\\
0.996875	0.614661963550852\\
1.03125	0.62082304526749\\
    };
    \addlegendentry{\textcolor{black}{\ours}}
        
	\end{axis}
\end{tikzpicture}& 
         \newcommand{\pckLineWidth}{3pt}
\newcommand{\plotWidth}{0.55\columnwidth}
\newcommand{\plotHeight}{0.44\columnwidth}
\newcommand{\pckTitle}{}

\pgfplotsset{%
    label style = {font=\normalfont},
    tick label style = {font=\normalfont},
    title style =  {font=\large},
    legend style={  fill= gray!10,
                    fill opacity=0.6, 
                    font=\normalfont,
                    draw=gray!20, %
                    text opacity=1}
}
\begin{tikzpicture}[scale=0.5, transform shape]
	\begin{axis}[
		width=\plotWidth,
		height=\plotHeight,
		grid=major,
		title=\pckTitle,
		legend style={
			at={(0.97,0.03)},
			anchor=south east,
			legend columns=1},
		legend cell align={left},
        title style={yshift=-0.1cm},
        xlabel={Conformal Distortion Threshold},
        xmin=0,
        xmax=1,
        ylabel near ticks,
        xtick={0, 0.25, 0.5, 0.75, 1},
        ymin=0,
        ymax=0.65,
        ytick={0, 0.20, 0.40, 0.60, 0.65, 0.80},
        yticklabels={$0$, $20$, $40$, $60$, $\phantom{100}$, $80$, $100$},
	]
    \addplot [color=cPLOT2, smooth, line width=\pckLineWidth]
    table[row sep=crcr]{%
0	0\\
0.034375	0.0368627450980392\\
0.06875	0.0725816993464052\\
0.103125	0.104607843137255\\
0.1375	0.135816993464052\\
0.171875	0.164722222222222\\
0.20625	0.189362745098039\\
0.240625	0.213104575163399\\
0.275	0.236732026143791\\
0.309375	0.257941176470588\\
0.34375	0.278807189542484\\
0.378125	0.297696078431373\\
0.4125	0.31640522875817\\
0.446875	0.332761437908497\\
0.48125	0.348954248366013\\
0.515625	0.36468954248366\\
0.55	0.379575163398693\\
0.584375	0.392843137254902\\
0.61875	0.405408496732026\\
0.653125	0.417908496732026\\
0.6875	0.42983660130719\\
0.721875	0.441437908496732\\
0.75625	0.451552287581699\\
0.790625	0.461437908496732\\
0.825	0.470915032679739\\
0.859375	0.479575163398693\\
0.89375	0.487614379084967\\
0.928125	0.495441176470588\\
0.9625	0.502859477124183\\
0.996875	0.51016339869281\\
1.03125	0.517222222222222\\
    };
    \addlegendentry{\textcolor{black}{\caoetal}}
    \addplot [color=cPLOT4, smooth, line width=\pckLineWidth]
    table[row sep=crcr]{%
0	0\\
0.034375	0.0142647058823529\\
0.06875	0.0270098039215686\\
0.103125	0.0400816993464052\\
0.1375	0.0534313725490196\\
0.171875	0.0661437908496732\\
0.20625	0.0793137254901961\\
0.240625	0.091421568627451\\
0.275	0.103774509803922\\
0.309375	0.115849673202614\\
0.34375	0.128611111111111\\
0.378125	0.139852941176471\\
0.4125	0.151699346405229\\
0.446875	0.163316993464052\\
0.48125	0.173986928104575\\
0.515625	0.18578431372549\\
0.55	0.196454248366013\\
0.584375	0.207173202614379\\
0.61875	0.218415032679739\\
0.653125	0.228545751633987\\
0.6875	0.238137254901961\\
0.721875	0.248218954248366\\
0.75625	0.258333333333333\\
0.790625	0.267924836601307\\
0.825	0.277712418300654\\
0.859375	0.286388888888889\\
0.89375	0.295588235294118\\
0.928125	0.304493464052288\\
0.9625	0.313235294117647\\
0.996875	0.32202614379085\\
1.03125	0.331029411764706\\
    };
    \addlegendentry{\textcolor{black}{\kernelm}}
    \addplot [color=cPLOT0, smooth, line width=\pckLineWidth]
    table[row sep=crcr]{%
0	0\\
0.034375	0.0345363408521303\\
0.06875	0.0732832080200501\\
0.103125	0.109874686716792\\
0.1375	0.144962406015038\\
0.171875	0.175538847117794\\
0.20625	0.204678362573099\\
0.240625	0.231010860484545\\
0.275	0.256808688387636\\
0.309375	0.280100250626566\\
0.34375	0.301203007518797\\
0.378125	0.32078529657477\\
0.4125	0.340334168755221\\
0.446875	0.359415204678363\\
0.48125	0.376357560568087\\
0.515625	0.392715121136174\\
0.55	0.408203842940685\\
0.584375	0.422155388471178\\
0.61875	0.435355054302423\\
0.653125	0.447385129490393\\
0.6875	0.458446115288221\\
0.721875	0.470108604845447\\
0.75625	0.480768588137009\\
0.790625	0.490325814536341\\
0.825	0.499933166248956\\
0.859375	0.508654970760234\\
0.89375	0.516758563074353\\
0.928125	0.523859649122807\\
0.9625	0.530893901420217\\
0.996875	0.538345864661654\\
1.03125	0.545129490392648\\
    };
    \addlegendentry{\textcolor{black}{\smcomb}}
    \addplot [color=cPLOT3, dotted, smooth, line width=\pckLineWidth]
    table[row sep=crcr]{%
0	0\\
0.034375	0.0405346700083542\\
0.06875	0.0787802840434419\\
0.103125	0.114903926482874\\
0.1375	0.149390142021721\\
0.171875	0.180534670008354\\
0.20625	0.208487886382623\\
0.240625	0.234853801169591\\
0.275	0.260066833751044\\
0.309375	0.283157894736842\\
0.34375	0.304043441938179\\
0.378125	0.323375104427736\\
0.4125	0.343040935672515\\
0.446875	0.361904761904762\\
0.48125	0.378496240601504\\
0.515625	0.394937343358396\\
0.55	0.410375939849624\\
0.584375	0.424260651629073\\
0.61875	0.437159565580618\\
0.653125	0.449005847953216\\
0.6875	0.460334168755221\\
0.721875	0.471829573934837\\
0.75625	0.482289055973266\\
0.790625	0.491812865497076\\
0.825	0.501136173767753\\
0.859375	0.51015873015873\\
0.89375	0.517827903091061\\
0.928125	0.524928989139515\\
0.9625	0.531829573934837\\
0.996875	0.539431913116124\\
1.03125	0.54593149540518\\
    };
    \addlegendentry{\textcolor{black}{\fastdog}}
    \addplot [color=cPLOT5, smooth, dashed, line width=\pckLineWidth]
    table[row sep=crcr]{%
0	0\\
0.034375	0.040702614379085\\
0.06875	0.0790686274509804\\
0.103125	0.115735294117647\\
0.1375	0.149852941176471\\
0.171875	0.180522875816993\\
0.20625	0.208496732026144\\
0.240625	0.235098039215686\\
0.275	0.26078431372549\\
0.309375	0.283741830065359\\
0.34375	0.304509803921569\\
0.378125	0.324264705882353\\
0.4125	0.343954248366013\\
0.446875	0.363055555555556\\
0.48125	0.379885620915033\\
0.515625	0.396699346405229\\
0.55	0.412173202614379\\
0.584375	0.425522875816993\\
0.61875	0.438235294117647\\
0.653125	0.450179738562091\\
0.6875	0.461372549019608\\
0.721875	0.472679738562092\\
0.75625	0.483415032679739\\
0.790625	0.493088235294118\\
0.825	0.502222222222222\\
0.859375	0.510996732026144\\
0.89375	0.519052287581699\\
0.928125	0.526372549019608\\
0.9625	0.533186274509804\\
0.996875	0.540866013071895\\
1.03125	0.547369281045752\\
    };
    \addlegendentry{\textcolor{black}{\ours}}
        
	\end{axis}
\end{tikzpicture}&
         \newcommand{\pckLineWidth}{3pt}
\newcommand{\plotWidth}{0.55\columnwidth}
\newcommand{\plotHeight}{0.44\columnwidth}
\newcommand{\pckTitle}{}

\pgfplotsset{%
    label style = {font=\normalfont},
    tick label style = {font=\normalfont},
    title style =  {font=\large},
    legend style={  fill= gray!10,
                    fill opacity=0.6, 
                    font=\normalfont,
                    draw=gray!20, %
                    text opacity=1}
}
\begin{tikzpicture}[scale=0.5, transform shape]
	\begin{axis}[
		width=\plotWidth,
		height=\plotHeight,
		grid=major,
		title=\pckTitle,
		legend style={
			at={(0.97,0.03)},
			anchor=south east,
			legend columns=1},
		legend cell align={left},
        title style={yshift=-0.1cm},
        xlabel={Conformal Distortion Threshold},
        xmin=0,
        xmax=1,
        ylabel near ticks,
        xtick={0, 0.25, 0.5, 0.75, 1},
        ymin=0,
        ymax=0.65,
        ytick={0, 0.20, 0.40, 0.60, 0.65, 0.80},
        yticklabels={$0$, $20$, $40$, $60$, $\phantom{100}$, $80$, $100$},
	]
    \addplot [color=cPLOT2, smooth, line width=\pckLineWidth]
    table[row sep=crcr]{%
0	0\\
0.034375	0.0387111111111111\\
0.06875	0.0728\\
0.103125	0.106266666666667\\
0.1375	0.135955555555556\\
0.171875	0.163\\
0.20625	0.192088888888889\\
0.240625	0.215822222222222\\
0.275	0.237355555555556\\
0.309375	0.257311111111111\\
0.34375	0.276288888888889\\
0.378125	0.294688888888889\\
0.4125	0.311622222222222\\
0.446875	0.327066666666667\\
0.48125	0.341688888888889\\
0.515625	0.355288888888889\\
0.55	0.368933333333333\\
0.584375	0.3822\\
0.61875	0.394\\
0.653125	0.406311111111111\\
0.6875	0.416244444444444\\
0.721875	0.425911111111111\\
0.75625	0.4354\\
0.790625	0.444622222222222\\
0.825	0.453066666666667\\
0.859375	0.460777777777778\\
0.89375	0.468888888888889\\
0.928125	0.475422222222222\\
0.9625	0.4816\\
0.996875	0.4888\\
1.03125	0.495733333333333\\
    };
    \addlegendentry{\textcolor{black}{\caoetal}}
    \addplot [color=cPLOT4, smooth, line width=\pckLineWidth]
    table[row sep=crcr]{%
0	0\\
0.034375	0.0153111111111111\\
0.06875	0.0291777777777778\\
0.103125	0.0441333333333333\\
0.1375	0.0577777777777778\\
0.171875	0.0716444444444444\\
0.20625	0.0849555555555556\\
0.240625	0.0990444444444444\\
0.275	0.112688888888889\\
0.309375	0.126222222222222\\
0.34375	0.138422222222222\\
0.378125	0.1508\\
0.4125	0.162777777777778\\
0.446875	0.174177777777778\\
0.48125	0.186444444444444\\
0.515625	0.198533333333333\\
0.55	0.209088888888889\\
0.584375	0.2202\\
0.61875	0.231577777777778\\
0.653125	0.241733333333333\\
0.6875	0.252155555555556\\
0.721875	0.2626\\
0.75625	0.273333333333333\\
0.790625	0.283222222222222\\
0.825	0.292688888888889\\
0.859375	0.3022\\
0.89375	0.312111111111111\\
0.928125	0.321777777777778\\
0.9625	0.330311111111111\\
0.996875	0.340177777777778\\
1.03125	0.3482\\
    };
    \addlegendentry{\textcolor{black}{\kernelm}}
    \addplot [color=cPLOT0, smooth, line width=\pckLineWidth]
    table[row sep=crcr]{%
0	0\\
0.034375	0.0429111111111111\\
0.06875	0.0919333333333333\\
0.103125	0.135844444444444\\
0.1375	0.176711111111111\\
0.171875	0.215577777777778\\
0.20625	0.251222222222222\\
0.240625	0.2846\\
0.275	0.314066666666667\\
0.309375	0.342466666666667\\
0.34375	0.367666666666667\\
0.378125	0.391266666666667\\
0.4125	0.4132\\
0.446875	0.432866666666667\\
0.48125	0.450555555555556\\
0.515625	0.467155555555556\\
0.55	0.483355555555556\\
0.584375	0.498844444444444\\
0.61875	0.512955555555556\\
0.653125	0.526688888888889\\
0.6875	0.538466666666667\\
0.721875	0.549933333333333\\
0.75625	0.561311111111111\\
0.790625	0.5706\\
0.825	0.580088888888889\\
0.859375	0.589133333333333\\
0.89375	0.598222222222222\\
0.928125	0.605444444444444\\
0.9625	0.611933333333333\\
0.996875	0.618555555555556\\
1.03125	0.625288888888889\\
    };
    \addlegendentry{\textcolor{black}{\smcomb}}
    \addplot [color=cPLOT3, dotted, smooth, line width=\pckLineWidth]
    table[row sep=crcr]{%
0	0\\
0.034375	0.0500888888888889\\
0.06875	0.0984666666666667\\
0.103125	0.142177777777778\\
0.1375	0.182933333333333\\
0.171875	0.220666666666667\\
0.20625	0.255822222222222\\
0.240625	0.289111111111111\\
0.275	0.318533333333333\\
0.309375	0.346444444444444\\
0.34375	0.370977777777778\\
0.378125	0.394422222222222\\
0.4125	0.416266666666667\\
0.446875	0.435533333333333\\
0.48125	0.453311111111111\\
0.515625	0.4694\\
0.55	0.486177777777778\\
0.584375	0.501111111111111\\
0.61875	0.514844444444444\\
0.653125	0.528511111111111\\
0.6875	0.540088888888889\\
0.721875	0.551688888888889\\
0.75625	0.562733333333333\\
0.790625	0.572133333333333\\
0.825	0.581666666666667\\
0.859375	0.590755555555556\\
0.89375	0.599577777777778\\
0.928125	0.606355555555556\\
0.9625	0.6128\\
0.996875	0.619733333333333\\
1.03125	0.626155555555556\\
    };
    \addlegendentry{\textcolor{black}{\fastdog}}
    \addplot [color=cPLOT5, smooth, dashed, line width=\pckLineWidth]
    table[row sep=crcr]{%
0	0\\
0.034375	0.0499555555555556\\
0.06875	0.0977111111111111\\
0.103125	0.141266666666667\\
0.1375	0.181844444444444\\
0.171875	0.219511111111111\\
0.20625	0.254533333333333\\
0.240625	0.287755555555556\\
0.275	0.317244444444444\\
0.309375	0.344933333333333\\
0.34375	0.369288888888889\\
0.378125	0.392888888888889\\
0.4125	0.414711111111111\\
0.446875	0.433755555555556\\
0.48125	0.451488888888889\\
0.515625	0.467733333333333\\
0.55	0.484555555555556\\
0.584375	0.499466666666667\\
0.61875	0.513333333333333\\
0.653125	0.526733333333333\\
0.6875	0.538444444444444\\
0.721875	0.550022222222222\\
0.75625	0.561088888888889\\
0.790625	0.570533333333333\\
0.825	0.579955555555556\\
0.859375	0.589022222222222\\
0.89375	0.597866666666667\\
0.928125	0.604622222222222\\
0.9625	0.611044444444444\\
0.996875	0.617933333333333\\
1.03125	0.624466666666667\\
    };
    \addlegendentry{\textcolor{black}{\ours}}
        
	\end{axis}
\end{tikzpicture}& 
         \newcommand{\pckLineWidth}{3pt}
\newcommand{\plotWidth}{0.55\columnwidth}
\newcommand{\plotHeight}{0.44\columnwidth}
\newcommand{\pckTitle}{}

\pgfplotsset{%
    label style = {font=\normalfont},
    tick label style = {font=\normalfont},
    title style =  {font=\large},
    legend style={  fill= gray!10,
                    fill opacity=0.6, 
                    font=\normalfont,
                    draw=gray!20, %
                    text opacity=1}
}
\begin{tikzpicture}[scale=0.5, transform shape]
	\begin{axis}[
		width=\plotWidth,
		height=\plotHeight,
		grid=major,
		title=\pckTitle,
		legend style={
			at={(0.97,0.03)},
			anchor=south east,
			legend columns=1},
		legend cell align={left},
        title style={yshift=-0.1cm},
        xlabel={Conformal Distortion Threshold},
        xmin=0,
        xmax=1,
        ylabel near ticks,
        xtick={0, 0.25, 0.5, 0.75, 1},
        ymin=0,
        ymax=0.65,
        ytick={0, 0.20, 0.40, 0.60, 0.65, 0.80},
        yticklabels={$0$, $20$, $40$, $60$, $\phantom{100}$, $80$, $100$},
	]
    \addplot [color=cPLOT2, smooth, line width=\pckLineWidth]
    table[row sep=crcr]{%
0	0\\
0.034375	0.0456704980842912\\
0.06875	0.086360153256705\\
0.103125	0.123409961685824\\
0.1375	0.158927203065134\\
0.171875	0.189272030651341\\
0.20625	0.219463601532567\\
0.240625	0.247931034482759\\
0.275	0.273256704980843\\
0.309375	0.296704980842912\\
0.34375	0.318084291187739\\
0.378125	0.338927203065134\\
0.4125	0.358199233716475\\
0.446875	0.376590038314176\\
0.48125	0.393716475095785\\
0.515625	0.409616858237548\\
0.55	0.423946360153257\\
0.584375	0.43904214559387\\
0.61875	0.453908045977012\\
0.653125	0.466245210727969\\
0.6875	0.477662835249042\\
0.721875	0.488544061302682\\
0.75625	0.497624521072797\\
0.790625	0.507892720306513\\
0.825	0.516704980842912\\
0.859375	0.525402298850575\\
0.89375	0.533448275862069\\
0.928125	0.540689655172414\\
0.9625	0.548122605363985\\
0.996875	0.555134099616858\\
1.03125	0.561532567049808\\
    };
    \addlegendentry{\textcolor{black}{\caoetal}}
    \addplot [color=cPLOT4, smooth, line width=\pckLineWidth]
    table[row sep=crcr]{%
0	0\\
0.034375	0.0140444444444444\\
0.06875	0.0276666666666667\\
0.103125	0.0421555555555556\\
0.1375	0.0551333333333333\\
0.171875	0.0693333333333333\\
0.20625	0.0820444444444444\\
0.240625	0.0952444444444445\\
0.275	0.107377777777778\\
0.309375	0.120488888888889\\
0.34375	0.133155555555556\\
0.378125	0.144888888888889\\
0.4125	0.156755555555556\\
0.446875	0.168511111111111\\
0.48125	0.179933333333333\\
0.515625	0.1918\\
0.55	0.202955555555556\\
0.584375	0.213177777777778\\
0.61875	0.223111111111111\\
0.653125	0.232822222222222\\
0.6875	0.242888888888889\\
0.721875	0.252888888888889\\
0.75625	0.2624\\
0.790625	0.271822222222222\\
0.825	0.281888888888889\\
0.859375	0.290933333333333\\
0.89375	0.3008\\
0.928125	0.310755555555556\\
0.9625	0.3198\\
0.996875	0.328844444444444\\
1.03125	0.337511111111111\\
    };
    \addlegendentry{\textcolor{black}{\kernelm}}
    \addplot [color=cPLOT0, smooth, line width=\pckLineWidth]
    table[row sep=crcr]{%
0	0\\
0.034375	0.0393686868686869\\
0.06875	0.0833585858585859\\
0.103125	0.122676767676768\\
0.1375	0.157424242424242\\
0.171875	0.19\\
0.20625	0.221212121212121\\
0.240625	0.250934343434343\\
0.275	0.278914141414141\\
0.309375	0.304015151515152\\
0.34375	0.327121212121212\\
0.378125	0.35040404040404\\
0.4125	0.370530303030303\\
0.446875	0.39020202020202\\
0.48125	0.408686868686869\\
0.515625	0.42510101010101\\
0.55	0.439545454545455\\
0.584375	0.453636363636364\\
0.61875	0.467651515151515\\
0.653125	0.480176767676768\\
0.6875	0.491994949494949\\
0.721875	0.503232323232323\\
0.75625	0.513459595959596\\
0.790625	0.523333333333333\\
0.825	0.532323232323232\\
0.859375	0.541010101010101\\
0.89375	0.548611111111111\\
0.928125	0.555934343434343\\
0.9625	0.563106060606061\\
0.996875	0.570681818181818\\
1.03125	0.57719696969697\\
    };
    \addlegendentry{\textcolor{black}{\smcomb}}
    \addplot [color=cPLOT3, dotted, smooth, line width=\pckLineWidth]
    table[row sep=crcr]{%
0	0\\
0.034375	0.0457070707070707\\
0.06875	0.0887878787878788\\
0.103125	0.128005050505051\\
0.1375	0.161691919191919\\
0.171875	0.194823232323232\\
0.20625	0.225984848484848\\
0.240625	0.255479797979798\\
0.275	0.282575757575758\\
0.309375	0.307954545454545\\
0.34375	0.330227272727273\\
0.378125	0.352929292929293\\
0.4125	0.373813131313131\\
0.446875	0.393080808080808\\
0.48125	0.411161616161616\\
0.515625	0.427550505050505\\
0.55	0.442020202020202\\
0.584375	0.455454545454545\\
0.61875	0.469242424242424\\
0.653125	0.48219696969697\\
0.6875	0.493611111111111\\
0.721875	0.504646464646465\\
0.75625	0.514974747474747\\
0.790625	0.524520202020202\\
0.825	0.533611111111111\\
0.859375	0.542222222222222\\
0.89375	0.549873737373737\\
0.928125	0.556843434343434\\
0.9625	0.564217171717172\\
0.996875	0.571666666666667\\
1.03125	0.578131313131313\\
    };
    \addlegendentry{\textcolor{black}{\fastdog}}
    \addplot [color=cPLOT5, smooth, dashed, line width=\pckLineWidth]
    table[row sep=crcr]{%
0	0\\
0.034375	0.0470818070818071\\
0.06875	0.0899145299145299\\
0.103125	0.13003663003663\\
0.1375	0.163614163614164\\
0.171875	0.196678876678877\\
0.20625	0.228278388278388\\
0.240625	0.258632478632479\\
0.275	0.286251526251526\\
0.309375	0.311355311355311\\
0.34375	0.334065934065934\\
0.378125	0.357289377289377\\
0.4125	0.377948717948718\\
0.446875	0.397240537240537\\
0.48125	0.414652014652015\\
0.515625	0.43030525030525\\
0.55	0.444639804639805\\
0.584375	0.458363858363858\\
0.61875	0.471892551892552\\
0.653125	0.484444444444444\\
0.6875	0.495824175824176\\
0.721875	0.506764346764347\\
0.75625	0.517460317460318\\
0.790625	0.526837606837607\\
0.825	0.535555555555556\\
0.859375	0.544200244200244\\
0.89375	0.552112332112332\\
0.928125	0.558925518925519\\
0.9625	0.565982905982906\\
0.996875	0.573040293040293\\
1.03125	0.57978021978022\\
    };
    \addlegendentry{\textcolor{black}{\ours}}
        
	\end{axis}
\end{tikzpicture}
         \vspace{0.2em}
         \\
         \hspace{-0.4cm}
         \pgfplotsset{%
    label style = {font=\normalfont},
    tick label style = {font=\normalfont},
    title style =  {font=\large,yshift=-0.1cm},
    legend style={  fill= gray!10,
                    fill opacity=0.6, 
                    font=\normalfont,
                    draw=gray!20, %
                    text opacity=1}
}
\begin{tikzpicture}[scale=0.5, transform shape]
    \begin{axis}[
		width=0.55\columnwidth,
		height=0.44\columnwidth,
        xticklabel style={rotate=45},
		title={},
		boxplot/draw direction=y,
		grid=major,
		ymin=-5,
		ymax=500,
		xmin=0.85,
		xmax=5.15,
            ymajorgrids=true,
		xtick={1,2,3,4,5,6},
		xticklabels={\caoetal, \kernelm, \smcomb, \fastdog, \ours, $\phantom{1}$},
		ylabel={Runtime [s] ($\leftarrow$)},
		every boxplot/.style={mark=x,every mark/.append style={mark size=5pt}},
		boxplotcolor/.style={color=#1,very thick,fill=#1!50,mark options={color=#1,fill=#1!70}},
            boxplot/box extend=0.4%
		]
		\addplot+[boxplotcolor=cPLOT2, boxplot prepared={
			lower whisker=3.4018,
			lower quartile=4.1115,
			median=4.3764,
			upper quartile=4.3764,
			upper whisker=6.5097
			}]
		coordinates {};%
            \addplot+[boxplotcolor=cPLOT4, boxplot prepared={
			lower whisker=0.0418,
			lower quartile=0.0464,
			median=0.043,
			upper quartile=0.046,
			upper whisker=0.1
			}]
		coordinates {};%
		\addplot+[boxplotcolor=cPLOT0, boxplot prepared={
			lower whisker=201.1398,
			lower quartile=262.6728,
			median=343.7218,
			upper quartile=343.7218,
			upper whisker=1118.0387
			}]
		coordinates {};
		\addplot+[boxplotcolor=cPLOT3, boxplot prepared={
			lower whisker=48.6354,
			lower quartile=109.6462,
			median=177.7282,
			upper quartile=177.7282,
			upper whisker=15896.477
			}]
		coordinates {};
		\addplot+[boxplotcolor=cPLOT5, boxplot prepared={
			lower whisker=25.6313,
			lower quartile=38.6686,
			median=50.9252,
			upper quartile=50.9252,
			upper whisker=25913.9883
			}]
		coordinates {};

	\end{axis}
\end{tikzpicture}&
         \hspace{0.12cm}
         \pgfplotsset{%
    label style = {font=\normalfont},
    tick label style = {font=\normalfont},
    title style =  {font=\large,yshift=-0.1cm},
    legend style={  fill= gray!10,
                    fill opacity=0.6, 
                    font=\normalfont,
                    draw=gray!20, %
                    text opacity=1}
}
\begin{tikzpicture}[scale=0.5, transform shape]
    \begin{axis}[
		width=0.55\columnwidth,
		height=0.44\columnwidth,
		title={},
        xticklabel style={rotate=45},
		boxplot/draw direction=y,
		grid=major,
		ymin=-5,
		ymax=500,
		xmin=0.85,
		xmax=5.15,
            ymajorgrids=true,
		xtick={1,2,3,4,5,6},
		xticklabels={\caoetal, \kernelm, \smcomb, \fastdog, \ours, $\phantom{1}$},
		every boxplot/.style={mark=x,every mark/.append style={mark size=5pt}},
		boxplotcolor/.style={color=#1,very thick,fill=#1!50,mark options={color=#1,fill=#1!70}},
            boxplot/box extend=0.4%
		]
		\addplot+[boxplotcolor=cPLOT2, boxplot prepared={
			lower whisker=45.9591,
			lower quartile=50.1603,
			median=65.6395,
			upper quartile=65.6395,
			upper whisker=107.349,
			}]
		coordinates {};%
              \addplot+[boxplotcolor=cPLOT4, boxplot prepared={
			lower whisker=0.04,
			lower quartile=0.042,
			median=0.043,
			upper quartile=0.043,
			upper whisker=0.11
			}]
		coordinates {};%
		\addplot+[boxplotcolor=cPLOT0, boxplot prepared={
			lower whisker=213.7211,
			lower quartile=349.7185,
			median=440.5502,
			upper quartile=440.5502,
			upper whisker=1588.4532,
			}]
		coordinates {};
		\addplot+[boxplotcolor=cPLOT3, boxplot prepared={
			lower whisker=85.2238,
			lower quartile=195.9231,
			median=285.7218,
			upper quartile=285.7218,
			upper whisker=1869.0995,
			}]
		coordinates {};
		\addplot+[boxplotcolor=cPLOT5, boxplot prepared={
			lower whisker=51.3797,
			lower quartile=83.5229,
			median=121.3219,
			upper quartile=121.3219,
			upper whisker=3719.0378,
			}]
		coordinates {};
	\end{axis}
\end{tikzpicture}&
         \hspace{0.12cm}
         \pgfplotsset{%
    label style = {font=\normalfont},
    tick label style = {font=\normalfont},
    title style =  {font=\large,yshift=-0.1cm},
    legend style={  fill= gray!10,
                    fill opacity=0.6, 
                    font=\normalfont,
                    draw=gray!20, %
                    text opacity=1}
}
\begin{tikzpicture}[scale=0.5, transform shape]
    \begin{axis}[
		width=0.55\columnwidth,
		height=0.44\columnwidth,
		title={},
        xticklabel style={rotate=45},
		boxplot/draw direction=y,
		grid=major,
		ymin=-5,
		ymax=500,
		xmin=0.85,
		xmax=5.15,
            ymajorgrids=true,
		xtick={1,2,3,4,5,6},
		xticklabels={\caoetal, \kernelm, \smcomb, \fastdog, \ours, $\phantom{1}$},
		every boxplot/.style={mark=x,every mark/.append style={mark size=5pt}},
		boxplotcolor/.style={color=#1,very thick,fill=#1!50,mark options={color=#1,fill=#1!70}},
            boxplot/box extend=0.4%
		]
		\addplot+[boxplotcolor=cPLOT2, boxplot prepared={
			lower whisker=2.8842,
			lower quartile=2.9532,
			median=2.9964,
			upper quartile=2.9964,
			upper whisker=3.4098
			}]
		coordinates {};%
              \addplot+[boxplotcolor=cPLOT4, boxplot prepared={
			lower whisker=0.053,
			lower quartile=0.23,
			median=0.35,
			upper quartile=0.35,
			upper whisker=0.65
			}]
		coordinates {};%
		\addplot+[boxplotcolor=cPLOT0, boxplot prepared={
			lower whisker=229.6037,
			lower quartile=328.6241,
			median=415.8495,
			upper quartile=415.8495,
			upper whisker=1179.2788
			}]
		coordinates {};
		\addplot+[boxplotcolor=cPLOT3, boxplot prepared={
			lower whisker=37.9024,
			lower quartile=83.68,
			median=119.8609,
			upper quartile=119.8609,
			upper whisker=15609.0992
			}]
		coordinates {};
		\addplot+[boxplotcolor=cPLOT5, boxplot prepared={
			lower whisker=23.6381,
			lower quartile=34.7832,
			median=47.3197,
			upper quartile=47.3197,
			upper whisker=448.7586
			}]
		coordinates {};
	\end{axis}
\end{tikzpicture}&
         \hspace{0.12cm}
         \pgfplotsset{%
    label style = {font=\normalfont},
    tick label style = {font=\normalfont},
    title style =  {font=\large,yshift=-0.1cm},
    legend style={  fill= gray!10,
                    fill opacity=0.6, 
                    font=\normalfont,
                    draw=gray!20, %
                    text opacity=1}
}
\begin{tikzpicture}[scale=0.5, transform shape]
    \begin{axis}[
		width=0.55\columnwidth,
		height=0.44\columnwidth,
		title={},
        xticklabel style={rotate=45},
		boxplot/draw direction=y,
		grid=major,
		ymin=-5,
		ymax=500,
		xmin=0.85,
		xmax=5.15,
            ymajorgrids=true,
		xtick={1,2,3,4,5,6},
		xticklabels={\caoetal, \kernelm, \smcomb, \fastdog, \ours, $\phantom{1}$},
		every boxplot/.style={mark=x,every mark/.append style={mark size=5pt}},
		boxplotcolor/.style={color=#1,very thick,fill=#1!50,mark options={color=#1,fill=#1!70}},
            boxplot/box extend=0.4%
		]
		\addplot+[boxplotcolor=cPLOT2, boxplot prepared={
			lower whisker=25.3918,
			lower quartile=25.6928,
			median=25.8189,
			upper quartile=25.8189,
			upper whisker=26.8532
			}]
		coordinates {};%
              \addplot+[boxplotcolor=cPLOT4, boxplot prepared={
			lower whisker=0.039,
			lower quartile=0.042,
			median=0.043,
			upper quartile=0.043,
			upper whisker=0.092
			}]
		coordinates {};%
		\addplot+[boxplotcolor=cPLOT0, boxplot prepared={
			lower whisker=171.4311,
			lower quartile=230.4345,
			median=331.6253,
			upper quartile=331.6253,
			upper whisker=1391.6865
			}]
		coordinates {};
		\addplot+[boxplotcolor=cPLOT3, boxplot prepared={
			lower whisker=74.8833,
			lower quartile=140.8436,
			median=195.4911,
			upper quartile=195.4911,
			upper whisker=1276.0366
			}]
		coordinates {};
		\addplot+[boxplotcolor=cPLOT5, boxplot prepared={
			lower whisker=44.4505,
			lower quartile=64.1906,
			median=76.8724,
			upper quartile=76.8724,
			upper whisker=1083.6199
			}]
		coordinates {};
	\end{axis}
\end{tikzpicture}%
    \end{tabular}%
    \vspace{-0.25cm}
    \caption{\textit{Top:} Percentage of correct keypoints w.r.t.\ geodesic error thresholds on FAUST, SMAL, DT4D-Intra and DT4D-Inter datasets. 
    The numbers in the legends are mean geodesic errors ($\downarrow$). 
    \textit{Middle:} Conformal distortion errors on FAUST, SMAL, DT4D-Intra and DT4D-Inter datasets. 
    \textit{Bottom:} Statistics of runtime of all methods. 
    }
    \label{fig:pck}
\end{figure*}

\begin{figure}[!htb]
    \hspace{-1.5cm}
    \def\pathOurs{figs/qualitative/ours/}
\def\pathCao{figs/qualitative/caoetal/}
\def\srcEnd{_M.png}
\def\trgtEnd{_N.png}
\def\columnOne{Standing2HMagicAttack01043-GoalkeeperScoop046}
\def\columnTwo{DancingRunningMan285-StandingReactLargeFromLeft021}
\def\columnTwoTwo{Standing2HMagicAttack01081-ShortLeftSideStep006}
\def\columnThree{Floating029-BrooklynUprock146}
\def\columnFour{Shuffling092-Shuffling152}
\def\columnFive{camel_a-rhino}
\def\columnSix{dog-giraffe_b}
\def\columnSeven{horse_01-lion_05}
\def\columnEight{dog_01-cow_03}
\def\columnNine{tr_reg_083-tr_reg_094}
\def\columnTen{tr_reg_081-tr_reg_095}
\def\columnEleven{tr_reg_081-tr_reg_091}
\def\heightQ{1.55cm}
\def\widthQ{1.6cm}
\def\hspaceCols{-0.35cm}
\begin{tabular}{cccccccccccc}
    &%
    \scircled{1} &\hspace{\hspaceCols}  \scircled{2} &\hspace{\hspaceCols}  \scircled{3} &\hspace{\hspaceCols}  \scircled{4} &\hspace{\hspaceCols}  \scircled{5} &\hspace{\hspaceCols}  \scircled{6} &\hspace{\hspaceCols}  \scircled{7} &\hspace{\hspaceCols}  \scircled{8} &\hspace{\hspaceCols}  \scircled{9} &\hspace{\hspaceCols}  \scircled{10} &\hspace{\hspaceCols}\scircled{11}\\
    \setlength{\tabcolsep}{0pt} 
    \rotatedCentering{90}{\heightQ}{Source}&
    \includegraphics[height=\heightQ, width=\widthQ]{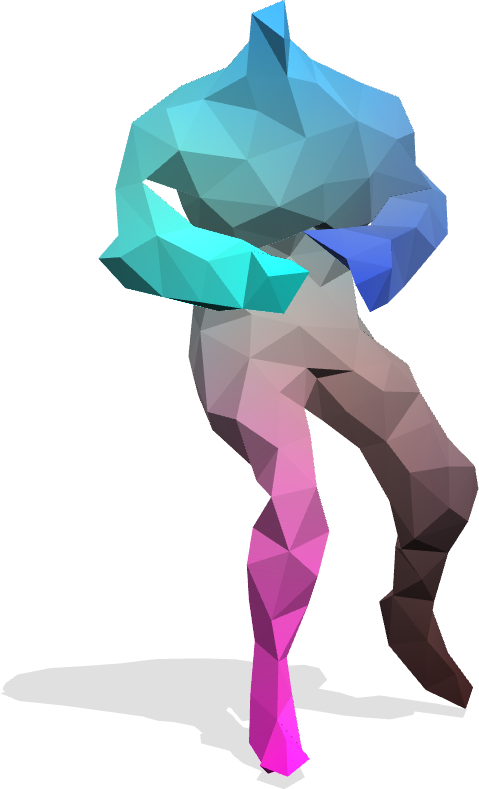}&
    \hspace{\hspaceCols}
    \includegraphics[height=\heightQ, width=\widthQ]{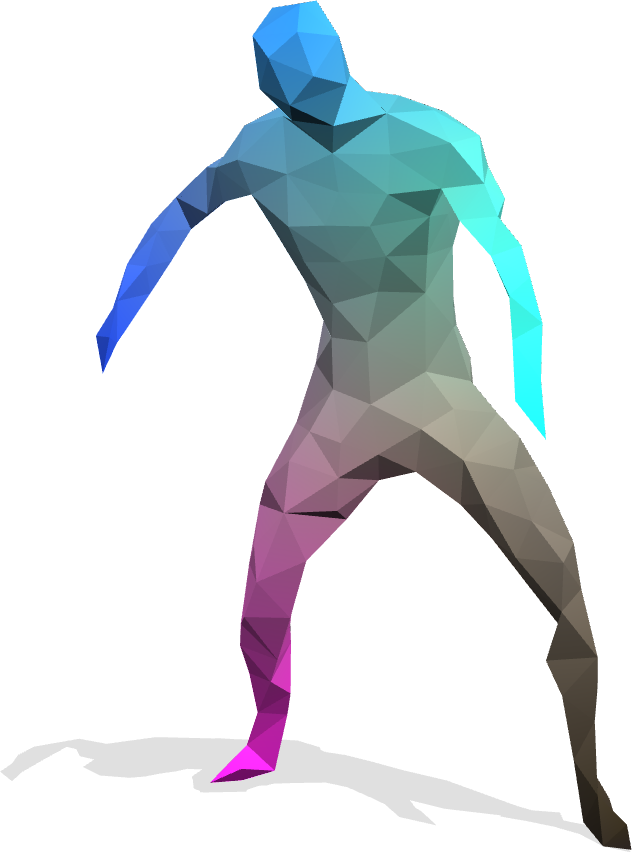}&
    \hspace{\hspaceCols}
    \includegraphics[height=\heightQ, width=\widthQ]{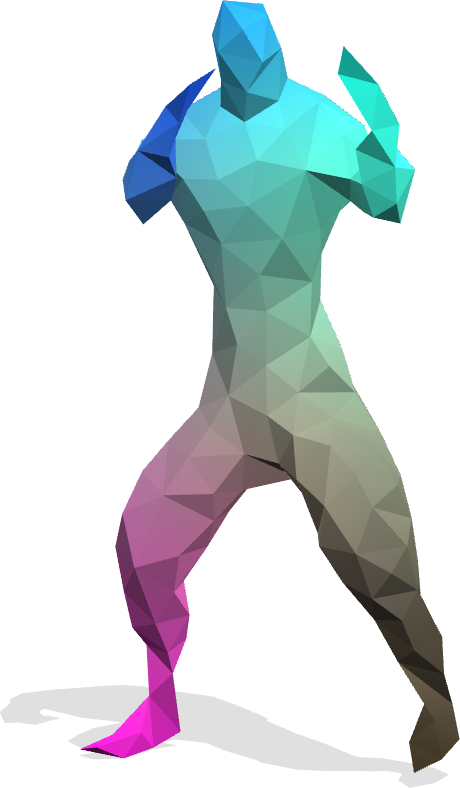}&
    \hspace{\hspaceCols}
    \includegraphics[height=\heightQ, width=\widthQ]{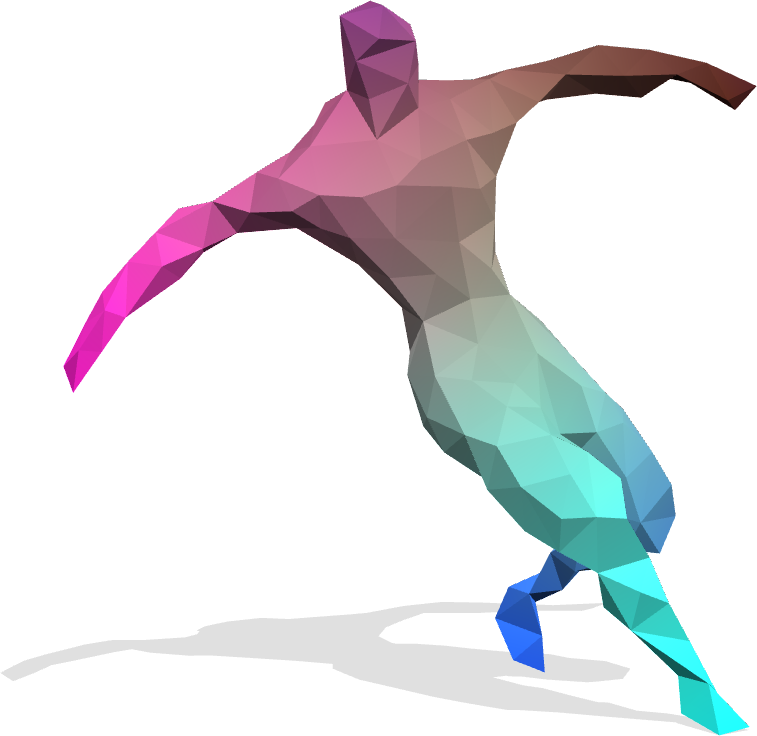}&
    \hspace{\hspaceCols}
    \includegraphics[height=\heightQ, width=\widthQ]{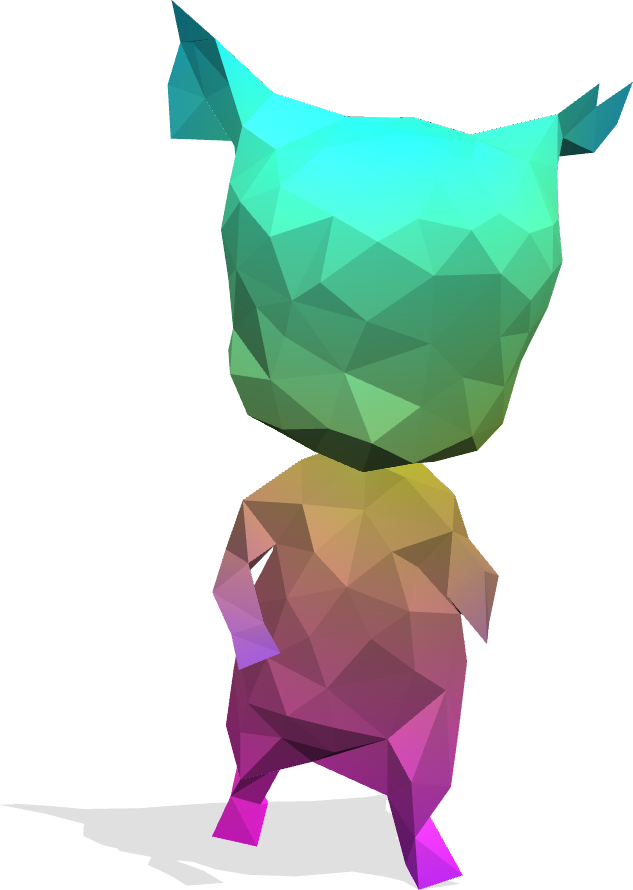}&
    \hspace{\hspaceCols}
    \includegraphics[height=\heightQ, width=\widthQ]{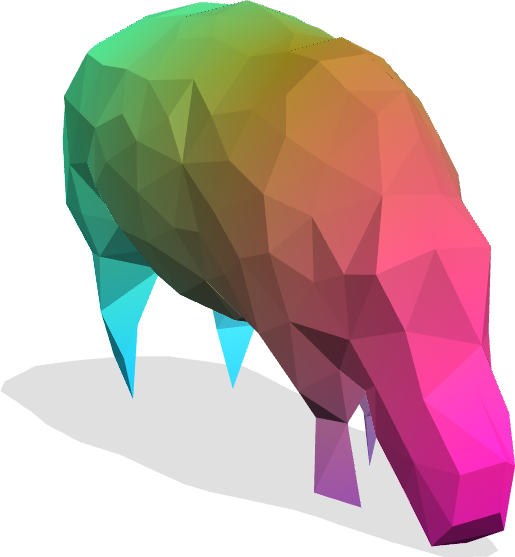}&
    \hspace{\hspaceCols}
    \includegraphics[height=\heightQ, width=\widthQ]{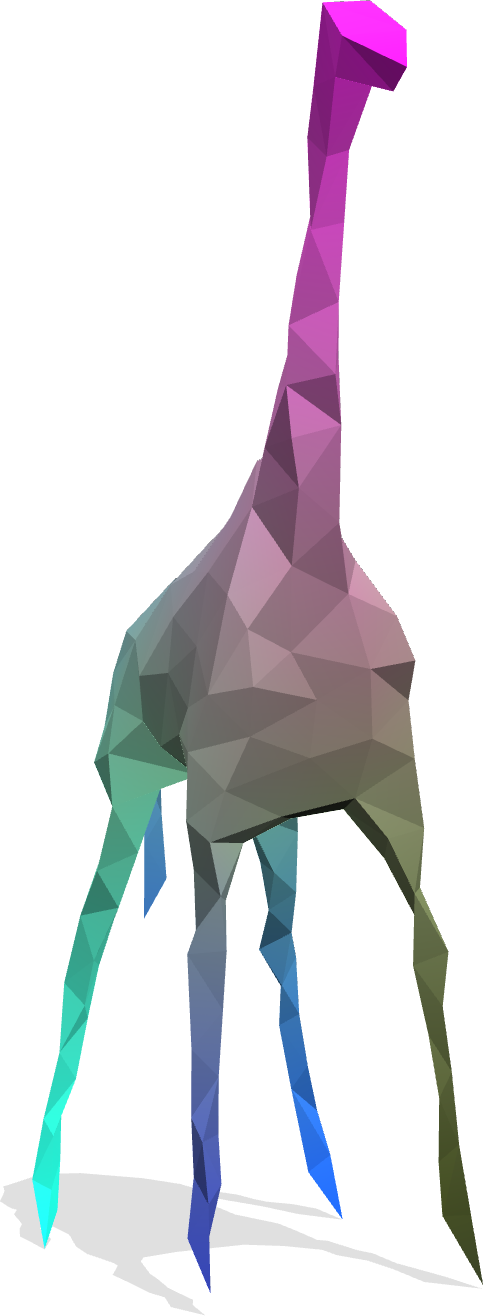}&
    \hspace{\hspaceCols}
    \includegraphics[height=\heightQ, width=\widthQ]{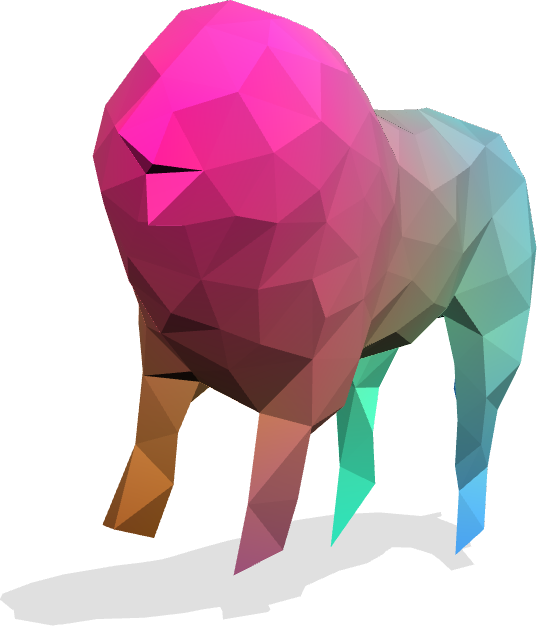}&
    \hspace{\hspaceCols}
    \includegraphics[height=\heightQ, width=\widthQ]{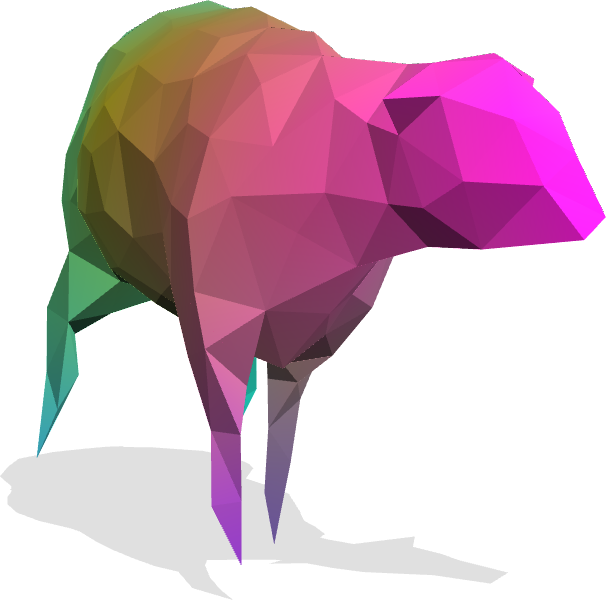}&
    \hspace{\hspaceCols}
    \includegraphics[height=\heightQ, width=\widthQ]{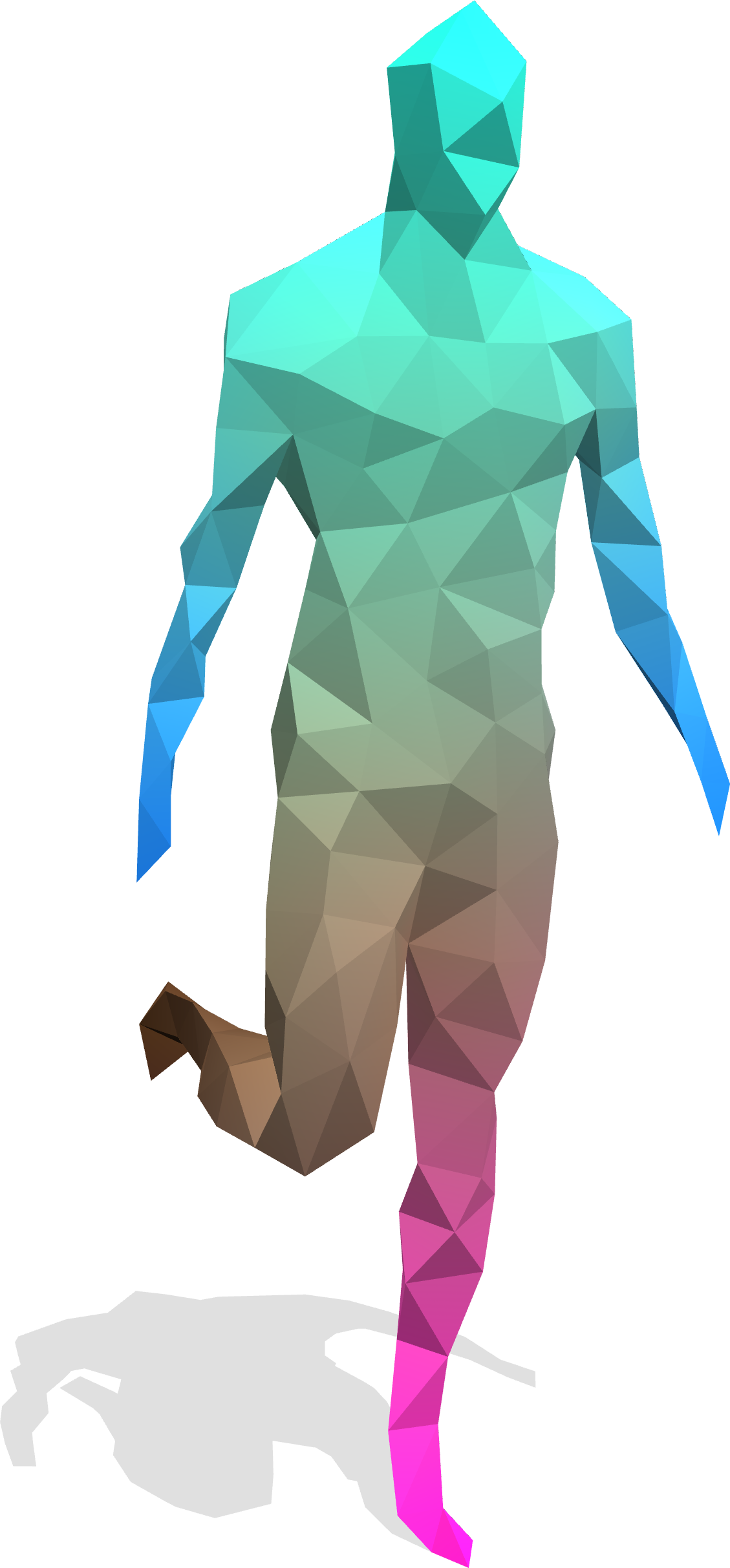}&
    \hspace{\hspaceCols}
    \includegraphics[height=\heightQ, width=\widthQ]{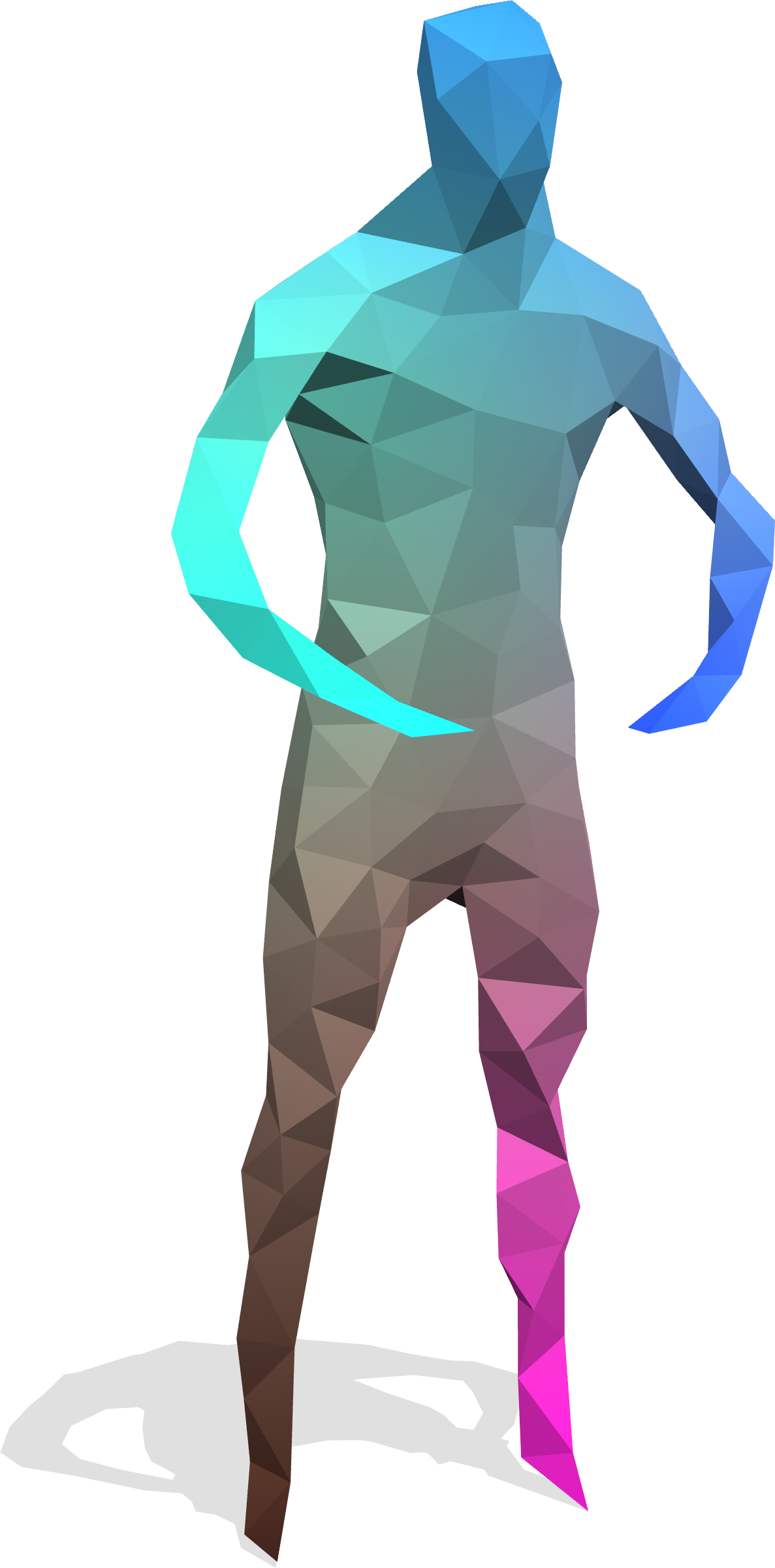}\\
    \rotatedCentering{90}{\heightQ}{\caoetal}&
    \includegraphics[height=\heightQ, width=\widthQ]{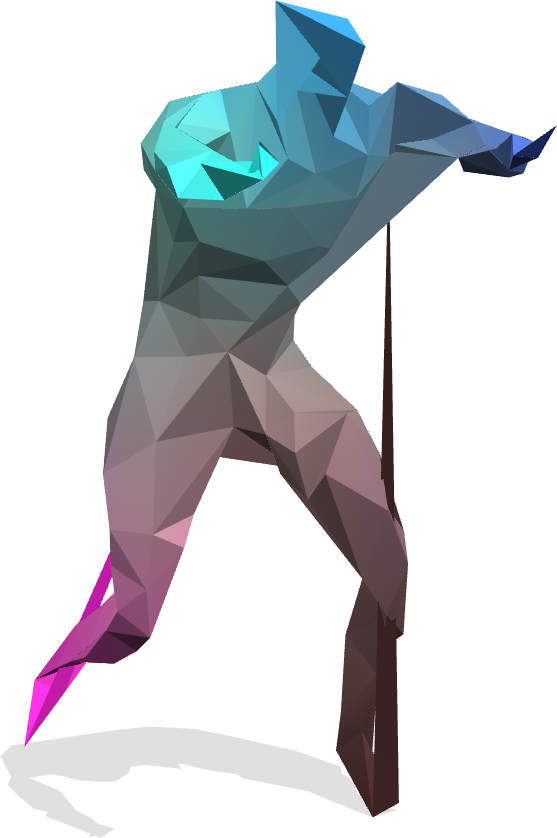}&
    \hspace{\hspaceCols}
    \includegraphics[height=\heightQ, width=\widthQ]{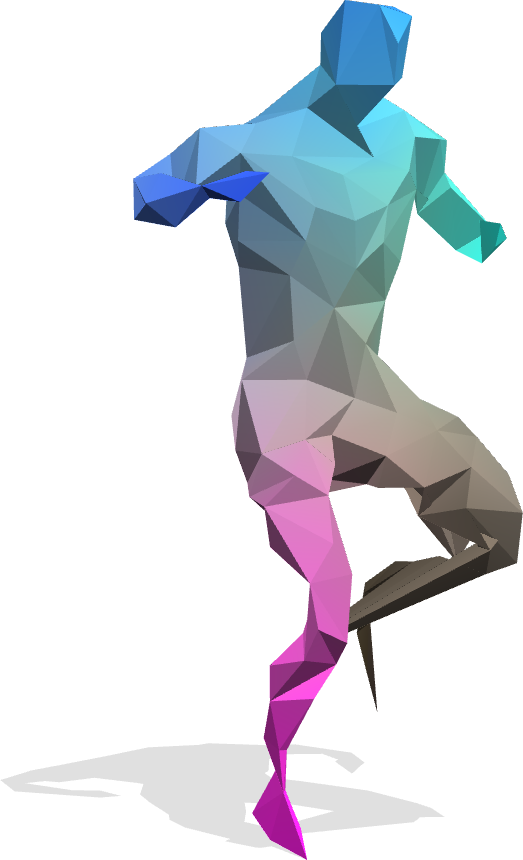}&
    \hspace{\hspaceCols}
    \includegraphics[height=\heightQ, width=\widthQ]{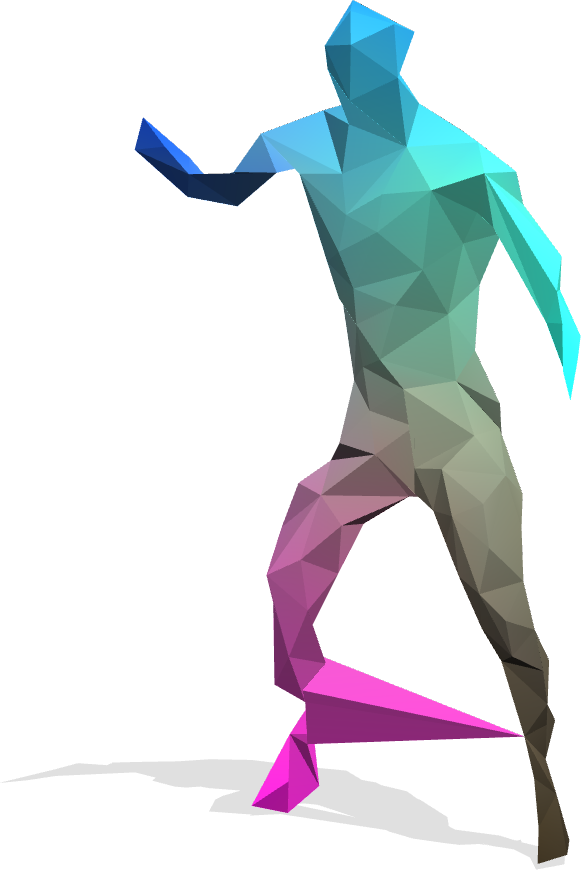}&
    \hspace{\hspaceCols}
    \includegraphics[height=\heightQ, width=\widthQ]{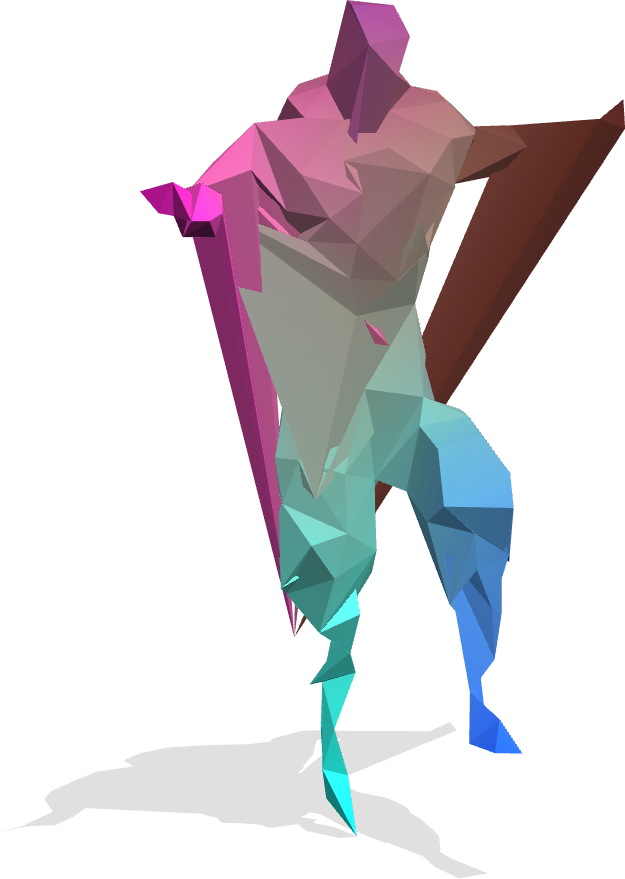}&
    \hspace{\hspaceCols}
    \includegraphics[height=\heightQ, width=\widthQ]{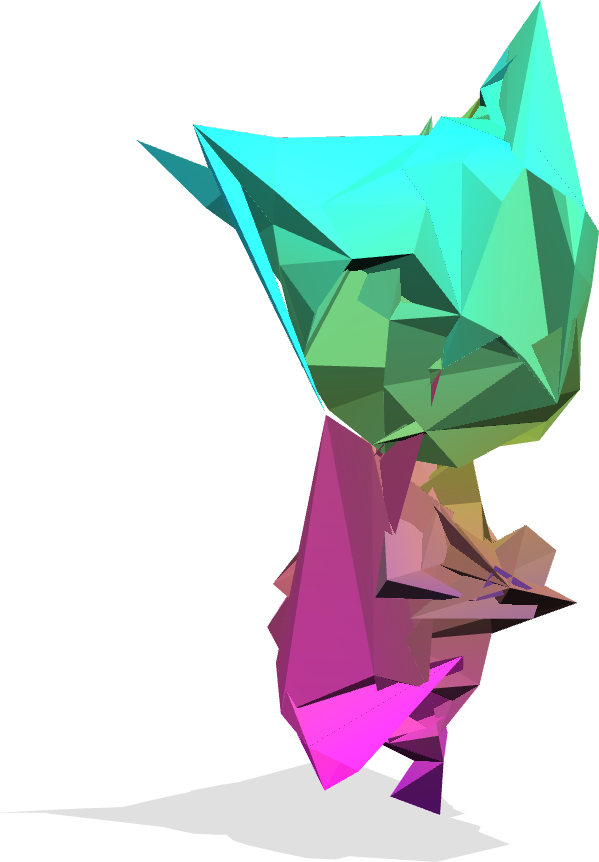}&
    \hspace{\hspaceCols}
    \includegraphics[height=\heightQ, width=\widthQ]{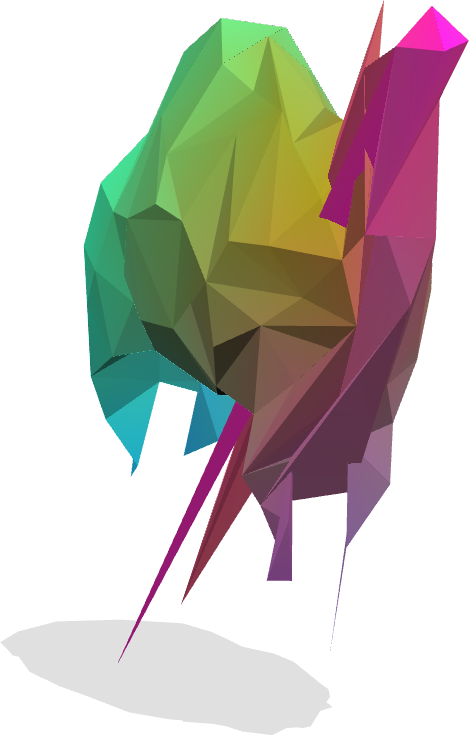}&
    \hspace{\hspaceCols}
    \includegraphics[height=\heightQ, width=\widthQ]{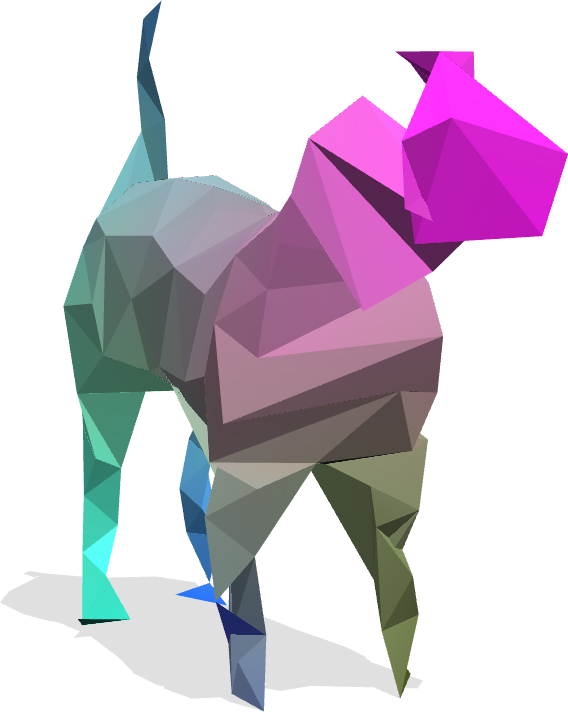}&
    \hspace{\hspaceCols}
    \includegraphics[height=\heightQ, width=\widthQ]{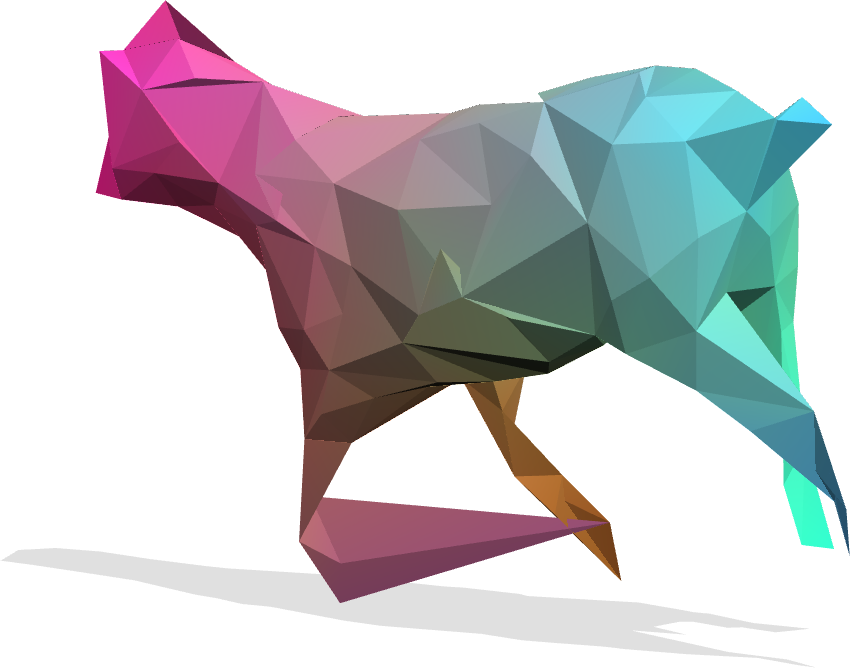}&
    \hspace{\hspaceCols}
    \includegraphics[height=\heightQ, width=\widthQ]{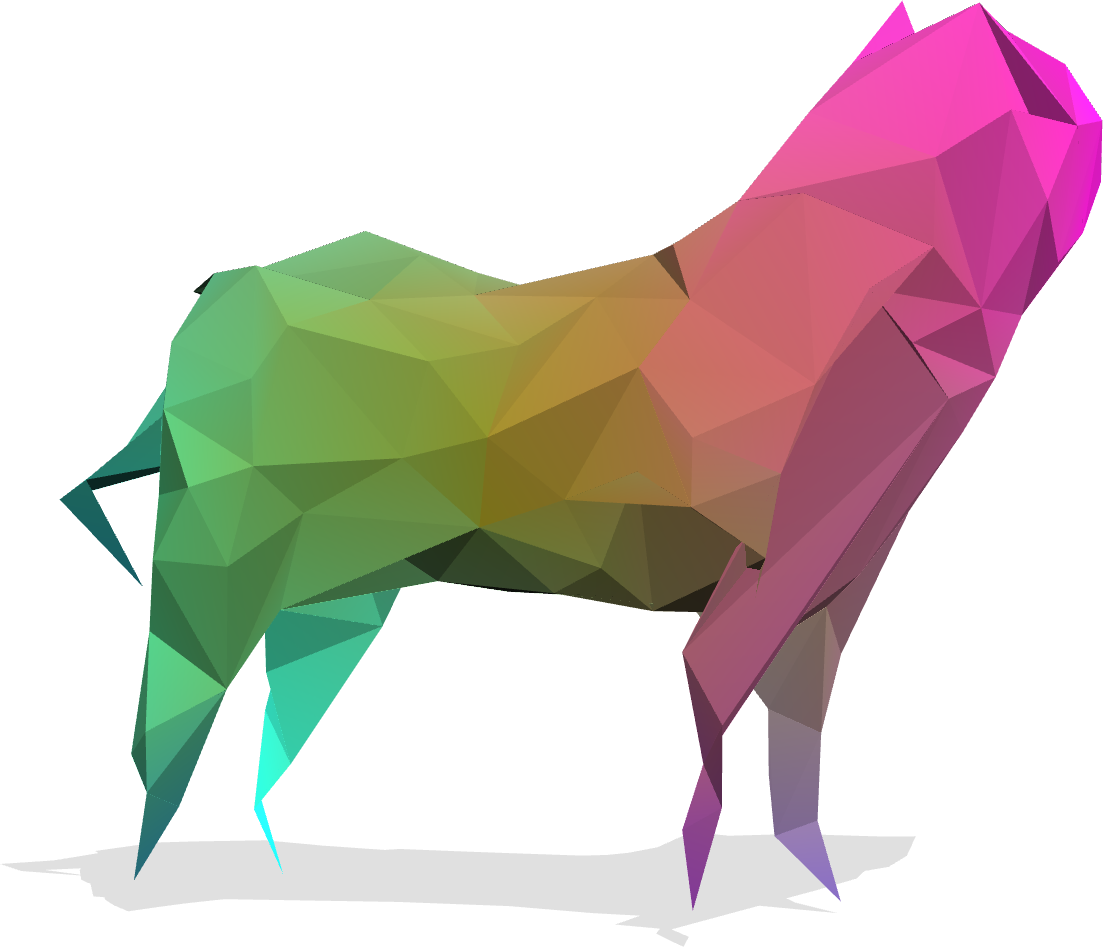}&
    \hspace{\hspaceCols}
    \includegraphics[height=\heightQ, width=\widthQ]{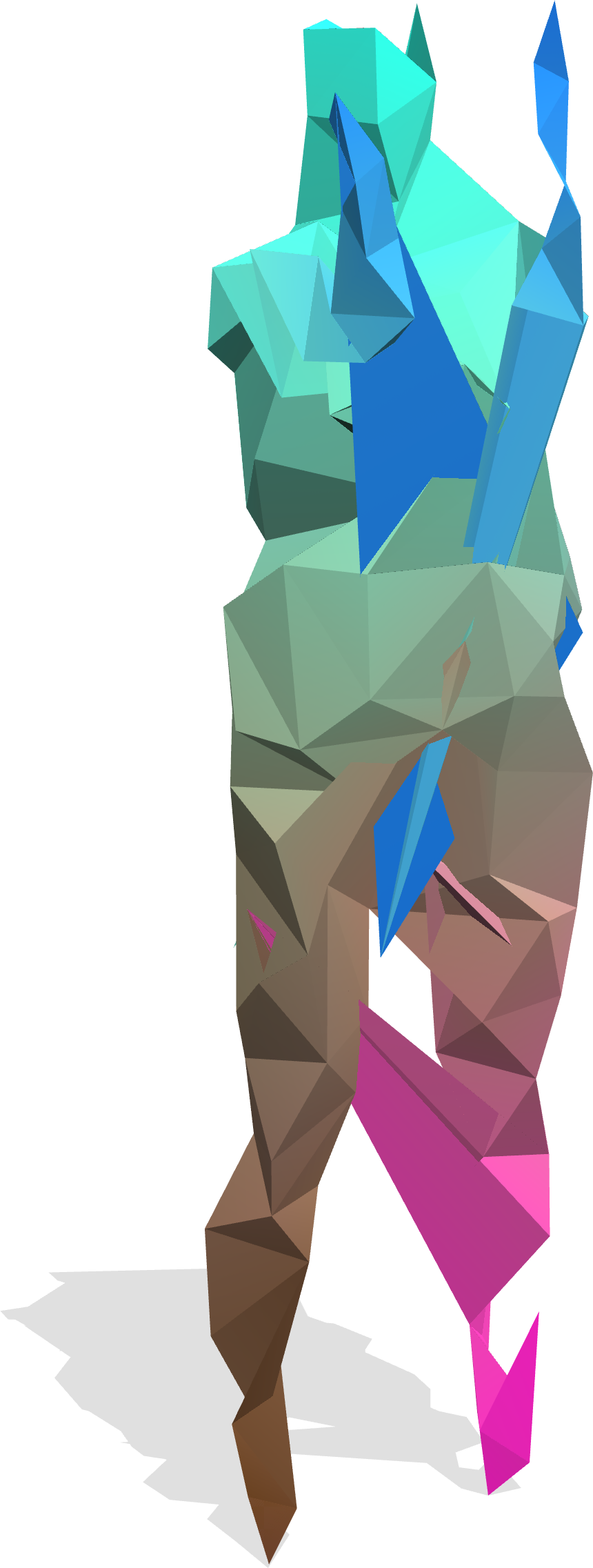}&
    \hspace{\hspaceCols}
    \includegraphics[height=\heightQ, width=\widthQ]{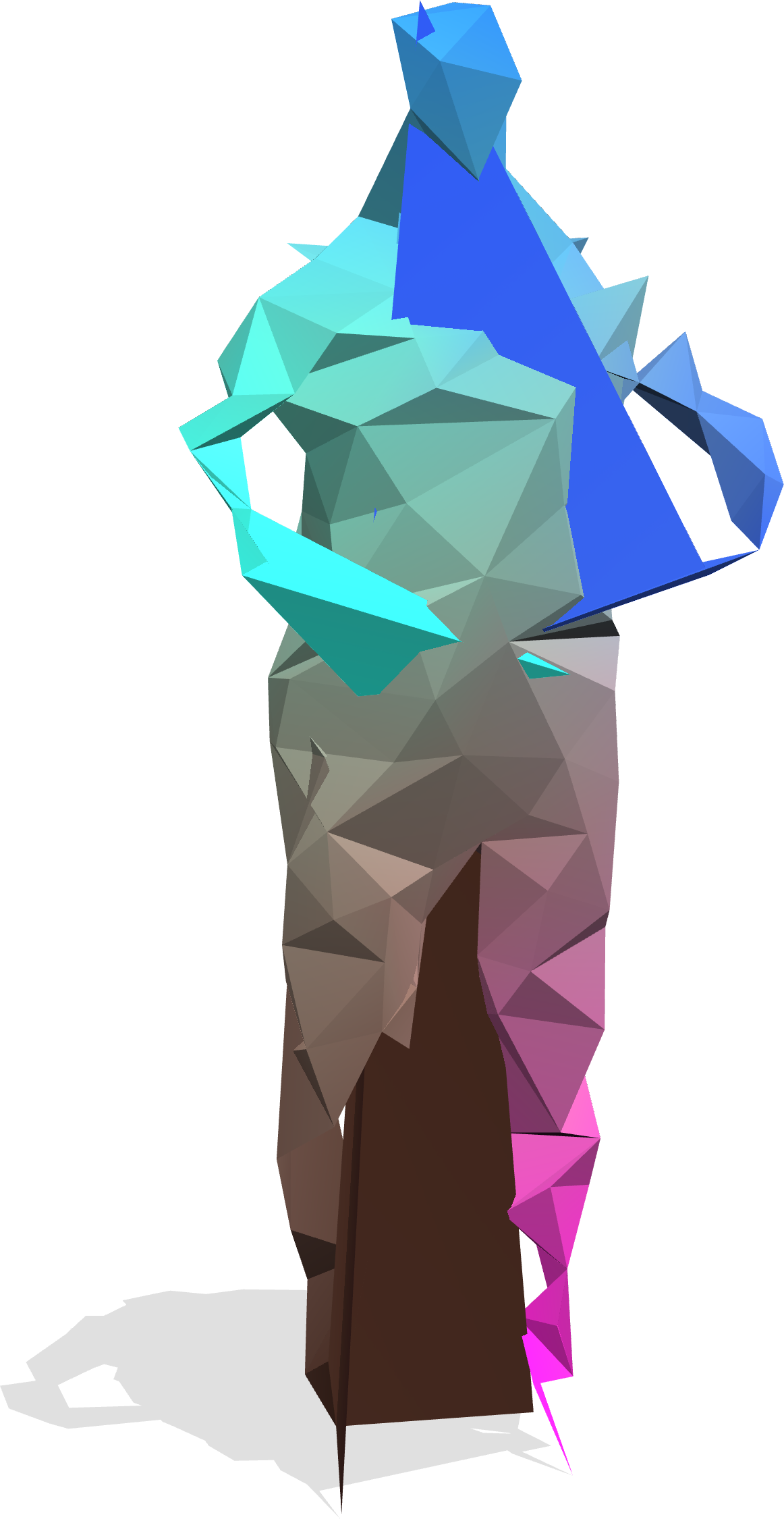}\\
    \rotatedCentering{90}{\heightQ}{\ours}&
    \includegraphics[height=\heightQ, width=\widthQ]{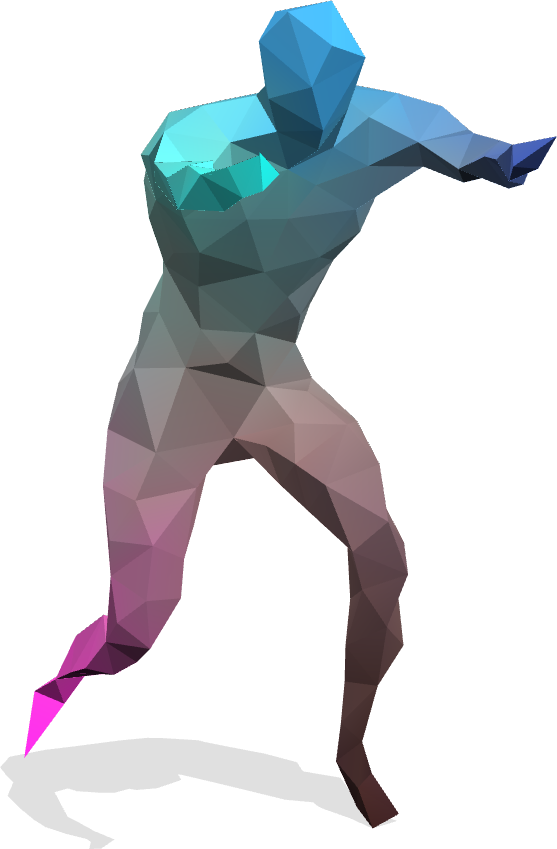}&
    \hspace{\hspaceCols}
    \includegraphics[height=\heightQ, width=\widthQ]{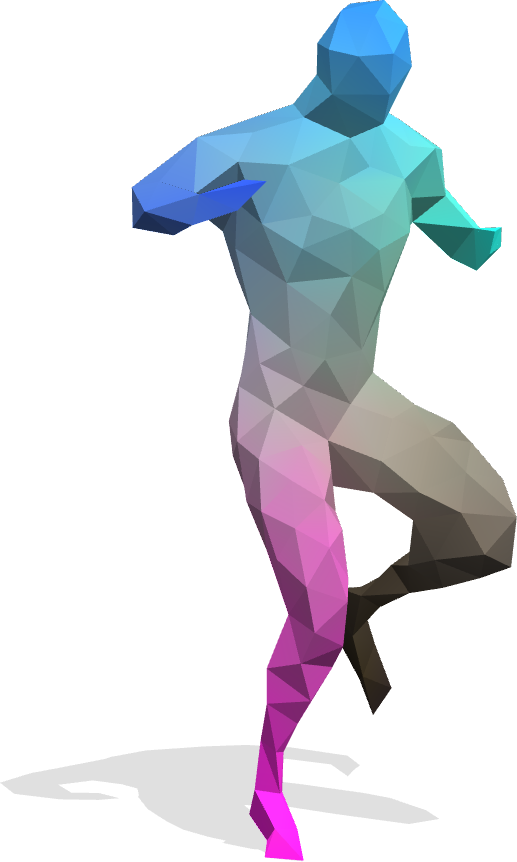}&
    \hspace{\hspaceCols}
    \includegraphics[height=\heightQ, width=\widthQ]{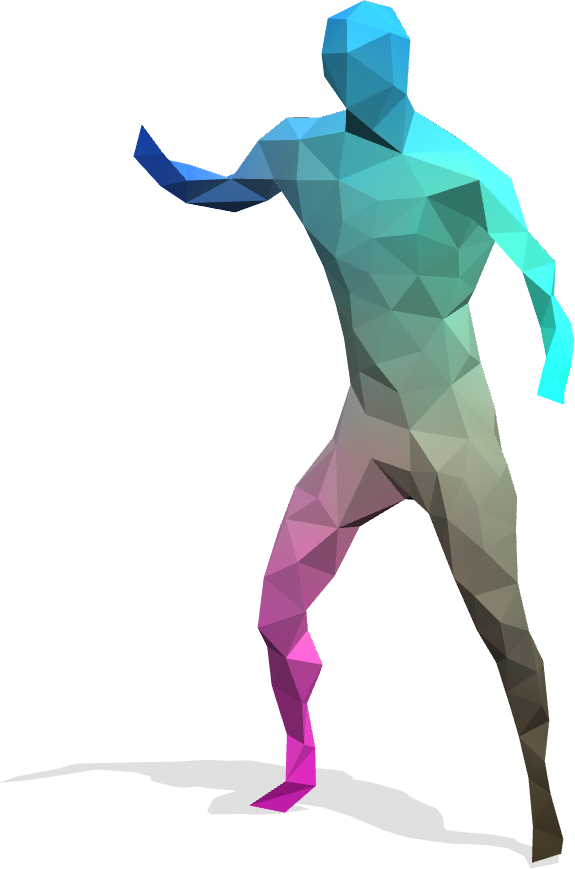}&
    \hspace{\hspaceCols}
    \includegraphics[height=\heightQ, width=\widthQ]{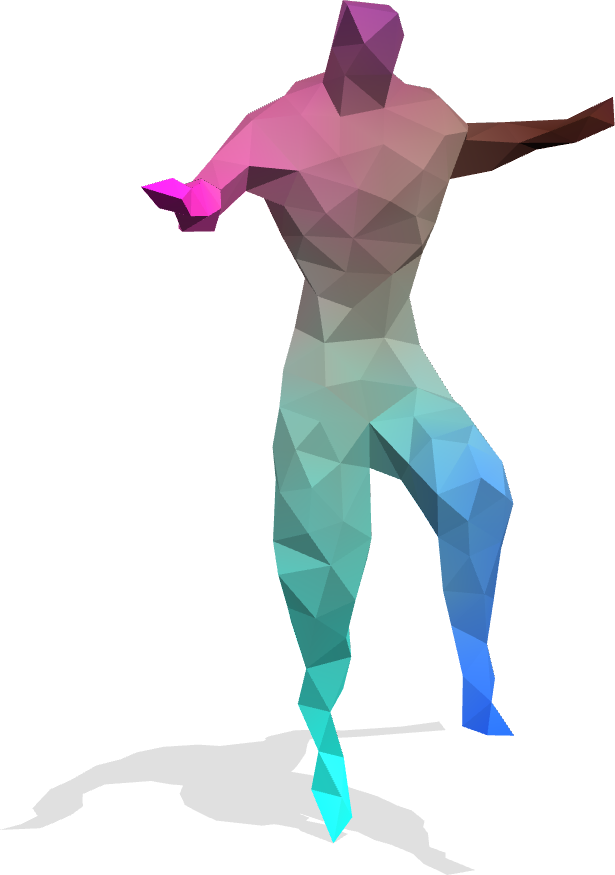}&
    \hspace{\hspaceCols}
    \includegraphics[height=\heightQ, width=\widthQ]{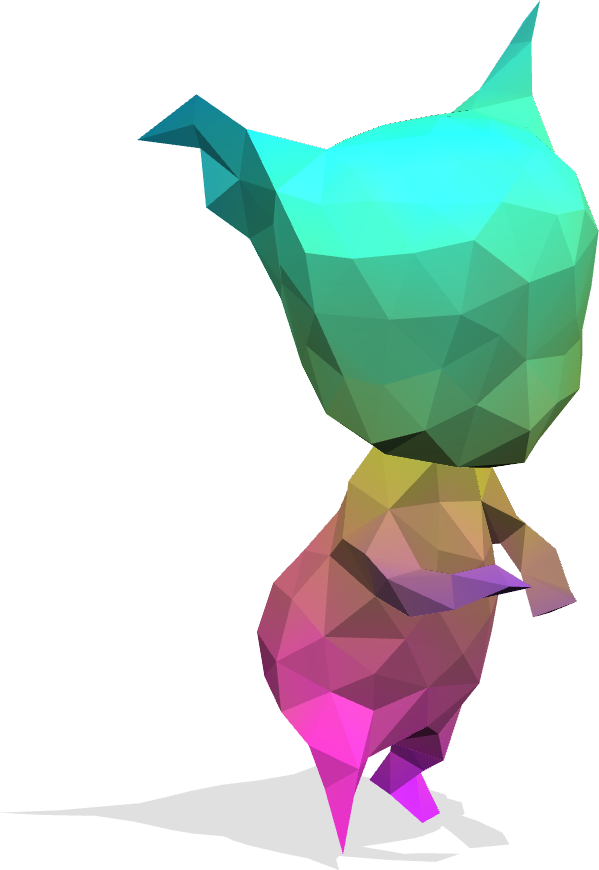}&
    \hspace{\hspaceCols}
    \includegraphics[height=\heightQ, width=\widthQ]{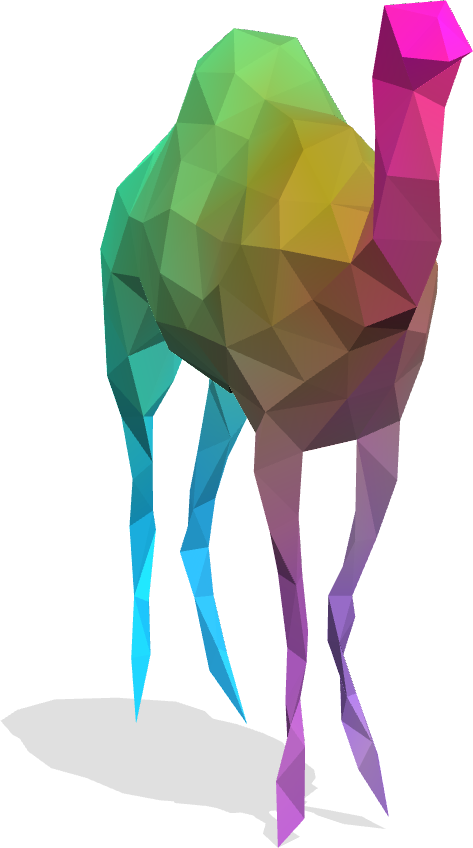}&
    \hspace{\hspaceCols}
    \includegraphics[height=\heightQ, width=\widthQ]{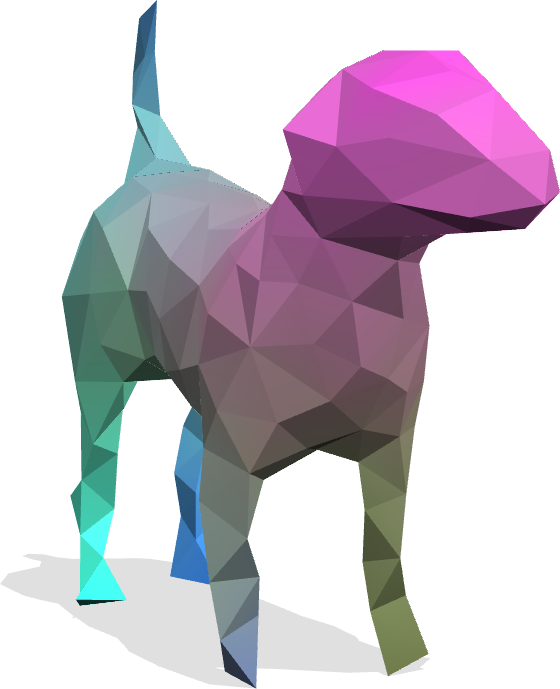}&
    \hspace{\hspaceCols}
    \includegraphics[height=\heightQ, width=\widthQ]{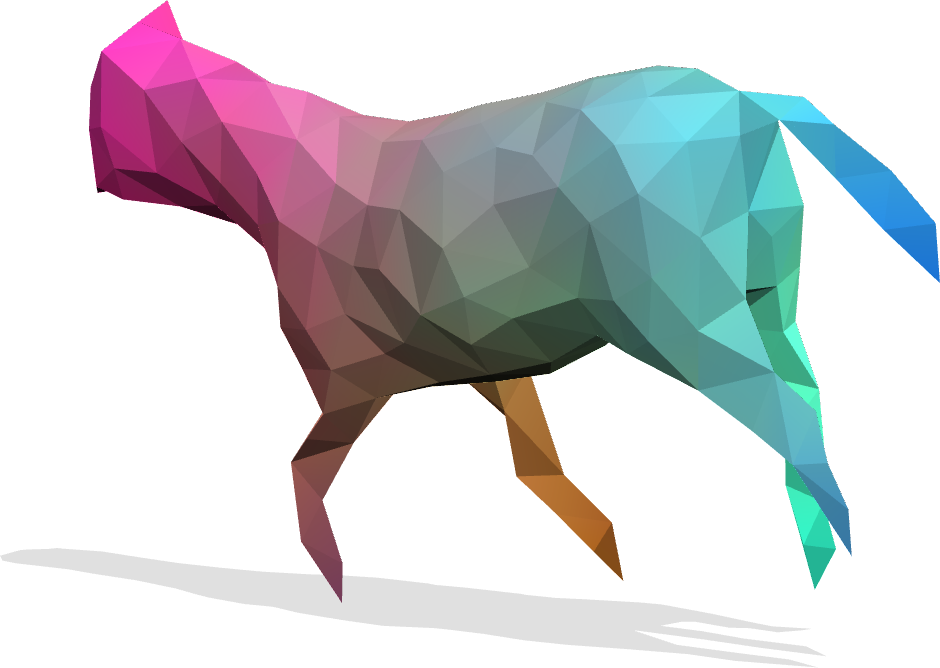}&
    \hspace{\hspaceCols}
    \includegraphics[height=\heightQ, width=\widthQ]{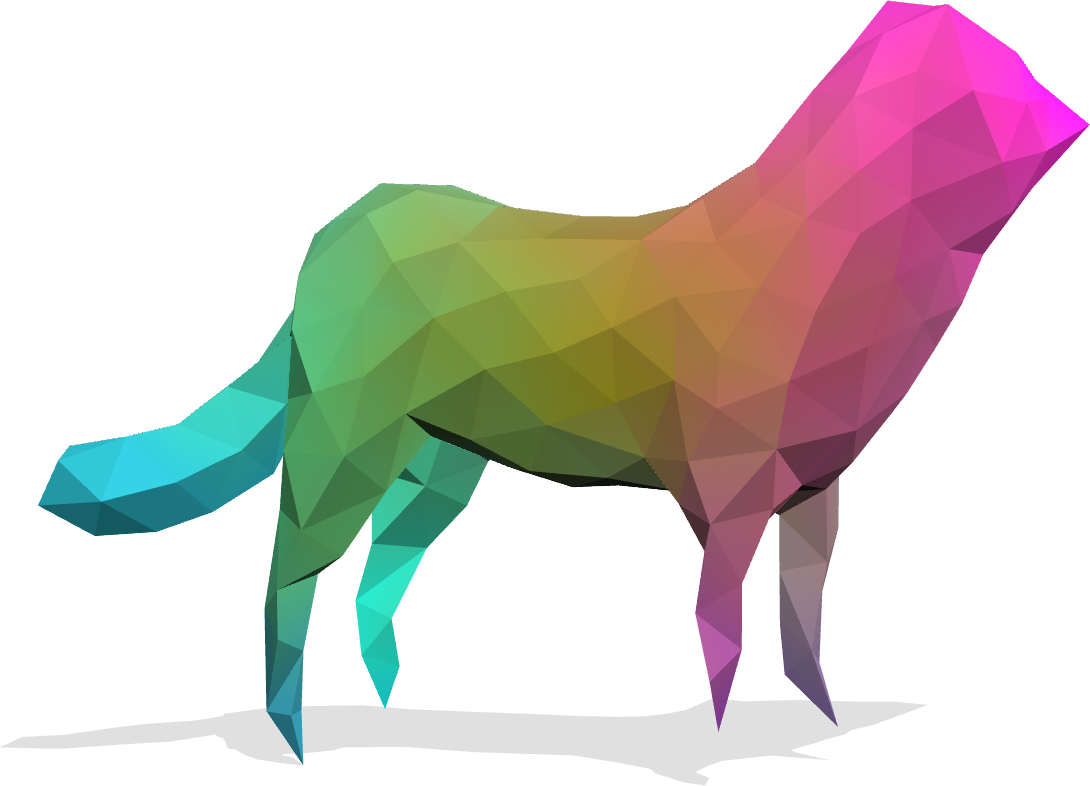}&
    \hspace{\hspaceCols}
    \includegraphics[height=\heightQ, width=\widthQ]{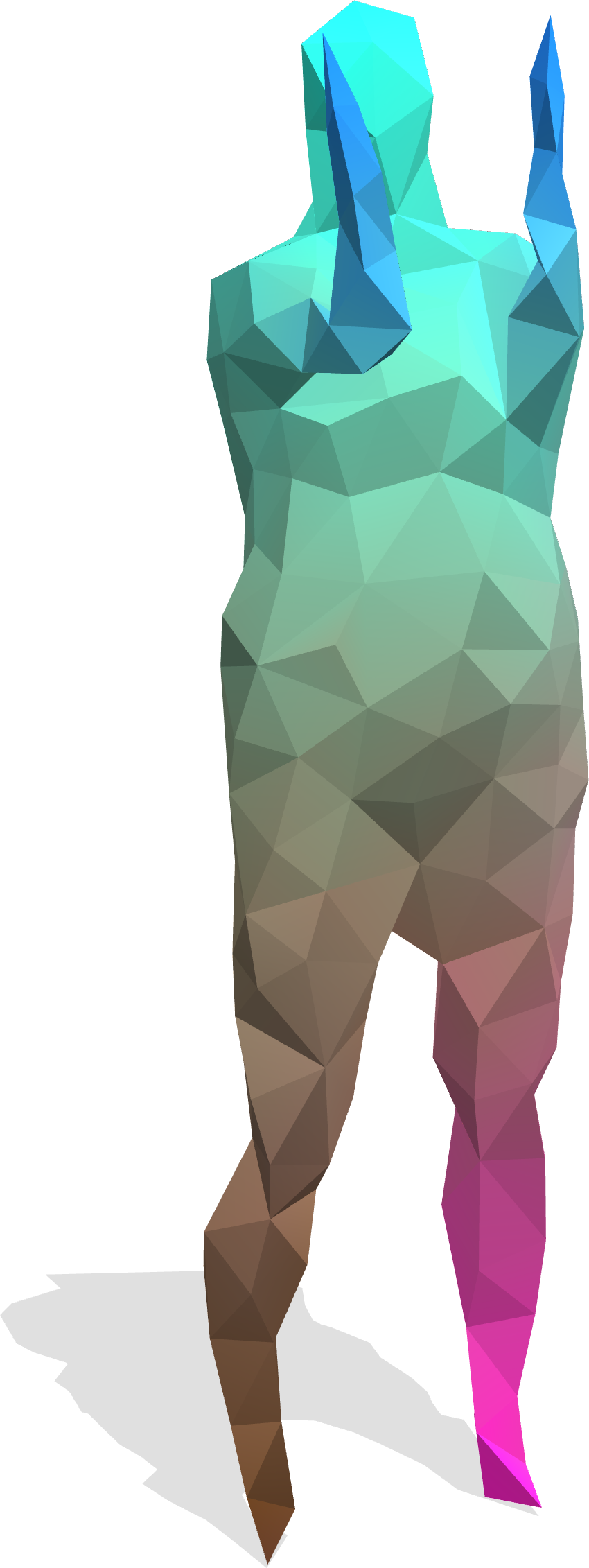}&
    \hspace{\hspaceCols}
    \includegraphics[height=\heightQ, width=\widthQ]{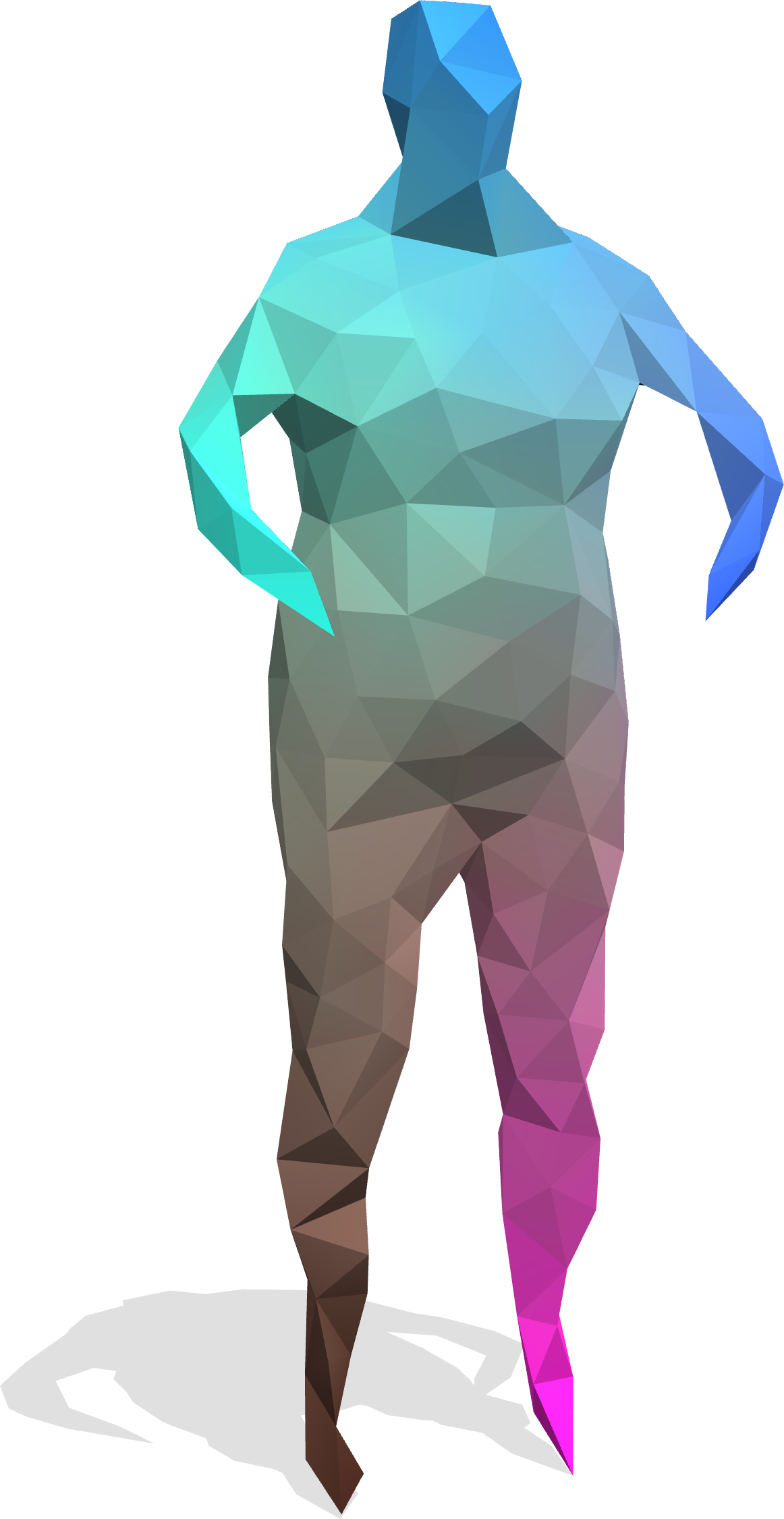}\\
\end{tabular}
    \vspace{-0.25cm}
    \caption{
    Qualitative Comparison on DT4D-Inter (\scircled{1}-\scircled{3}), DT4D-Intra (\scircled{4},\scircled{5}), SHREC'20 (\scircled{6},\scircled{7}), SMAL (\scircled{8},\scircled{9}) and FAUST (\scircled{10},\scircled{11}). 
    To better visualise the quality of matchings we transfer triangulation and the color from source to target shapes. 
    }
    \label{fig:qualitative-comparison}
\end{figure}

In~\cref{fig:pck}, we compare shape matching results w.r.t.\ mean geodesic errors, conformal distortion errors, and computation times.
Results show that the geometric consistency constraints help to decrease mean geodesic errors by sometimes over $50\%$. At the same time smoothness (by means of conformal distortion errors) of the resulting matching increases consistently across all datasets.
Only on SMAL dataset our approach lags behind by a small margin w.r.t.\ mean geodesic error but still produces a smoother matching.
Note that mean geodesic error does not directly quantify geometric consistency and even though results might have small error values, solutions can still be visually implausible (which can also be seen from qualitative results, cf.~\cref{fig:qualitative-comparison}, \cref{fig:appendix-qualitative}).

From the runtime perspective in~\cref{fig:pck} (\textit{bottom}) our method is by far the fastest method among the ILP-based ones (\smcomb, \fastdog, \ours). Moreover our dual optimisation scheme also helps in solving the primal problem~\eqref{eq:ilp-sm}. It produces most amount of globally optimal solutions, and fails the least number of times as compared to other ILP solvers, cf.~\cref{tab:opt-inf}. 
Note that mean geodesic errors of all ILP methods yield very similar matching quality since these methods solve the same optimisation problem. 

Furthermore, we observe in~\cref{fig:pck} that \kernelm~struggles on all datasets, likely because it searches over the space of permutation matrices to find shape correspondence which is not a valid assumption for shapes with different discretisation. Thus, we do not include \kernelm~in further comparisons.

In ~\cref{fig:qualitative-comparison} and~\cref{fig:appendix-qualitative} we visualize shape matching solutions. For showcasing the effect of geometric consistency we transfer triangulation from one shape to the other via the computed matching. We observe that geometric consistency constraints drastically improve solution quality in our approach as compared to \caoetal\ which does not impose geometric consistency. 

\myparagraph{Ablation: Faster Optimisation.}
In~\cref{fig:convergence} we compare the convergence of our solver to the fastest existing approach \fastdog\ for solving the ILP formulation~\eqref{eq:ilp-sm}. 
We observe that our algorithm produces much better primal and dual objectives (in significantly less time) compared to \fastdog. 
Moreover, in~\cref{tab:opt-inf} we compare the number of globally optimal and infeasible solutions found by each solver.
Our solver finds globally optimal solutions in most cases and the smallest number of infeasible solutions. This shows that our improved dual optimisation also facilitates in primal recovery.
In~\cref{fig:runtime} we compare runtimes of all ILP-based methods on different problem sizes by varying the resolution (number of triangles) for 5 pairs from the FAUST dataset. We observe that \lprelax\ does not scale well with growing problem sizes. While \fastdog\ is faster than all other existing methods, we outperform it as well. 

Note that our solver related improvements from Sec.~\ref{sec:faster_optimisation} are applicable to ILPs beyond shape matching task as well. To this end we evaluate our solver on some of the difficult datasets used in~\cite{abbas2022fastdog} and obtain $5$--$10$ times runtime improvement. We refer to Sec.~\ref{sec:evaluation_other_ilps} for more details.

\begin{figure}[!h]
    \centering
    \begin{tabular}{cc}
        \hspace{-1cm}
        \newcommand{\drawsquare}[1]{%
\begin{tikzpicture}[#1]%
\node[fill=cPLOT3, inner sep=0pt,minimum size=7pt] {};
\end{tikzpicture}%
}

\newcommand{\drawplus}[1]{%
\begin{tikzpicture}[#1]%
\node[mark size=4pt, color=cPLOT3, line width=1.5pt] at (0, -1) {\pgfuseplotmark{+}};
\end{tikzpicture}%
}

\newcommand{\drawcircle}[1]{%
\begin{tikzpicture}[#1]%
\node[mark size=3pt, color=cPLOT5, line width=1.5pt] at (0, -1) {\pgfuseplotmark{o}};
\end{tikzpicture}%
}

\newcommand{\drawtriangle}[1]{%
\begin{tikzpicture}[#1]%
\node[regular polygon,regular polygon sides=3, fill=cPLOT5, inner sep=0pt,minimum size=9pt] {};
\end{tikzpicture}%
}

\pgfplotsset{%
	label style = {font=\large},
	tick label style = {font=\large},
	title style =  {font=\Large},
	legend style={  fill= gray!10,
		fill opacity=0.6, 
		font=\normalfont,
		draw=gray!20, %
		text opacity=1}
}
\newcommand{\runtimeLineWidth}{3pt}
\newcommand{\rtCplotWidth}{0.8\columnwidth}
\newcommand{\rtCplotHeight}{0.28\columnwidth}
\pgfplotscreateplotcyclelist{plot_colors}{
smooth,line width=\runtimeLineWidth,color=cPLOT3\\
smooth,line width=\runtimeLineWidth,color=cPLOT5\\%
}
\begin{tikzpicture}[scale=0.5, transform shape]
	\begin{axis}[%
            cycle list name=plot_colors,
		width=\rtCplotWidth,
		height=\rtCplotHeight,
		title=FAUST: Solver Convergence,
		title style={yshift=-0.25cm},
		scale only axis,
		grid=major,
		legend style={
			at={(0.98,0.98)},
			anchor=north east,
			legend columns=1,
			legend cell align={left}},
             xlabel={Time [s]},
             ylabel shift = -0.1cm,
             ylabel={\large $\leftarrow$ Relative Gaps $\leftarrow$},
		xmin=-20,
		xmax=999,
		ymin=0.00001,
		ymax=0.1,
            ymode=log,
		ytick={0.000001, 0.00001, 0.0001, 0.001, 0.01, 0.1, 1},
		ylabel near ticks,
		]

    \addplot table [x index = 0, y index = 1, col sep=comma] {figs/tikz/logs/faust_fastdog_final_lb_results_500_upwards.csv};
    \addlegendentry{\fastdog\, \drawplus{}};

    \addplot table [x index = 0, y index = 1, col sep=comma] {figs/tikz/logs/faust_v2_lbfgs_split_final_lb_results_500_upwards.csv};
    \addlegendentry{\ours\, \drawcircle{}};

    \node[color=cPLOT3, inner sep=0pt, scale=2.5, minimum size=7pt] at (372.29020833333334, 0.00260066786755601) {\pgfuseplotmark{+}};

    \node[color=cPLOT5, inner sep=1pt, scale=1.8, line width=1pt] at (120.79632727272725, 0.002610783941259561) {\pgfuseplotmark{o}};
    \coordinate (start) at (910, 0.0001);
    \coordinate (end) at (78, 0.0001);

    \draw[->, red!70, decorate, decoration={zigzag, amplitude=1mm, segment length=3mm, pre=lineto, pre length=3mm, post=lineto, post length=3mm}] (start) -- node[midway, below] {\large $11\times$ faster} (end);
    \end{axis}
\end{tikzpicture}&
        \newcommand{\drawplus}[1]{%
\begin{tikzpicture}[#1]%
\node[mark size=4pt, color=cPLOT3, line width=1.5pt] at (0, -1) {\pgfuseplotmark{+}};
\end{tikzpicture}%
}

\newcommand{\drawcircle}[1]{%
\begin{tikzpicture}[#1]%
\node[mark size=3pt, color=cPLOT5, line width=1.5pt] at (0, -1) {\pgfuseplotmark{o}};
\end{tikzpicture}%
}

\pgfplotsset{%
	label style = {font=\large},
	tick label style = {font=\large},
	title style =  {font=\Large},
	legend style={  fill= gray!10,
		fill opacity=0.6, 
		font=\normalfont,
		draw=gray!20, %
		text opacity=1}
}
\newcommand{\runtimeLineWidth}{3pt}
\newcommand{\rtCplotWidth}{0.8\columnwidth}
\newcommand{\rtCplotHeight}{0.28\columnwidth}
\pgfplotscreateplotcyclelist{plot_colors}{
smooth,line width=\runtimeLineWidth,color=cPLOT3\\
smooth,line width=\runtimeLineWidth,color=cPLOT5\\%
}
\begin{tikzpicture}[scale=0.5, transform shape]
	\begin{axis}[%
            cycle list name=plot_colors,
		width=\rtCplotWidth,
		height=\rtCplotHeight,
		title=DT4D-Intra: Solver Convergence,
		title style={yshift=-0.25cm},
		scale only axis,
		grid=major,
		legend style={
			at={(0.98,0.98)},
			anchor=north east,
			legend columns=1,
			legend cell align={left}},
             xlabel={Time [s]},
		xmin=-10,
		xmax=499,
		ymin=0.000005,
		ymax=0.1,
            ymode=log,
		ytick={0.000001, 0.00001, 0.0001, 0.001, 0.01, 0.1, 1},
		ylabel near ticks,
		]

    \addplot table [x index = 0, y index = 1, col sep=comma] {figs/tikz/logs/dt4d_fastdog_final_lb_results_500_upwards.csv};
    \addlegendentry{\fastdog\, \drawplus{}};

    \addplot table [x index = 0, y index = 1, col sep=comma] {figs/tikz/logs/dtd_lbfgs_split_final_lb_results_500_upwards.csv};
    \addlegendentry{\ours\, \drawcircle{}};

    \node[color=cPLOT3, inner sep=0pt, scale=2.5, minimum size=7pt] at (260.09934545454547, 0.0025344734923885746) {\pgfuseplotmark{+}};

    \node[color=cPLOT5, inner sep=1pt, scale=1.8, line width=1pt] at (75.4741636363636, 0.0011593452992476594) {\pgfuseplotmark{o}};

    \coordinate (start) at (437, 0.00003);
    \coordinate (end) at (47, 0.00003);

    \draw[->, red!70, decorate, decoration={zigzag, amplitude=1mm, segment length=3mm, pre=lineto, pre length=3mm, post=lineto, post length=3mm}] (start) -- node[midway, below] {\large $9\times$ faster} (end);
    \end{axis}
\end{tikzpicture}
    \end{tabular}%
    \vspace{-0.25cm}
    \caption{{Convergence} plots averaged over a total of 55 instances with varying shape resolutions in $\{500, 550, \cdots, 1000\}$. 
    The curves (---) and markers ($+$, $\circ$) represent quality of dual objectives (by relative dual gaps) and primal objectives (by primal-dual gaps), respectively. 
    Our combined solver improvements (Sec~\ref{sec:faster_optimisation}) of quasi-newton updates and constraint splitting give up to $11\times$ improvement in runtime of dual optimisation which also aids in faster primal solutions.
    }
    \label{fig:convergence}
\end{figure}

\myparagraph{Ablation: Energy Adaptation.}
In \cref{tab:energy-compare}, we compare runtimes and geodesic errors when the ILP~\eqref{eq:ilp-sm} is solved with the elastic energy of~\cite{windheuser2011large} or with our proposed energy~\eqref{eq:cost}.
Our solver is the fastest with both energies and furthermore our energy plays a crucial role in decreasing matching errors and runtimes.

\begin{table}[!h]
\centering
{\footnotesize
    \begin{tabular}{@{}lrrr rrr@{}}
    \toprule
     & \multicolumn{3}{c}{\textbf{Elastic Energy}} & \multicolumn{3}{c}{\textbf{Our Energy}} \\ \cmidrule(l){2-4} \cmidrule(l){5-7} 
      & Time [s.] &  Geo.\ Error & \# Inf. & Time [s.] &  Geodesic Error  & \# Inf. \\ \midrule
     {\smcomb}& 648& 0.087& 12& 376& 0.042 &9\\
     {\fastdog}& 925& 0.083&36 & 327& \textbf{0.041} &4\\
     {\ours}& {423}& 0.084&8 & \textbf{134}& \textbf{0.041} &\textbf{2}\\
     \bottomrule
    \end{tabular}
}
    \caption{
    Comparison of solvers for~\eqref{eq:ilp-sm} using elastic energy of~\cite{windheuser2011large} and our energy~\eqref{eq:cost}. We report mean runtime in seconds (Time [s.]), mean geodesic errors (Geo.\ Error), and number of infeasible solutions (\# Inf.) on FAUST dataset (with $450$ faces). 
    }
    \label{tab:energy-compare}
\end{table}

\subsection{Scaling to Higher Resolutions}
\begin{figure}[!h]
    \vspace{-0.5cm}
    \centering
    \begin{tabular}{cc}
        \begin{tabular}{c}
             \includegraphics[width=0.35\columnwidth,trim={0.8cm 0.8cm 0.8cm 0.8cm}]{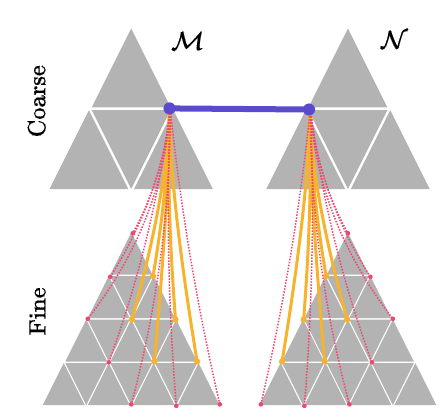} 
        \end{tabular}
        \hspace{-1cm}
        &
        \def\pathOurs{figs/qualitative/c2f_ours/}
\def\pathCao{figs/qualitative/c2f_caoetal/}
\def\srcEnd{_M.png}
\def\trgtEnd{_N.png}
\def\columnOne{tr_reg_087-tr_reg_090}
\def\columnTwo{tr_reg_080-tr_reg_090}
\def\columnTwoTwo{Shuffling165-Shuffling092}
\def\columnThree{DancingRunningMan312-DancingRunningMan279}
\def\columnFour{Falling230-Falling271}
\def\columnFive{Floating099-KettlebellSwing047}
\def\columnSix{Floating099-KettlebellSwing047}
\def\heightQ{1.55cm}
\def\widthQ{1.6cm}
\def\hspaceCols{-0.15cm}
\begin{tabular}{ccccccc}
    &\hspace{\hspaceCols} \scircled{1} &\hspace{\hspaceCols}  \scircled{2} &\hspace{\hspaceCols}  \scircled{3} &\hspace{\hspaceCols}  \scircled{4} &\hspace{\hspaceCols}  \scircled{5} &\hspace{\hspaceCols}  \scircled{6}\\
    \setlength{\tabcolsep}{0pt} 
    \rotatedCentering{90}{\heightQ}{Source}&
    \hspace{\hspaceCols}
    \includegraphics[height=\heightQ, width=\widthQ]{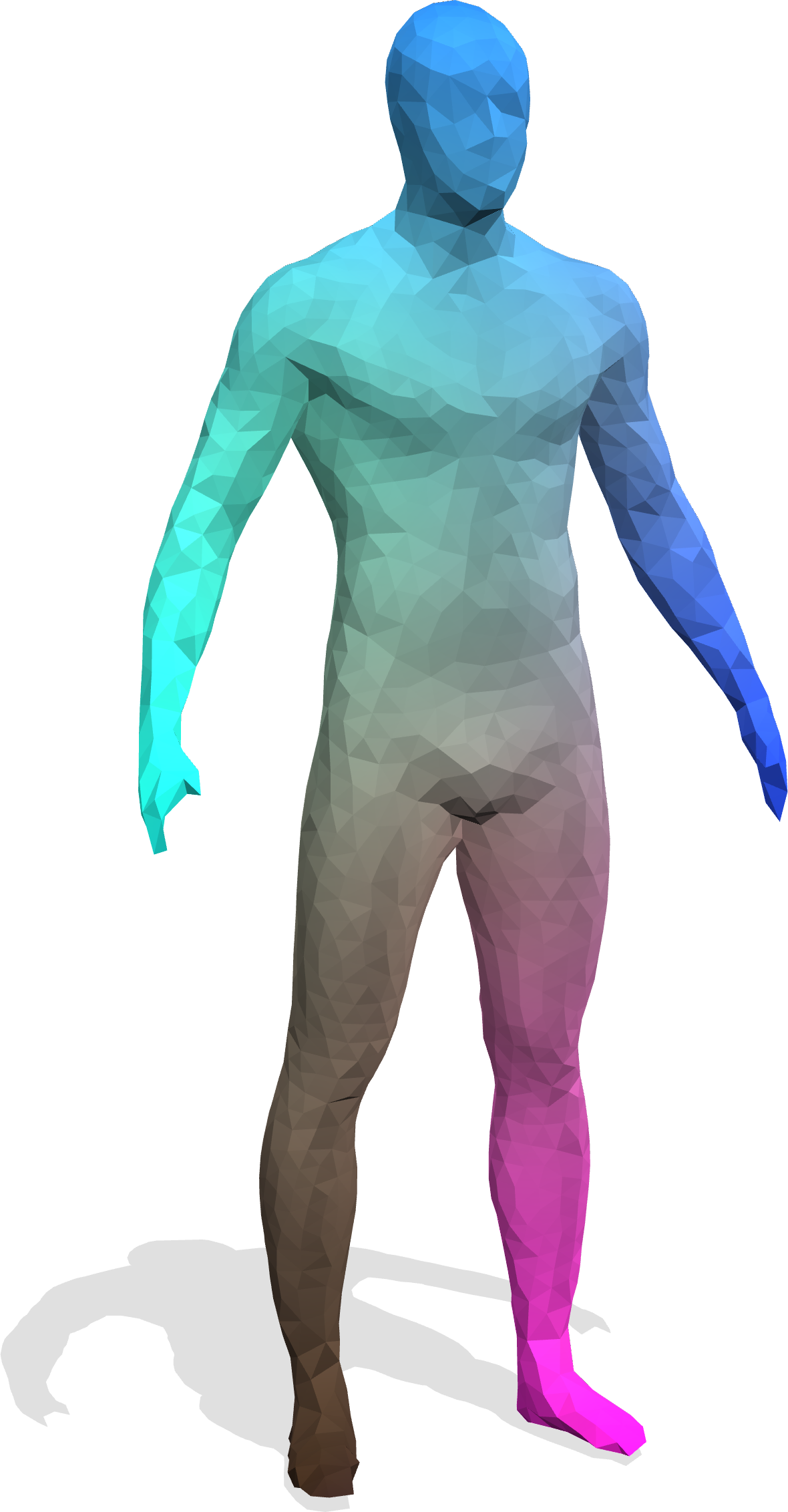}&
    \hspace{\hspaceCols}
    \includegraphics[height=\heightQ, width=\widthQ]{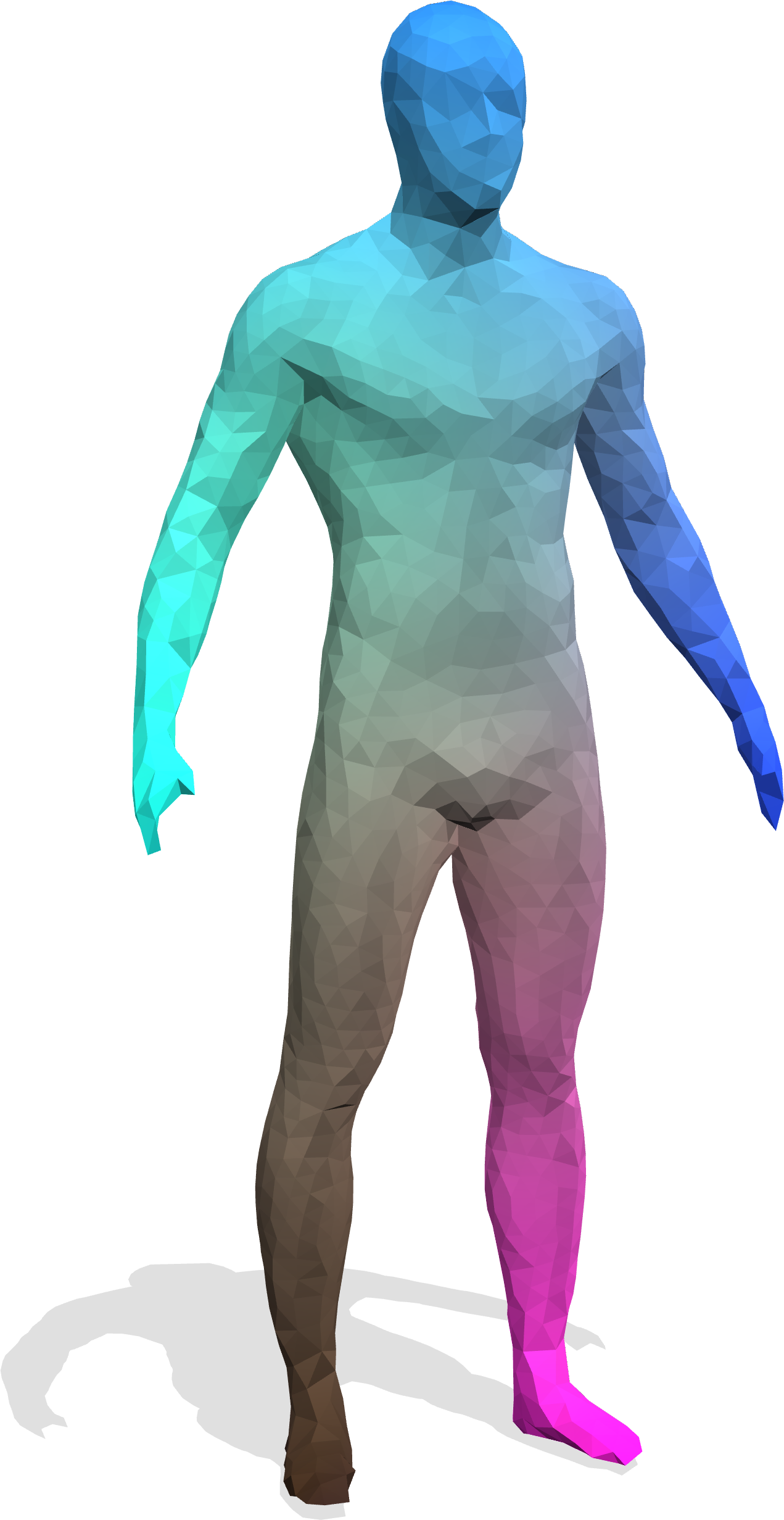}&
    \hspace{\hspaceCols}
    \includegraphics[height=\heightQ, width=\widthQ]{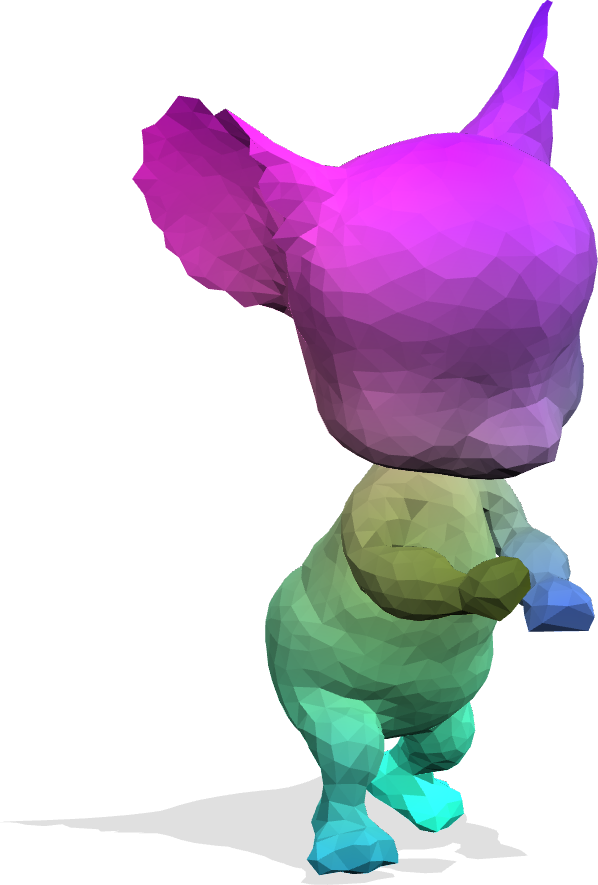}&
    \hspace{\hspaceCols}
    \includegraphics[height=\heightQ, width=\widthQ]{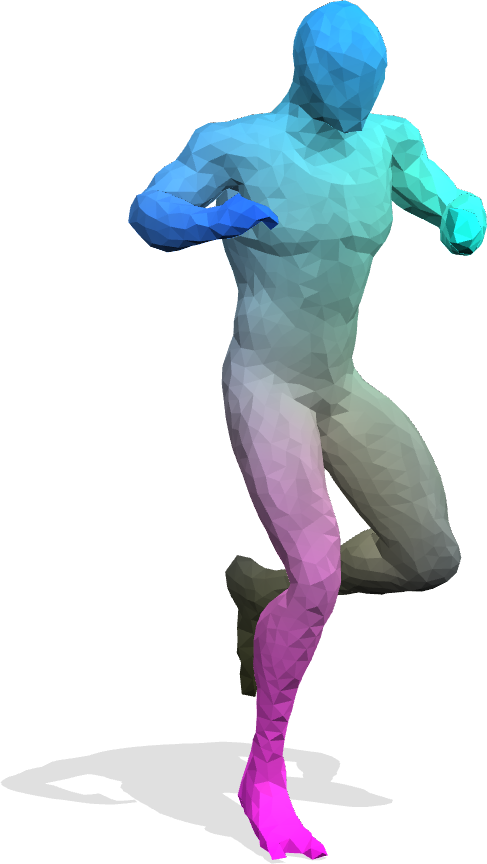}&
    \hspace{\hspaceCols}
    \includegraphics[height=\heightQ, width=\widthQ]{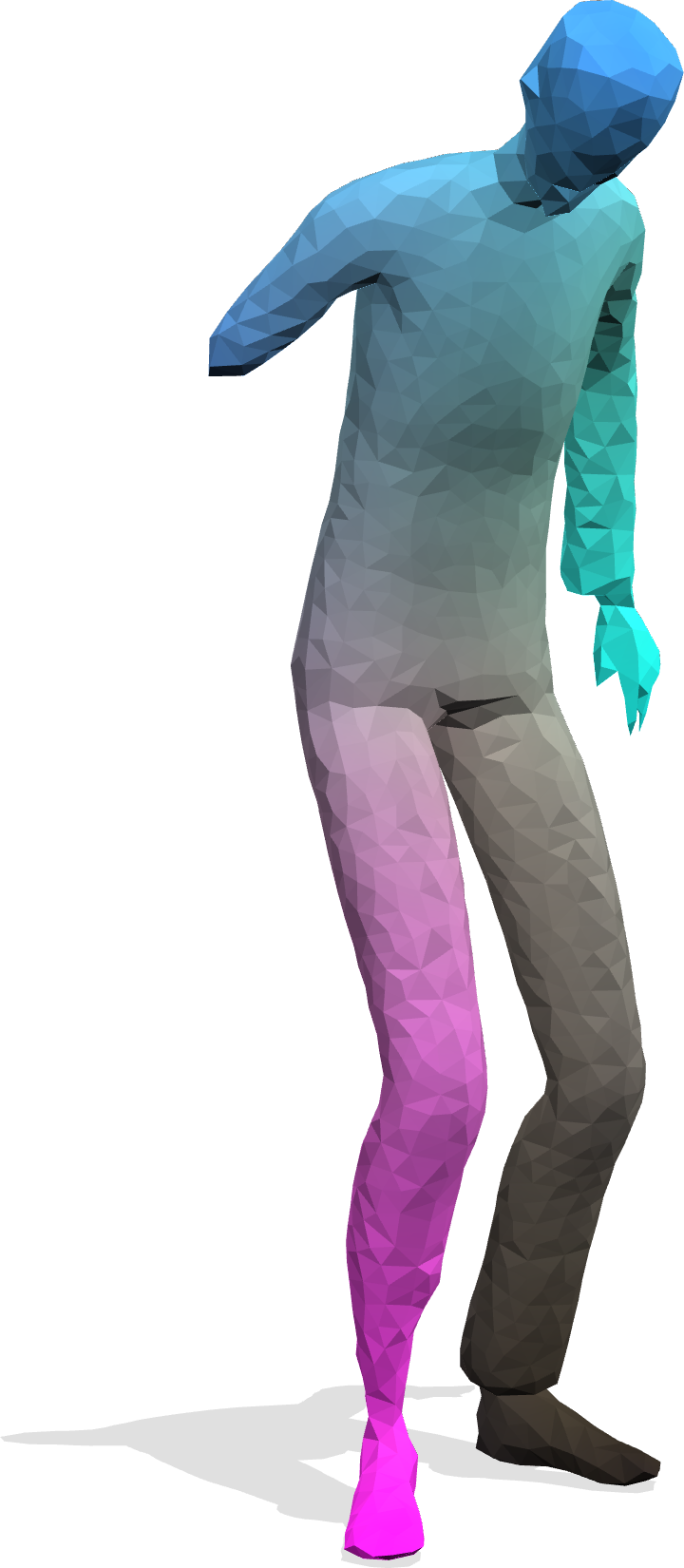}&
    \hspace{\hspaceCols}
    \includegraphics[height=\heightQ, width=\widthQ]{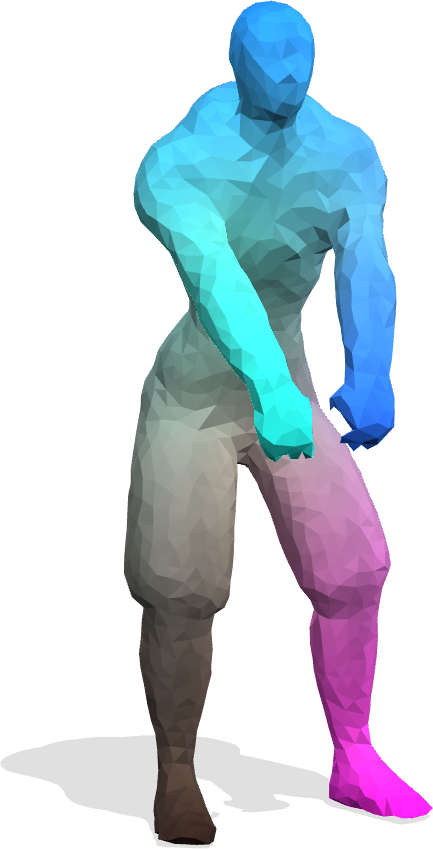}\\
    \rotatedCentering{90}{\heightQ}{\caoetal}&
    \hspace{\hspaceCols}
    \includegraphics[height=\heightQ, width=\widthQ]{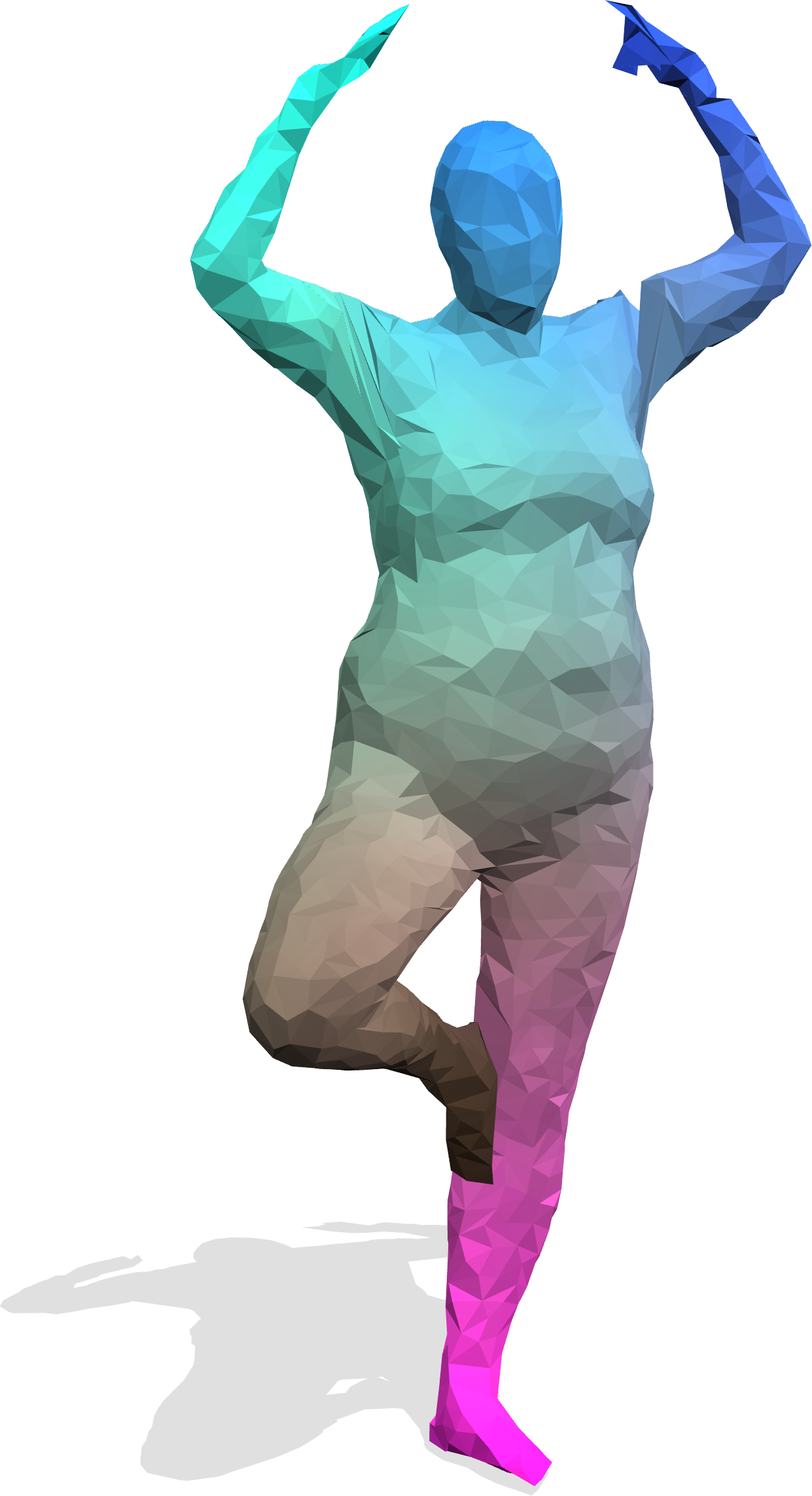}&
    \hspace{\hspaceCols}
    \includegraphics[height=\heightQ, width=\widthQ]{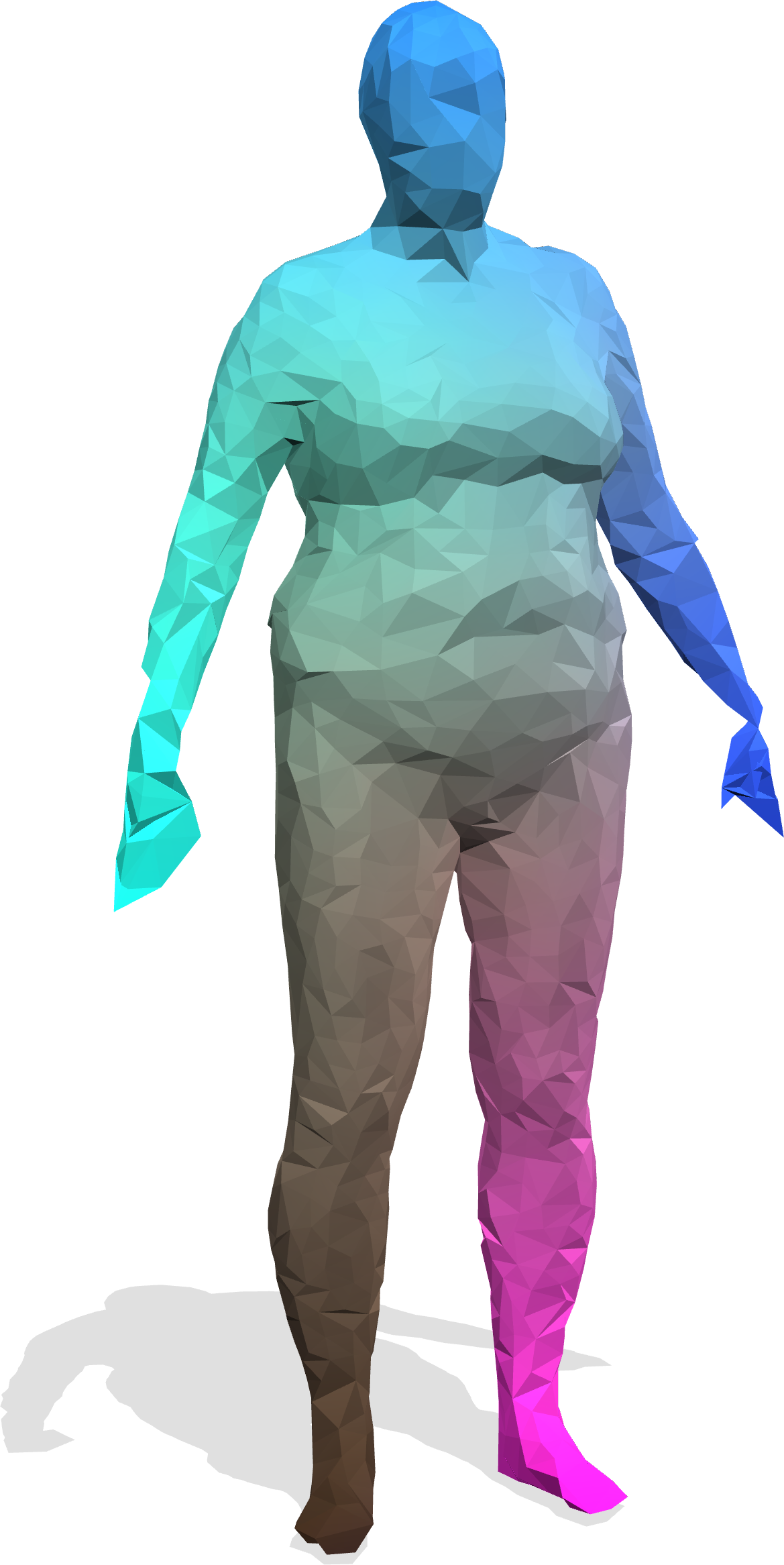}&
    \hspace{\hspaceCols}
    \includegraphics[height=\heightQ, width=\widthQ]{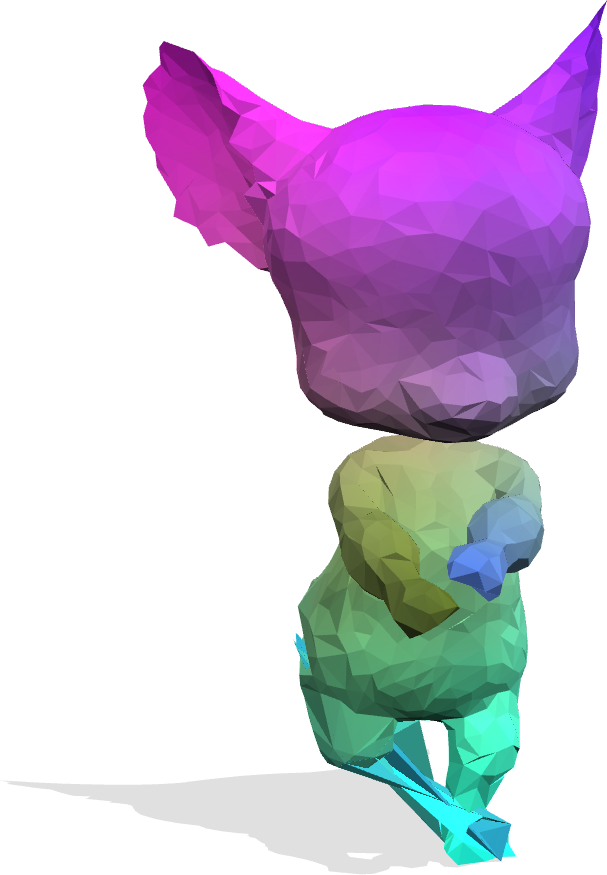}&
    \hspace{\hspaceCols}
    \includegraphics[height=\heightQ, width=\widthQ]{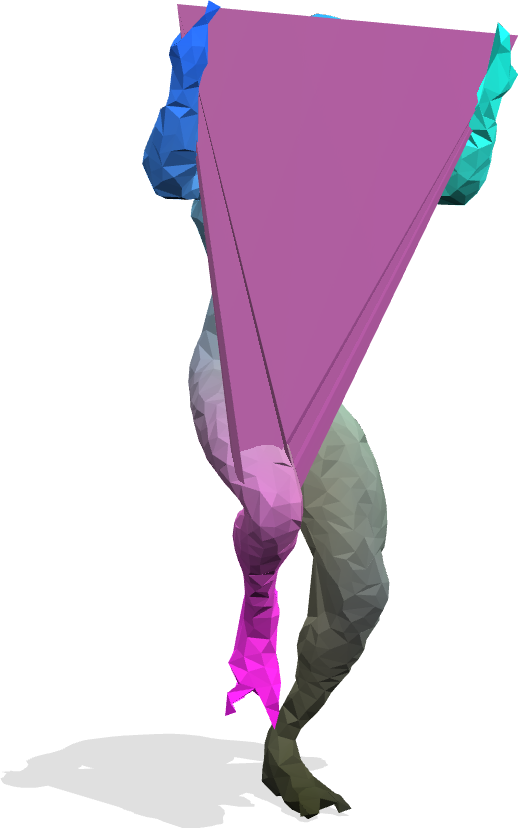}&
    \hspace{\hspaceCols}
    \includegraphics[height=\heightQ, width=\widthQ]{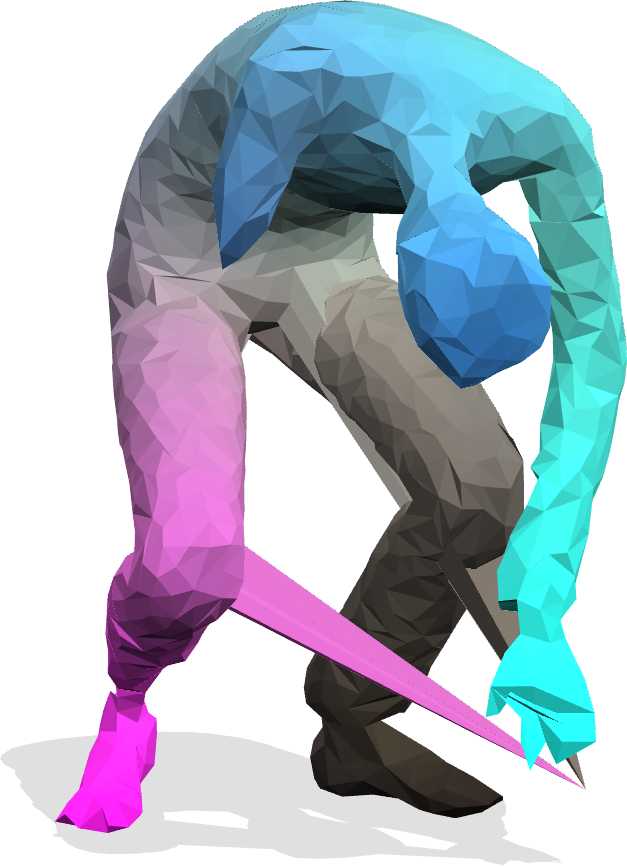}&
    \hspace{\hspaceCols}
    \includegraphics[height=\heightQ, width=\widthQ]{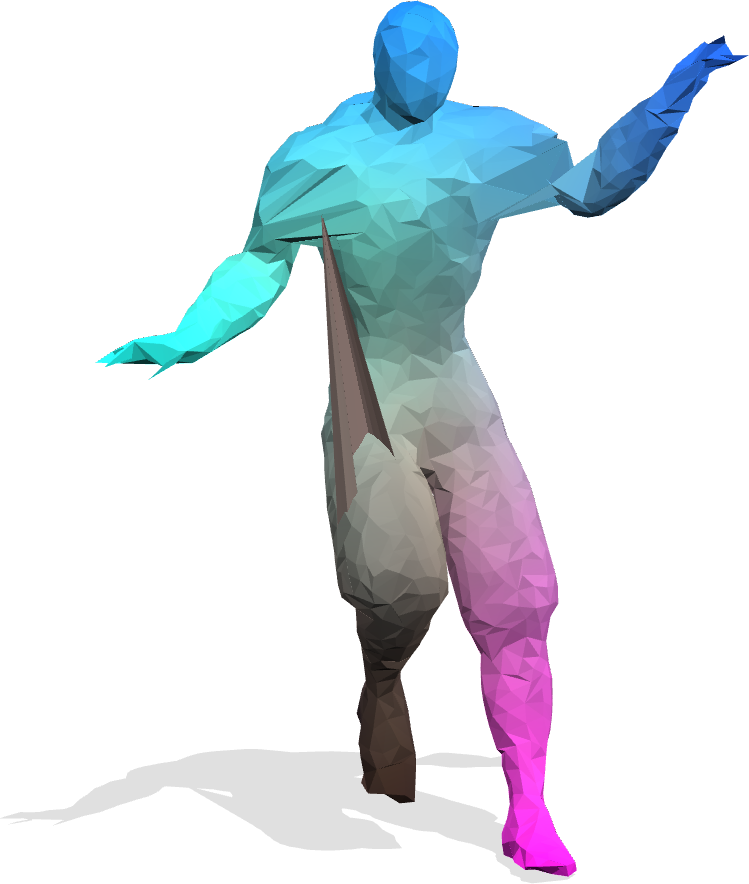}\\
    \rotatedCentering{90}{\heightQ}{\ours}&
    \hspace{\hspaceCols}
    \includegraphics[height=\heightQ, width=\widthQ]{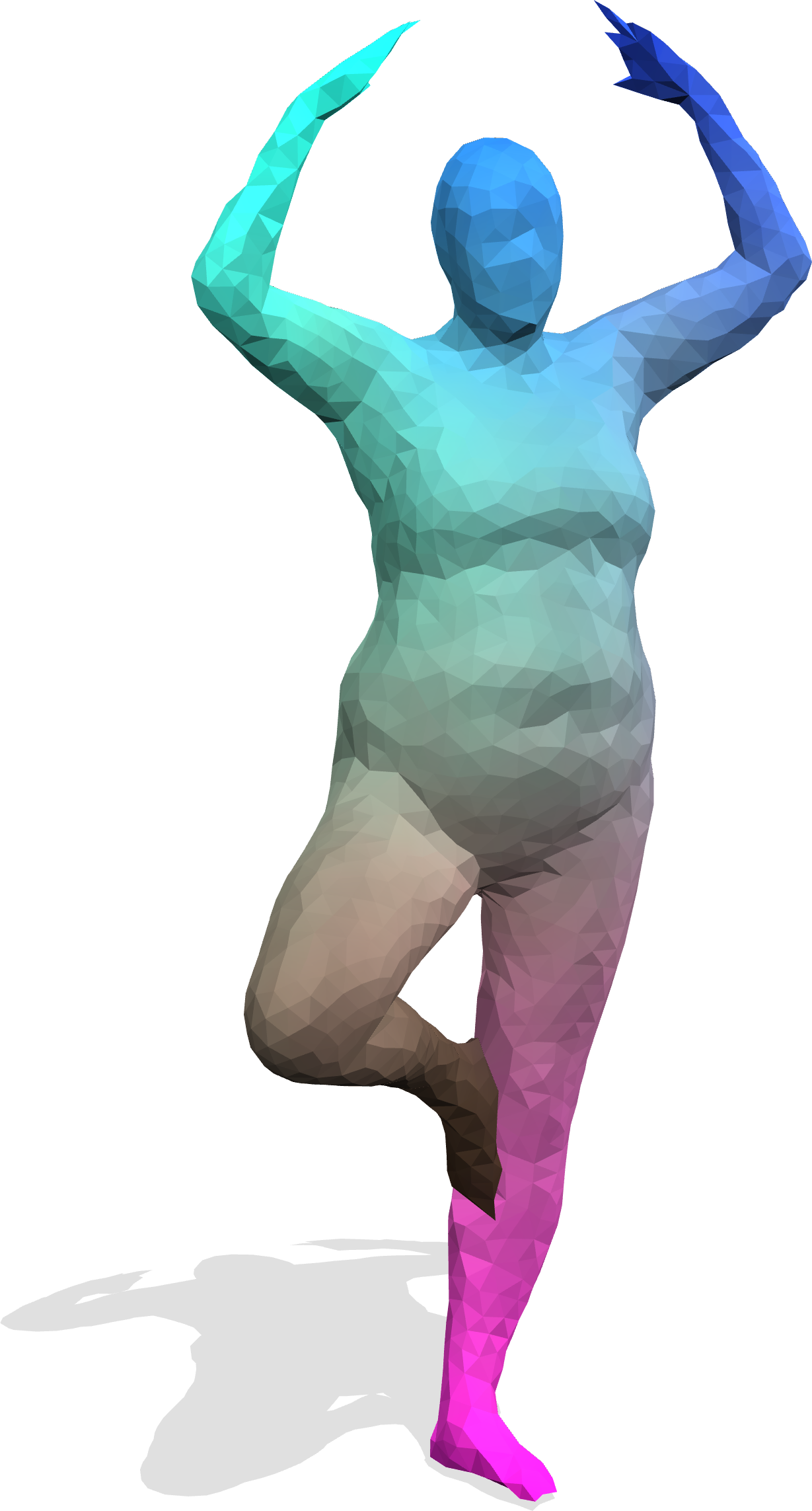}&
    \hspace{\hspaceCols}
    \includegraphics[height=\heightQ, width=\widthQ]{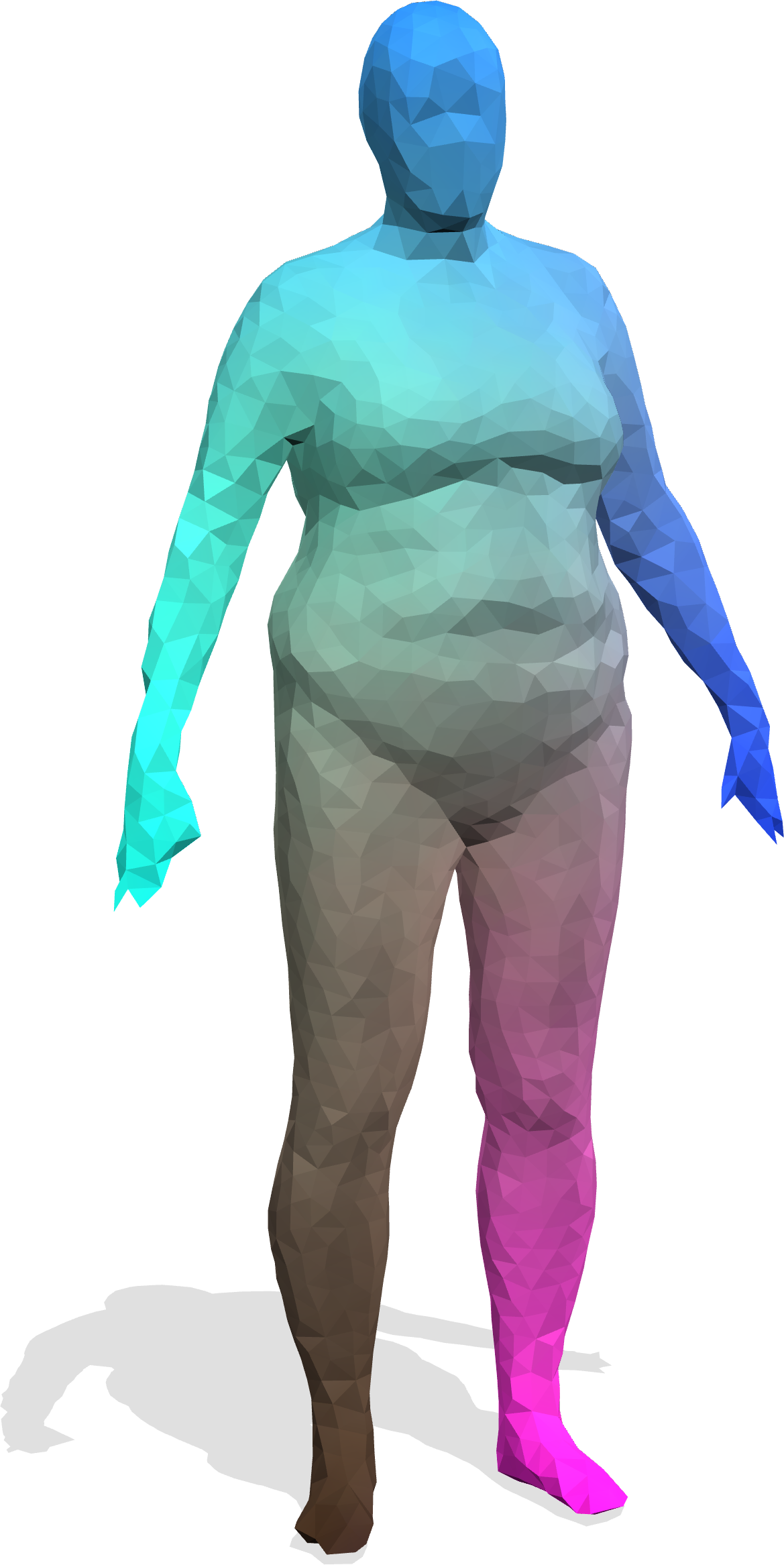}&
    \hspace{\hspaceCols}
    \includegraphics[height=\heightQ, width=\widthQ]{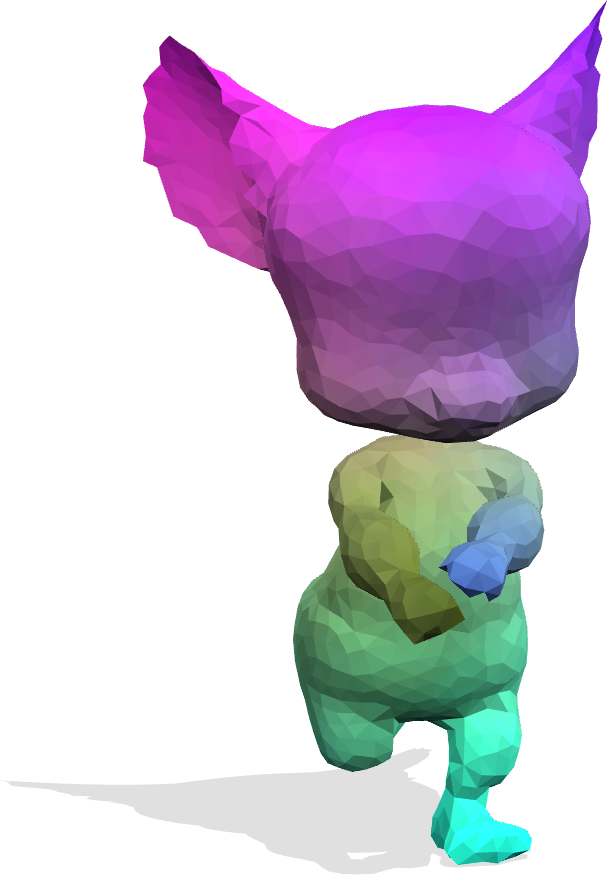}&
    \hspace{\hspaceCols}
    \includegraphics[height=\heightQ, width=\widthQ]{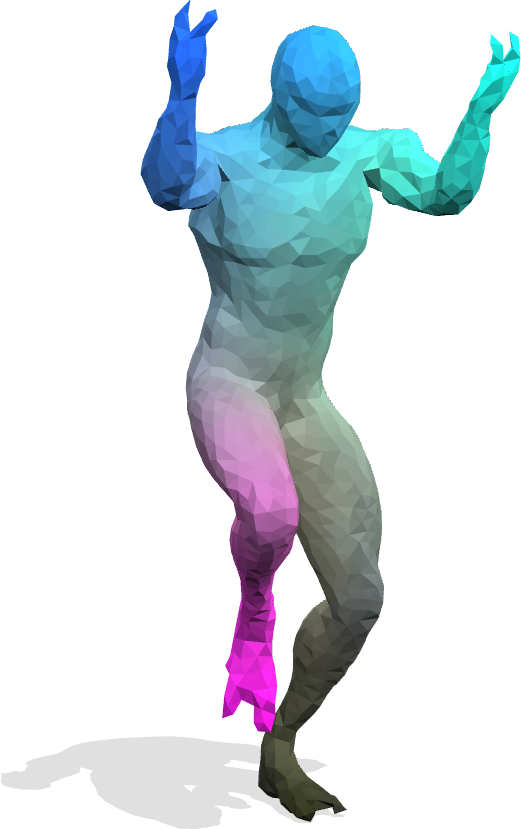}&
    \hspace{\hspaceCols}
    \includegraphics[height=\heightQ, width=\widthQ]{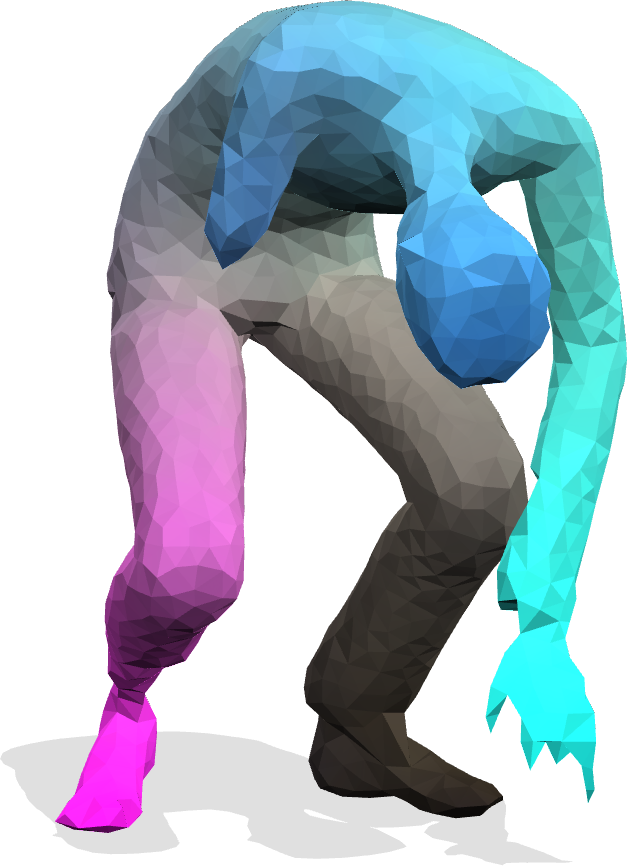}&
    \hspace{\hspaceCols}
    \includegraphics[height=\heightQ, width=\widthQ]{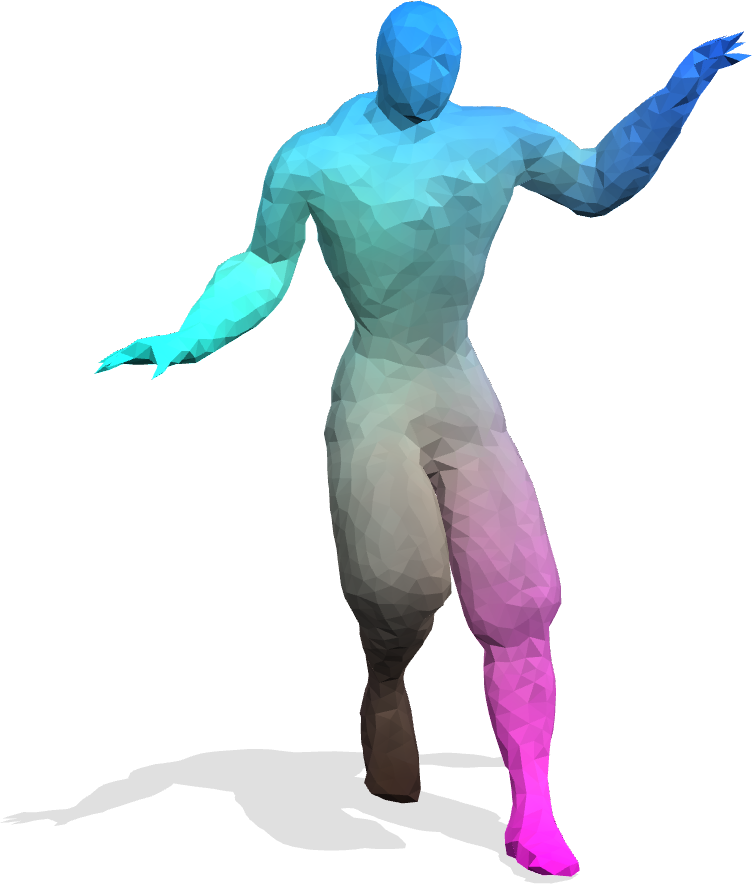}
\end{tabular}
    \end{tabular}
    \vspace{-0.25cm}
    \caption{\textit{Left:} Illustration of our coarse-to-fine pruning strategy. For two matched vertices on the coarse level ({\color{cBLUE} \rule[0.4mm]{0.3cm}{0.9mm}}) we only keep potential matchings on the fine level which belong to one-ring ({\color{cPLOT5} $\bullet$}) and two-ring neighbourhood ({\color{cPINK2} $\bullet$}) of respective vertex on coarse level. 
    \textit{Right:} Qualitative matching comparison via triangulation transfer on FAUST (\scircled{1}-\scircled{2}) and DT4D-Intra (\scircled{3}-\scircled{6}) datasets for shapes with $4k$ triangles. 
    }
    \label{fig:c2f}
\end{figure}

As a proof of concept we evaluate our approach on larger shapes resolutions by employing a coarse-to-fine strategy inspired by~\cite{windheuser2011large}. 
We use solutions from coarser shape resolutions to prune~\eqref{eq:ilp-sm} on finer ones. 
We start with a resolution of $500$ triangles and prune the subsequent ILPs with underlying shape resolutions of $1k$, $2k$, $3k$ up to $4k$ triangles. For pruning the ILPs we only allow matching between triangles which contain vertices of one-ring and two-ring neighbourhood of a given vertex-vertex matching on the coarser resolution (\cref{fig:c2f}).

We evaluate on datasets of FAUST, DT4D-Intra (by subsampling shapes to $4000$ triangles) with resulting mean geodesic errors of {0.0193/0.0161} for ours and {0.0197/0.0165} for \caoetal~resp. Note that scale of the geodesic error decreases due to lesser discretisation noise on higher resolutions. 
The average computation time of the whole coarse-to-fine pipeline is around 20 minutes
from which approximately $30\%$ was spent on dual optimisation. \rebuttalfix{While \caoetal~requires only few seconds however, it does not provide geometrically consistent matchings}.
In \cref{fig:c2f} (right), we show the corresponding qualitative results. Similar to previous results, our approach leads to better matching quality as compared to~\caoetal.

\section{Discussion \& Limitations}
Overall, our proposed solver empirically leads to favourable results compared to competing methods: we obtain the largest portion of globally optimal solutions ($90\%$), the smallest amount of infeasible solutions ($2\%$, cf.~\cref{tab:opt-inf}), and our solver is up to ${11}\times$ faster compared to previous approaches (\cref{fig:convergence}). Despite these positive aspects, there is also a shortcoming: we build upon the product space formalism~\cite{windheuser2011geometrically} and also inherit its quadratic space complexity (w.r.t. number of triangles). We found that for our GPU-based solver the memory of modern graphics cards constitutes the main bottleneck when scaling to higher shape resolutions. With a 48GB GPU, we can solve shape matching problems with up to 1000 triangles per shape, which is an order of magnitude smaller compared to purely learning-based shape matching methods (e.g.,~\cite{cao2023unsupervised}). 
As a proof of concept to circumvent the memory bottleneck, we employ a coarse-to-fine scheme allowing us to compute geometrically consistent matchings of shapes with larger resolutions (4000 triangles).
Yet, our runtimes for the coarse-to-fine scheme (20 minutes) are still higher than the ones of deep learning methods (few seconds).
Nevertheless, we consider our approach to be a large leap forward, especially because our matchings ensure geometric consistency.

\section{Conclusion}
Our work bridges the gap between learning-based shape matching and purely combinatorial optimisation formulations. While learning-based methods are fast and robust, they do not allow to enforce solutions with certain structural or geometric properties. In contrast, optimisation-based approaches allow for taking such properties into account, but they are generally slower and lead to challenging non-convex problems. To address these limitations, we pursue a two-fold strategy: (i) infuse learning-based feature embeddings into the cost function of a combinatorial problem, and 
(ii) propose strategies for better optimisation of the Lagrangian dual~\eqref{eq:dual-problem} by quasi-Newton updates and effectively utilising parallelism in modern GPUs. 
This allows us to solve practically relevant 3D shape matching problems faster as compared to previous approaches, often even to global optimality. We believe that on the one hand our approach is an important contribution for the field of 3D shape analysis per-se. On the other hand, many downstream applications in visual computing that build upon 3D correspondence problems (e.g. autonomous driving and medical image analysis) can benefit from geometrically consistent matchings. 
Lastly, due to its generic nature our approach can directly benefit from future work in more powerful feature extractors.

{
\bibliographystyle{splncs04}
\bibliography{bibliography}

\begin{thebibliography}{10}
\providecommand{\url}[1]{\texttt{#1}}
\providecommand{\urlprefix}{URL }
\providecommand{\doi}[1]{https://doi.org/#1}

\bibitem{abbas2022fastdog}
Abbas, A., Swoboda, P.: {FastDOG: Fast discrete optimization on GPU}. In: CVPR
  (2022)

\bibitem{attaiki2023understanding}
Attaiki, S., Ovsjanikov, M.: Understanding and improving features learned in
  deep functional maps. In: CVPR (2023)

\bibitem{attaiki2021dpfm}
Attaiki, S., Pai, G., Ovsjanikov, M.: {DPFM: Deep partial functional maps}. In:
  2021 International Conference on 3D Vision (3DV). pp. 175--185. IEEE (2021)

\bibitem{attene2010lightweight}
Attene, M.: A lightweight approach to repairing digitized polygon meshes. The
  visual computer  \textbf{26},  1393--1406 (2010)

\bibitem{bernard2020mina}
Bernard, F., Suri, Z.K., Theobalt, C.: {MINA: Convex mixed-integer programming
  for non-rigid shape alignment}. In: CVPR (2020)

\bibitem{bogo2014faust}
Bogo, F., Romero, J., Loper, M., Black, M.J.: {FAUST: Dataset and evaluation
  for 3D mesh registration}. In: CVPR (2014)

\bibitem{bryant1986graph}
Bryant, R.E.: Graph-based algorithms for boolean function manipulation.
  Computers, IEEE Transactions on  \textbf{100}(8),  677--691 (1986)

\bibitem{cao2022unsupervised}
Cao, D., Bernard, F.: Unsupervised deep multi-shape matching. In: ECCV (2022)

\bibitem{cao2023self}
Cao, D., Bernard, F.: {Self-Supervised Learning for Multimodal Non-Rigid 3D
  Shape Matching}. In: CVPR (2023)

\bibitem{cao2024spectral}
Cao, D., Eisenberger, M., El~Amrani, N., Cremers, D., Bernard, F.: Spectral
  meets spatial: Harmonising 3d shape matching and interpolation. In: CVPR
  (2024)

\bibitem{cao2023unsupervised}
Cao, D., Roetzer, P., Bernard, F.: Unsupervised learning of robust spectral
  shape matching. ACM Transactions on Graphics (TOG)  (2023)

\bibitem{coughlan2000efficient}
Coughlan, J., Yuille, A., English, C., Snow, D.: Efficient deformable template
  detection and localization without user initialization. Computer Vision and
  Image Understanding  \textbf{78}(3),  303--319 (2000)

\bibitem{donati2022deep}
Donati, N., Corman, E., Ovsjanikov, M.: Deep orientation-aware functional maps:
  Tackling symmetry issues in shape matching. In: CVPR (2022)

\bibitem{donati2020deep}
Donati, N., Sharma, A., Ovsjanikov, M.: Deep geometric functional maps: Robust
  feature learning for shape correspondence. In: CVPR (2020)

\bibitem{dyke2020track}
Dyke, R.M., Lai, Y.K., Rosin, P.L., Zappal{\`a}, S., Dykes, S., Guo, D., Li,
  K., Marin, R., Melzi, S., Yang, J.: {SHREC'20}: Shape correspondence with
  non-isometric deformations. Computers \& Graphics  \textbf{92},  28--43
  (2020)

\bibitem{ezuz2017deblurring}
Ezuz, D., Ben-Chen, M.: Deblurring and denoising of maps between shapes. In:
  Computer Graphics Forum. vol.~36, pp. 165--174. Wiley Online Library (2017)

\bibitem{ezuz2019elastic}
Ezuz, D., Heeren, B., Azencot, O., Rumpf, M., Ben-Chen, M.: Elastic
  correspondence between triangle meshes. In: Computer Graphics Forum. vol.~38,
  pp. 121--134. Wiley Online Library (2019)

\bibitem{felzenszwalb2005representation}
Felzenszwalb, P.F.: Representation and detection of deformable shapes. TPAMI
  \textbf{27}(2),  208--220 (2005)

\bibitem{gasparetto2017spatial}
Gasparetto, A., Cosmo, L., Rodola, E., Bronstein, M., Torsello, A.: {Spatial
  maps: From low rank spectral to sparse spatial functional representations}.
  In: 2017 International Conference on 3D Vision (3DV). pp. 477--485. IEEE
  (2017)

\bibitem{groueix20183d}
Groueix, T., Fisher, M., Kim, V.G., Russell, B.C., Aubry, M.: 3d-coded: 3d
  correspondences by deep deformation. In: European Conference on Computer
  Vision (2018)

\bibitem{gurobi}
{Gurobi Optimization, LLC}: {Gurobi Optimizer Reference Manual} (2023),
  \url{https://www.gurobi.com}

\bibitem{halimi2019unsupervised}
Halimi, O., Litany, O., Rodola, E., Bronstein, A.M., Kimmel, R.: Unsupervised
  learning of dense shape correspondence. In: CVPR (2019)

\bibitem{hormann2000mips}
Hormann, K., Greiner, G.: {MIPS: An efficient global parametrization method}.
  Curve and Surface Design: Saint-Malo 1999 pp. 153--162 (2000)

\bibitem{huang2013consistent}
Huang, Q.X., Guibas, L.: Consistent shape maps via semidefinite programming.
  In: Computer graphics forum. vol.~32, pp. 177--186. Wiley Online Library
  (2013)

\bibitem{libigl}
Jacobson, A., Panozzo, D., et~al.: {libigl}: A simple {C++} geometry processing
  library (2018), https://libigl.github.io/

\bibitem{jiang2023neural}
Jiang, P., Sun, M., Huang, R.: {Neural Intrinsic Embedding for Non-rigid Point
  Cloud Matching}. In: CVPR (2023)

\bibitem{kim2011blended}
Kim, V.G., Lipman, Y., Funkhouser, T.: Blended intrinsic maps. ACM transactions
  on graphics (TOG)  \textbf{30}(4),  1--12 (2011)

\bibitem{taocp4a}
Knuth, D.E.: The Art of Computer Programming: Combinatorial Algorithms, Part 1.
  Addison-Wesley Professional, 1st edn. (2011)

\bibitem{lahner2016efficient}
L\"{a}hner, Z., Rodola, E., Schmidt, F.R., Bronstein, M.M., Cremers, D.:
  Efficient globally optimal 2d-to-3d deformable shape matching. In: CVPR
  (2016)

\bibitem{lange2021efficient}
Lange, J.H., Swoboda, P.: Efficient message passing for 0--1 ilps with binary
  decision diagrams. In: International Conference on Machine Learning. pp.
  6000--6010. PMLR (2021)

\bibitem{li2022learning}
Li, L., Donati, N., Ovsjanikov, M.: Learning multi-resolution functional maps
  with spectral attention for robust shape matching. NIPS  (2022)

\bibitem{li2020shape}
Li, Q., Liu, S., Hu, L., Liu, X.: {Shape correspondence using anisotropic
  Chebyshev spectral CNNs}. In: CVPR (2020)

\bibitem{li20214dcomplete}
Li, Y., Takehara, H., Taketomi, T., Zheng, B., Nie{\ss}ner, M.: 4dcomplete:
  Non-rigid motion estimation beyond the observable surface. In: ICCV (2021)

\bibitem{litke2005image}
Litke, N., Droske, M., Rumpf, M., Schr{\"o}der, P.: An image processing
  approach to surface matching. In: Symposium on Geometry Processing. vol.~255,
  pp. 207--216 (2005)

\bibitem{liu1989lbfgs}
Liu, D.C., Nocedal, J.: {On the limited memory BFGS method for large scale
  optimization}. Mathematical programming  \textbf{45}(1-3),  503--528 (1989)

\bibitem{magnet2022smooth}
Magnet, R., Ren, J., Sorkine-Hornung, O., Ovsjanikov, M.: Smooth non-rigid
  shape matching via effective dirichlet energy optimization. In: International
  Conference on 3D Vision (3DV) (2022)

\bibitem{marin2020correspondence}
Marin, R., Rakotosaona, M.J., Melzi, S., Ovsjanikov, M.: Correspondence
  learning via linearly-invariant embedding. In: NeurIPS (2020)

\bibitem{maron2016point}
Maron, H., Dym, N., Kezurer, I., Kovalsky, S., Lipman, Y.: Point registration
  via efficient convex relaxation. ACM Transactions on Graphics (TOG)
  \textbf{35}(4), ~73 (2016)

\bibitem{meyer2003discrete}
Meyer, M., Desbrun, M., Schr{\"o}der, P., Barr, A.H.: Discrete
  differential-geometry operators for triangulated 2-manifolds. In:
  Visualization and mathematics III. pp. 35--57. Springer (2003)

\bibitem{nocedal1980updating_lbfgs}
Nocedal, J.: Updating quasi-newton matrices with limited storage. Mathematics
  of computation  \textbf{35}(151),  773--782 (1980)

\bibitem{ovsjanikov2012functional}
Ovsjanikov, M., Ben-Chen, M., Solomon, J., Butscher, A., Guibas, L.: Functional
  maps: a flexible representation of maps between shapes. ACM Transactions on
  Graphics (TOG)  \textbf{31}(4), ~30 (2012)

\bibitem{qi2017pointnet++}
Qi, C.R., Yi, L., Su, H., Guibas, L.J.: Pointnet++: Deep hierarchical feature
  learning on point sets in a metric space. NIPS  (2017)

\bibitem{ren2018continuous}
Ren, J., Poulenard, A., Wonka, P., Ovsjanikov, M.: Continuous and
  orientation-preserving correspondences via functional maps. ACM Transactions
  on Graphics (ToG)  \textbf{37},  1--16 (2018)

\bibitem{roetzer2023conjugate}
Roetzer, P., L{\"a}hner, Z., Bernard, F.: Conjugate product graphs for globally
  optimal 2d-3d shape matching. In: IEEE Conf. on Computer Vision and Pattern
  Recognition (CVPR) (2023)

\bibitem{roetzer2022scalable}
Roetzer, P., Swoboda, P., Cremers, D., Bernard, F.: A scalable combinatorial
  solver for elastic geometrically consistent 3d shape matching. In: CVPR
  (2022)

\bibitem{roufosse2019unsupervised}
Roufosse, J.M., Sharma, A., Ovsjanikov, M.: Unsupervised deep learning for
  structured shape matching. In: ICCV (2019)

\bibitem{sahilliouglu2018genetic}
Sahillio{\u{g}}lu, Y.: A genetic isometric shape correspondence algorithm with
  adaptive sampling. ACM Transactions on Graphics (ToG)  \textbf{37}(5),  1--14
  (2018)

\bibitem{sahilliouglu2020recent}
Sahillio{\u{g}}lu, Y.: Recent advances in shape correspondence. The Visual
  Computer  \textbf{36}(8),  1705--1721 (2020)

\bibitem{sankoff1983time}
Sankoff, D., Kruskal, J.B.: Time warps, string edits, and macromolecules: the
  theory and practice of sequence comparison. Addison-Wesley Publishing Company
  (1983)

\bibitem{schmidt2009planar}
Schmidt, F.R., T\"oppe, E., Cremers, D.: Efficient planar graph cuts with
  applications in computer vision. In: CVPR (2009)

\bibitem{schmidt2020inter}
Schmidt, P., Campen, M., Born, J., Kobbelt, L.: Inter-surface maps via
  constant-curvature metrics. ACM Transactions on Graphics (TOG)
  \textbf{39}(4),  119--1 (2020)

\bibitem{schmidt2023surface}
Schmidt, P., Pieper, D., Kobbelt, L.: Surface maps via adaptive triangulations.
  Computer Graphics Forum  \textbf{42}(2) (2023)

\bibitem{schoenemann2009combinatorial}
Schoenemann, T., Cremers, D.: A combinatorial solution for model-based image
  segmentation and real-time tracking. TPAMI  \textbf{32}(7),  1153--1164
  (2009)

\bibitem{schreiner2004inter}
Schreiner, J., Asirvatham, A., Praun, E., Hoppe, H.: Inter-surface mapping. In:
  ACM SIGGRAPH 2004 Papers, pp. 870--877 (2004)

\bibitem{sharma2020weakly}
Sharma, A., Ovsjanikov, M.: Weakly supervised deep functional maps for shape
  matching. NIPS  (2020)

\bibitem{sharma2011topologically}
Sharma, A., Horaud, R., Cech, J., Boyer, E.: Topologically-robust 3d shape
  matching based on diffusion geometry and seed growing. In: CVPR 2011. pp.
  2481--2488. IEEE (2011)

\bibitem{sharp2020diffusionnet}
Sharp, N., Attaiki, S., Crane, K., Ovsjanikov, M.: Diffusionnet: Discretization
  agnostic learning on surfaces. ACM Transactions on Graphics (TOG)  (2022)

\bibitem{solomon2016entropic}
Solomon, J., Peyr{\'e}, G., Kim, V.G., Sra, S.: Entropic metric alignment for
  correspondence problems. ACM Transactions on Graphics (ToG)  \textbf{35}(4),
  1--13 (2016)

\bibitem{sullivan2019pyvista}
Sullivan, C.B., Kaszynski, A.: {PyVista}: 3d plotting and mesh analysis through
  a streamlined interface for the visualization toolkit ({VTK}). Journal of
  Open Source Software  \textbf{4}(37), ~1450 (may 2019).
  \doi{10.21105/joss.01450}, \url{https://doi.org/10.21105/joss.01450}

\bibitem{takayama2022compatible}
Takayama, K.: Compatible intrinsic triangulations. ACM Transactions on Graphics
  (TOG)  \textbf{41}(4),  1--12 (2022)

\bibitem{tam2012registration}
Tam, G.K., Cheng, Z.Q., Lai, Y.K., Langbein, F.C., Liu, Y., Marshall, D.,
  Martin, R.R., Sun, X.F., Rosin, P.L.: Registration of 3d point clouds and
  meshes: A survey from rigid to nonrigid. IEEE transactions on visualization
  and computer graphics  \textbf{19}(7),  1199--1217 (2012)

\bibitem{thomas2019kpconv}
Thomas, H., Qi, C.R., Deschaud, J.E., Marcotegui, B., Goulette, F., Guibas,
  L.J.: Kpconv: Flexible and deformable convolution for point clouds. In: ICCV
  (2019)

\bibitem{trappolini2021shape}
Trappolini, G., Cosmo, L., Moschella, L., Marin, R., Melzi, S., Rodol{\`a}, E.:
  Shape registration in the time of transformers. Advances in Neural
  Information Processing Systems  (2021)

\bibitem{van2011survey}
Van~Kaick, O., Zhang, H., Hamarneh, G., Cohen-Or, D.: A survey on shape
  correspondence. In: Computer graphics forum. vol.~30, pp. 1681--1707. Wiley
  Online Library (2011)

\bibitem{vestner2017efficient}
Vestner, M., L{\"a}hner, Z., Boyarski, A., Litany, O., Slossberg, R., Remez,
  T., Rodola, E., Bronstein, A., Bronstein, M., Kimmel, R., et~al.: Efficient
  deformable shape correspondence via kernel matching. In: 2017 international
  conference on 3D vision (3DV). pp. 517--526. IEEE (2017)

\bibitem{vestner2017product}
Vestner, M., Litman, R., Rodola, E., Bronstein, A., Cremers, D.: {Product
  manifold filter: Non-rigid shape correspondence via kernel density estimation
  in the product space}. In: Proceedings of the IEEE Conference on Computer
  Vision and Pattern Recognition. pp. 3327--3336 (2017)

\bibitem{wiersma2020cnns}
Wiersma, R., Eisemann, E., Hildebrandt, K.: Cnns on surfaces using
  rotation-equivariant features. ACM Transactions on Graphics (ToG)
  \textbf{39}(4),  92--1 (2020)

\bibitem{windheuser2011geometrically}
Windheuser, T., Schlickewei, U., Schmidt, F.R., Cremers, D.: Geometrically
  consistent elastic matching of 3d shapes: A linear programming solution. In:
  ICCV (2011)

\bibitem{windheuser2011large}
Windheuser, T., Schlickwei, U., Schimdt, F.R., Cremers, D.: Large-scale integer
  linear programming for orientation preserving 3d shape matching. In: Computer
  Graphics Forum. vol.~30, pp. 1471--1480. Wiley Online Library (2011)

\bibitem{zuffi20173d}
Zuffi, S., Kanazawa, A., Jacobs, D.W., Black, M.J.: {3D menagerie: Modeling the
  3D shape and pose of animals}. In: CVPR (2017)

\end{thebibliography}

}
\clearpage
\appendix
\counterwithin{figure}{section}
\counterwithin{table}{section}
\section{Fast Discrete Optimisation for Geometrically Consistent 3D Shape Matching - Appendix}

\subsection{Evaluation on other ILP benchmark problems}
\label{sec:evaluation_other_ilps}
Although in this work we focus mainly on shape matching problems, our quasi-Newton solver is generally applicable in all cases which~\cite{abbas2022fastdog} can solve, i.e., $0$--$1$ integer linear programs. 
Moreover, our constraint splitting scheme is also generally applicable however, it is useful only in certain scenarios, i.e., a few constraints containing a larger amount of variables as compared to other constraints. 

\begin{figure}[htb!]
    \centering
    \begin{tabular}{cc}
        \hspace{-1cm}
        \includegraphics[width=0.45\textwidth]{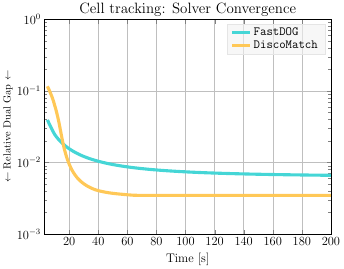}&
        \includegraphics[width=0.45\textwidth]{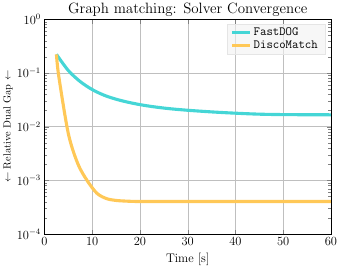}
    \end{tabular}%
    \caption{{Convergence comparison} of our quasi-Newton based solver via Alg.~\ref{alg:lbfgs} versus \texttt{FastDOG}~\cite{abbas2022fastdog} on problems unrelated to shape matching. We report results on the datasets which are most difficult to solve in~\cite{abbas2022fastdog} i.e., the largest instance of cell tracking problem (left) and $20$ instances of graph matching for developmental biology (right). In both datasets we reach better objectives of the Lagrangean relaxation~\eqref{eq:dual-problem} measured via gap to the optimal solution (lower values are better). 
    }
    \label{fig:convergence_comparison_other_ilps}
\end{figure}

To check efficacy of our contribution we evaluate our solver on the most difficult instances of cell tracking and graph matching datasets as used in~\cite{abbas2022fastdog}. We only utilise Alg.~\ref{alg:lbfgs} in our solver and do not use our constraint splitting scheme since both datasets contain constraints of roughly equal size. The results are reported in Figure~\ref{fig:convergence_comparison_other_ilps}. 

We observe that on both datasets our solver improves upon~\cite{abbas2022fastdog} by reaching better objectives as measured by gap to the optimum. It also matches the performance of~\cite{abbas2022fastdog} in significantly less time.

\subsection{Additional Analysis}\label{sec:appendix-dataset-analysis}
In \cref{tab:opt-inf}, we show number of infeasible and optimal solutions per dataset.
We observe that our method produces the most of amount of optimal and the least amount of infeasible solutions on all datasets.
\begin{table*}[!htb]
    \small
    \centering
    \resizebox{\textwidth}{!}{
    \begin{tabular}{@{}lcccccccccc|cc@{}}
    \toprule
    Dataset (\#Inst.)& \multicolumn{2}{c}{\textbf{FAUST}(190)} & \multicolumn{2}{c}{\textbf{DT4D-Intra}(100)} & \multicolumn{2}{c}{\textbf{DT4D-Inter}(100)} & \multicolumn{2}{c}{\textbf{SMAL}(136)} & \multicolumn{2}{c}{\textbf{SHREC'20}(18)} & \multicolumn{2}{c}{\textbf{Total}(544)}\\
     \cmidrule(l){2-3} \cmidrule(l){4-5} \cmidrule(l){6-7} \cmidrule(l){8-9} \cmidrule(l){10-11} 
     & \# Opt. & \# Inf. & \# Opt. & \# Inf. &  \# Opt. & \# Inf. &  \# Opt. & \# Inf. &  \# Opt. & \# Inf. &\# Opt. & \# Inf.  \\ \midrule
     {\smcomb}& 149& 9& 82& 2& 68& 11& 101& 6&  7& 2 & 407& 30\\
     {\fastdog}& 170& 4& 93& 1& 80& 12& 119& 3&  9& \textbf{0} &471&20\\
     {\ours}   & \textbf{174}& \textbf{2}& \textbf{97}& \textbf{0}& \textbf{81}&  \textbf{9}& \textbf{128}& \textbf{0}& \textbf{12}& \textbf{0} &\textbf{492}&\textbf{11}\\
     \bottomrule
    \end{tabular}}
    \caption{We compare number of \textbf{global optimal} ($\uparrow$) and \textbf{infeasible} ($\downarrow$) solutions of the individual ILP solvers per dataset. Our solver consistently yields most global optimal and least infeasible solutions. In total, our approach solves $90\%$ of instances to global optimality (\fastdog\ $87\%$, \smcomb\ $75\%$) and produces only $2\%$ of infeasible solutions (\fastdog\ $4\%$, \smcomb\ $6\%$).}
    \label{tab:opt-inf}
\end{table*}

Moreover in \cref{fig:runtime} we observe that our solver offers much better scaling to growing instances sizes and finds solutions with lower optimality gaps as compared to other ILP solvers. 

\begin{figure}[!htb]
    \centering
    \begin{tabular}{c|c}
        \hspace{-0.8cm}
        \begin{tabular}{cc}
            \multicolumn{2}{c}{\newcommand{\runtimeLineWidth}{3pt}
\newcommand{\rtCplotWidth}{0\columnwidth}
\newcommand{\rtCplotHeight}{0\columnwidth}

\newcommand{\drawcross}[1]{%
\begin{tikzpicture}[#1]%
\node[mark size=4pt, color=cPLOT6, line width=1.5pt] at (0, -1) {\pgfuseplotmark{x}};
\end{tikzpicture}%
}

\newcommand{\drawsquare}[1]{%
\begin{tikzpicture}[#1]%
\node[mark size=4pt, color=cPLOT0, line width=1.5pt] at (0, -1) {\pgfuseplotmark{square}};
\end{tikzpicture}%
}

\newcommand{\drawplus}[1]{%
\begin{tikzpicture}[#1]%
\node[mark size=4pt, color=cPLOT3, line width=1.5pt] at (0, -1) {\pgfuseplotmark{+}};
\end{tikzpicture}%
}

\newcommand{\drawcircle}[1]{%
\begin{tikzpicture}[#1]%
\node[mark size=3pt, color=cPLOT5, line width=1.5pt] at (0, -1) {\pgfuseplotmark{o}};
\end{tikzpicture}%
}

\pgfplotsset{%
	label style = {font=\normalfont},
	tick label style = {font=\normalfont},
	title style =  {font=\normalfont},
}
\begin{tikzpicture}[scale=0.5, transform shape]
	\begin{axis}[%
		width=\rtCplotWidth,
		height=\rtCplotHeight,
		title style={yshift=0.8cm},
		scale only axis,
		grid=major,
           legend style={at={(0.498,1.13)},anchor=south,legend columns=4,font=\large},
		ylabel={{\large Runtime [s]}},
		xlabel={\large Number of Triangles of $\mathcal{M}$ and $\mathcal{N}$},
		xmin=50,
		xmax=950,
		xtick={50, 250, 500, 750, 1000},
		xtick scale label code/.code={},
		ymin=0,
		ymax=750,
		ytick={250, 500, 750},
		ylabel near ticks,
		]

		\addplot [color=cPLOT6, smooth, line width=\runtimeLineWidth]
		table[row sep=crcr]{%
0 0\\
		};
		\addlegendentry{\textcolor{black}{\lprelax\ }\, \drawcross{} $\quad$}

		\addplot [color=cPLOT0, smooth, line width=\runtimeLineWidth]
		table[row sep=crcr]{%
0 0\\
		};
		\addlegendentry{\textcolor{black}{\smcomb\ }\, \drawsquare{} $\quad$}
		
		\addplot [color=cPLOT3, smooth, tension=0.2, line width=\runtimeLineWidth]
		table[row sep=crcr]{%
0 0\\
		};
		\addlegendentry{\textcolor{black}{\fastdog\ }\, \drawplus{} $\quad$}
		
		\addplot [color=cPLOT5, line width=\runtimeLineWidth, smooth]
		table[row sep=crcr]{%
0 0\\
		};
		\addlegendentry{\textcolor{black}{\ours\ }\, \drawcircle{}}
	\end{axis}
\end{tikzpicture}%
}\\
            \input{figs/tikz/runtime_shape_res}&
            \hspace{-0.6cm}
        \input{figs/tikz/runtime_shape_res_DT4d_intra}
        \end{tabular}
        &
        \hspace{-0.7cm}
        \begin{tabular}{c}
            \input{figs/tikz/scatter_faust}
        \end{tabular}
    \end{tabular}%
    \caption{\textit{Left:} {Runtime comparison} w.r.t.\ shape resolution. Thick lines are mean runtimes and markers show individual experiments. Note that the problem size grows quadratically with the number of triangles of the individual shapes, i.e.,~the search space size is approximately $21\cdot |T_\mathcal{M}||T_\mathcal{N}|$.
    \textit{Right:} Our novel combinatorial solver \protect\yellowcirclemarker is faster and finds tighter optimality gaps compared to previous combinatorial approaches \smcomb~\cite{roetzer2022scalable} and \fastdog~\cite{abbas2022fastdog}.
    }
    \label{fig:runtime}
\end{figure}

\begin{figure*}[!htb]
    \hspace{-1.65cm}
    \def\pathOurs{figs/qualitative/ours/}
\def\pathCao{figs/qualitative/caoetal/}
\def\srcEnd{_M.png}
\def\trgtEnd{_N.png}
\def\columnOne{Standing2HMagicAttack01058-Falling202}
\def\columnTwo{KettlebellSwing020-StandingReactLargeFromLeft021}
\def\columnTwoTwo{StandingCoverTurn007-Running046}
\def\columnThree{SwingDancing184-SwingDancing316}
\def\columnFour{Flair027-Flair023}
\def\columnFive{giraffe_a-leopard}
\def\columnSix{leopard-pig}
\def\columnSeven{horse_04-wolf_03}
\def\columnEight{dog_01-dog_08}
\def\columnNine{tr_reg_093-tr_reg_096}
\def\columnTen{tr_reg_082-tr_reg_087}
\def\columnEleven{tr_reg_081-tr_reg_098}
\def\heightQ{1.5cm}
\def\widthQ{1.4cm}
\def\hspaceCols{-0.4cm}
\newcommand{\geoerrorstyle}[1]{{\tiny\textcolor{gray}{#1}}}
\begin{tabular}{cccccccccccc}
    & \scircled{1} &\hspace{\hspaceCols}  \scircled{2} &\hspace{\hspaceCols}  \scircled{3} &\hspace{\hspaceCols}  \scircled{4} &\hspace{\hspaceCols}  \scircled{5} &\hspace{\hspaceCols}  \scircled{6} &\hspace{\hspaceCols}  \scircled{7} & \hspace{\hspaceCols} \scircled{8} &\hspace{\hspaceCols}  \scircled{9} &\hspace{\hspaceCols}  \scircled{10} &\hspace{\hspaceCols}\scircled{11}\\
    \setlength{\tabcolsep}{0pt} 
    \rotatedCentering{90}{\heightQ}{Source}&
    \hspace{0.0cm}
    \includegraphics[height=\heightQ, width=\widthQ]{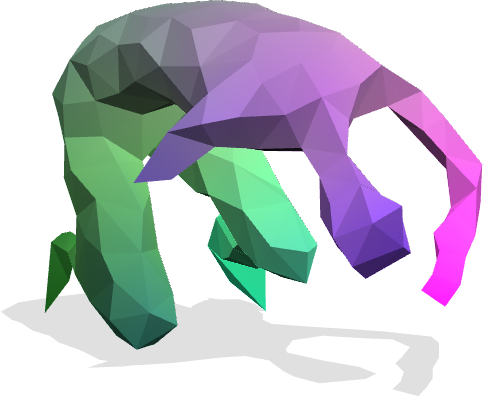}&
    \hspace{\hspaceCols}
    \includegraphics[height=\heightQ, width=\widthQ]{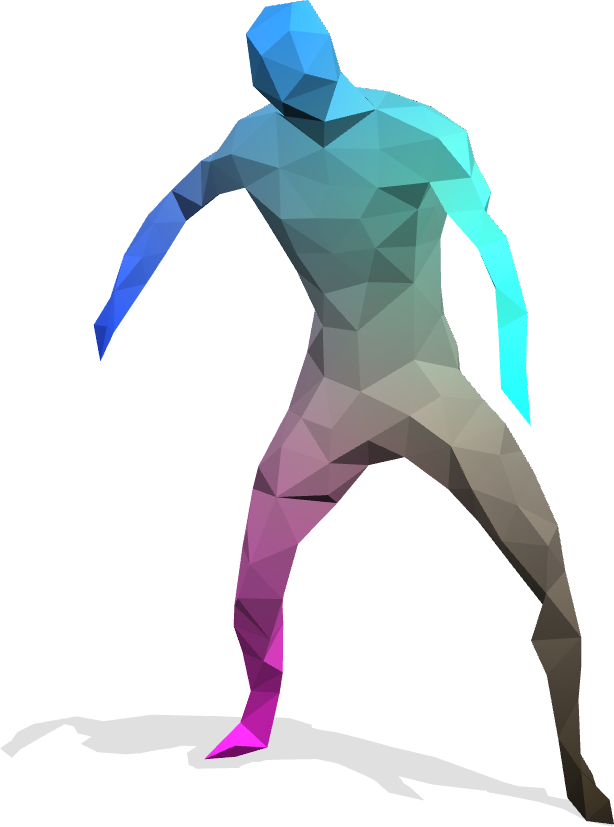}&
    \hspace{\hspaceCols}
    \includegraphics[height=\heightQ, width=\widthQ]{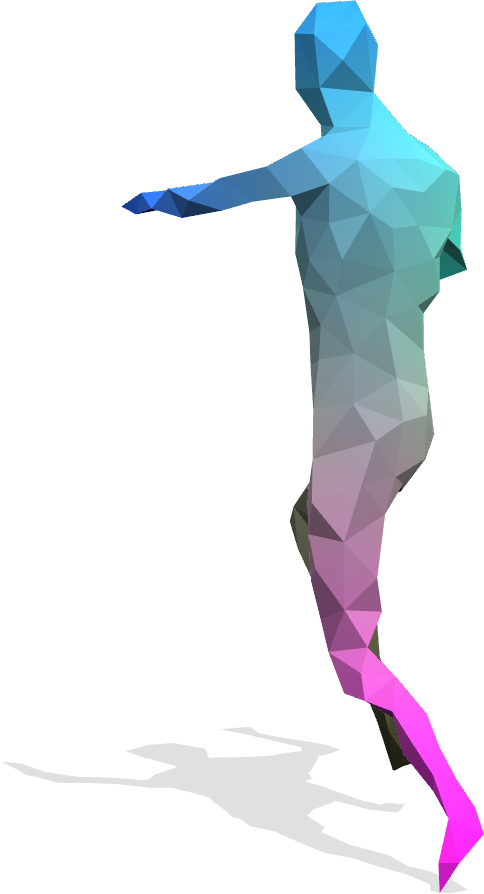}&
    \hspace{\hspaceCols}
    \includegraphics[height=\heightQ, width=\widthQ]{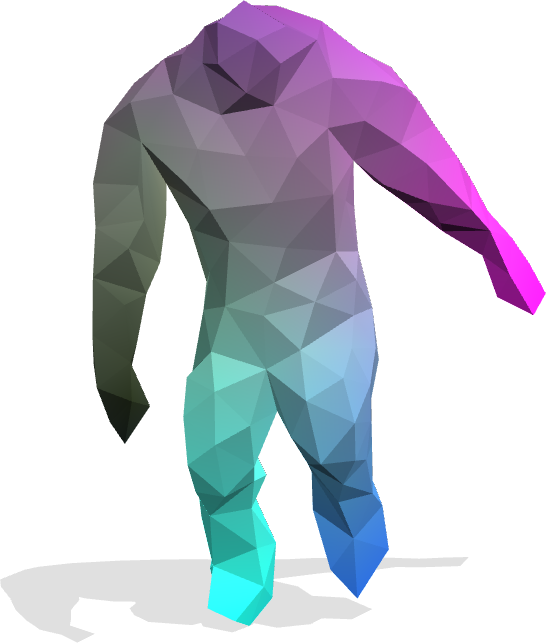}&
    \hspace{\hspaceCols}
    \includegraphics[height=\heightQ, width=\widthQ]{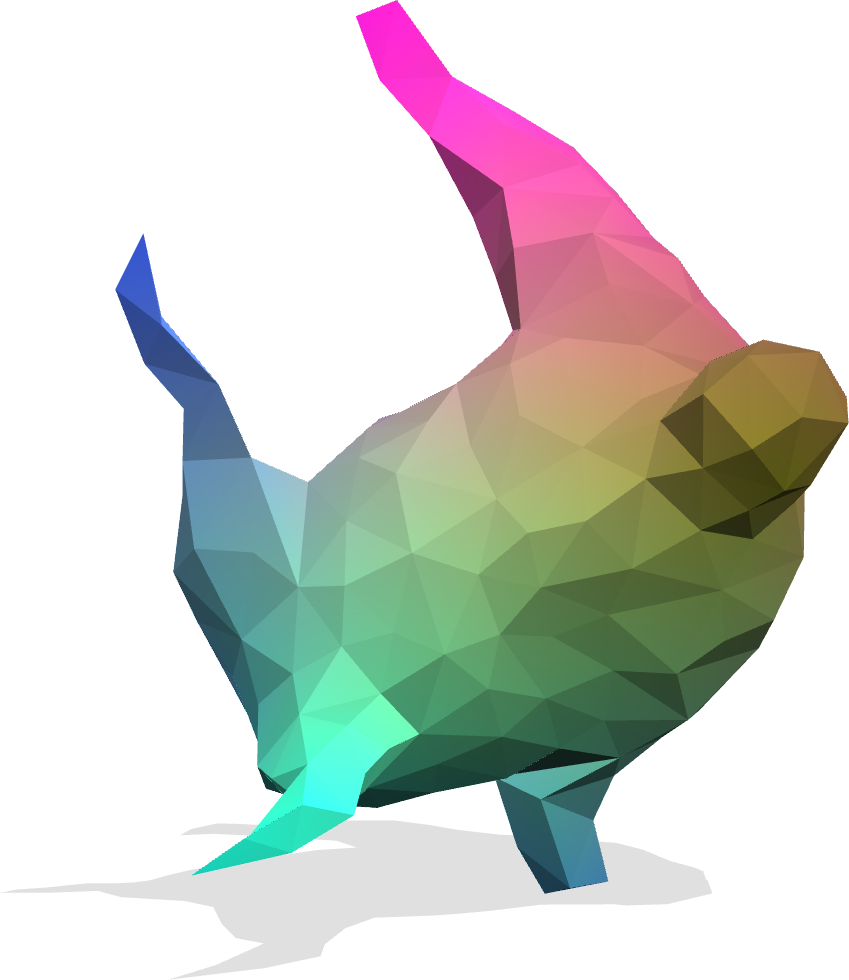}&
    \hspace{\hspaceCols}
    \includegraphics[height=\heightQ, width=\widthQ]{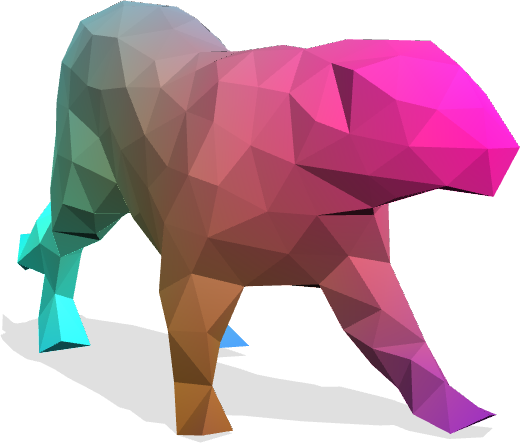}&
    \hspace{\hspaceCols}
    \includegraphics[height=\heightQ, width=\widthQ]{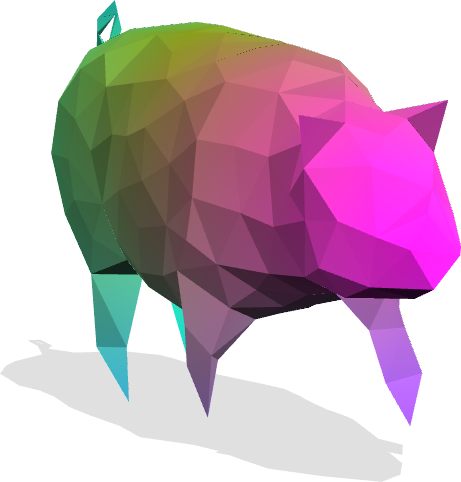}&
    \hspace{\hspaceCols}
    \includegraphics[height=\heightQ, width=\widthQ]{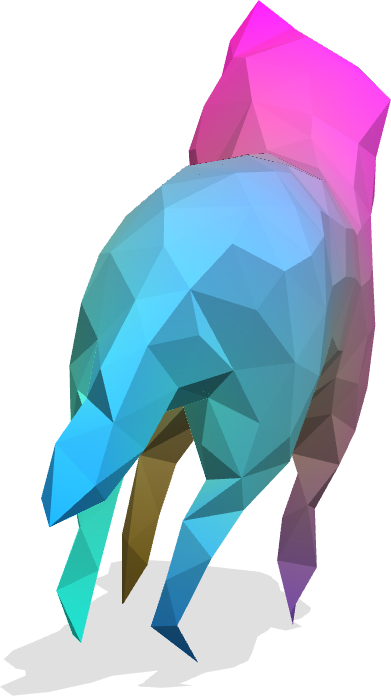}&
    \hspace{\hspaceCols}
    \includegraphics[height=\heightQ, width=\widthQ]{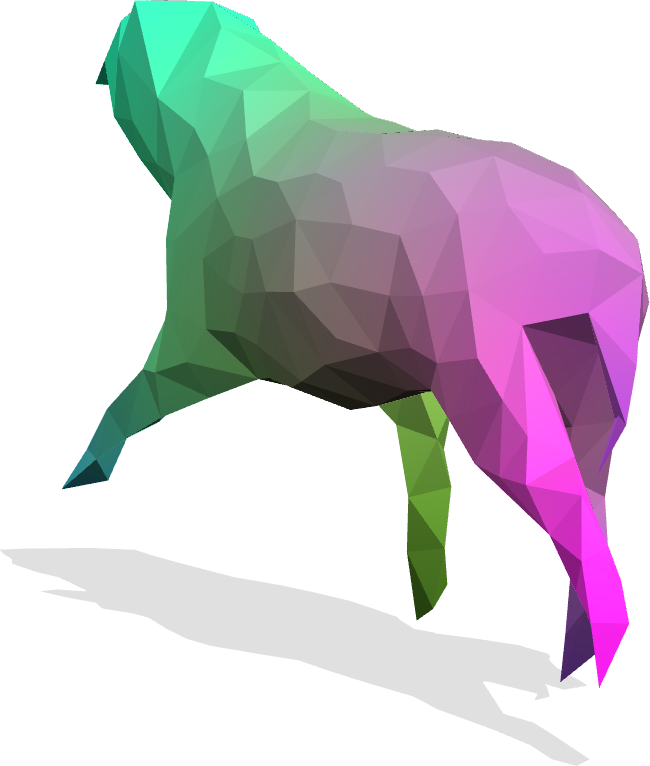}&
    \hspace{\hspaceCols}
    \includegraphics[height=\heightQ, width=\widthQ]{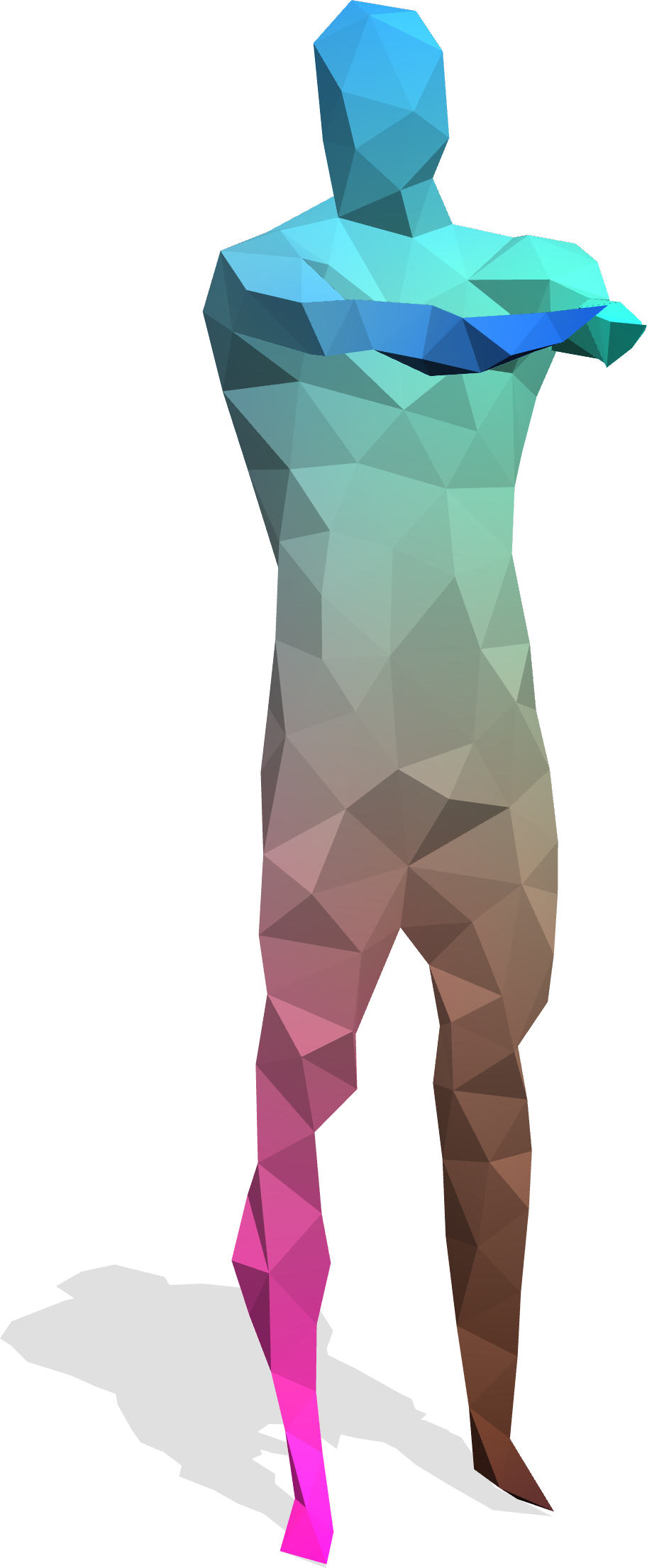}&
    \hspace{\hspaceCols}
    \includegraphics[height=\heightQ, width=\widthQ]{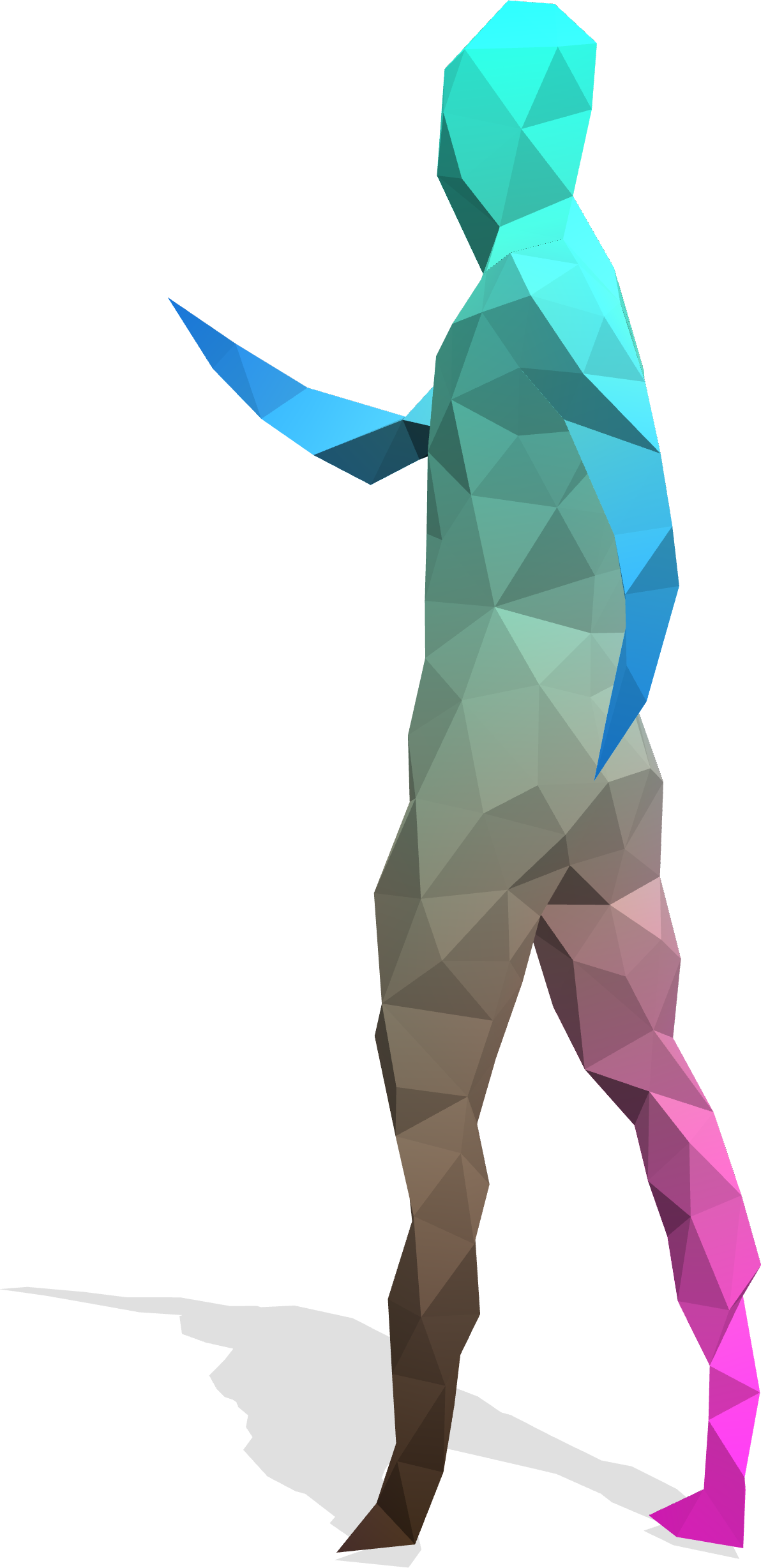}\\
    \rotatedCentering{90}{\heightQ}{\caoetal}&
    \hspace{\hspaceCols}
    \includegraphics[height=\heightQ, width=\widthQ]{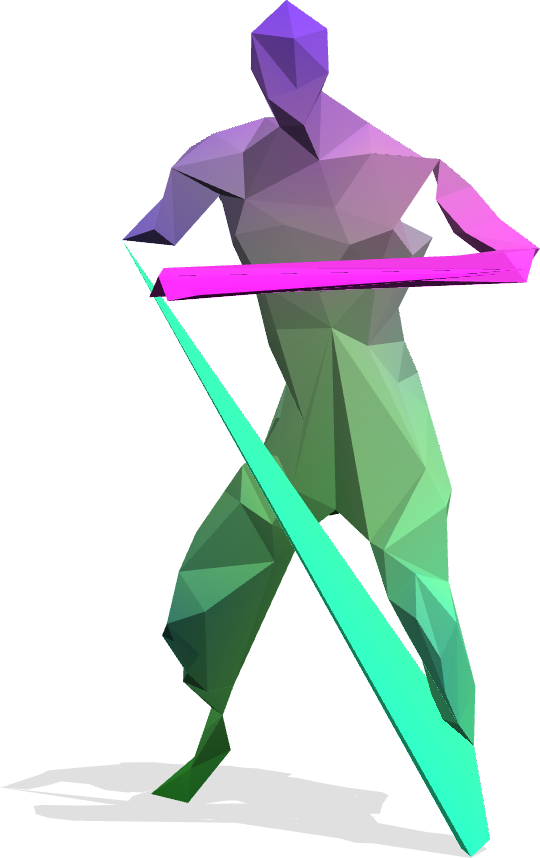}&
    \hspace{\hspaceCols}
    \includegraphics[height=\heightQ, width=\widthQ]{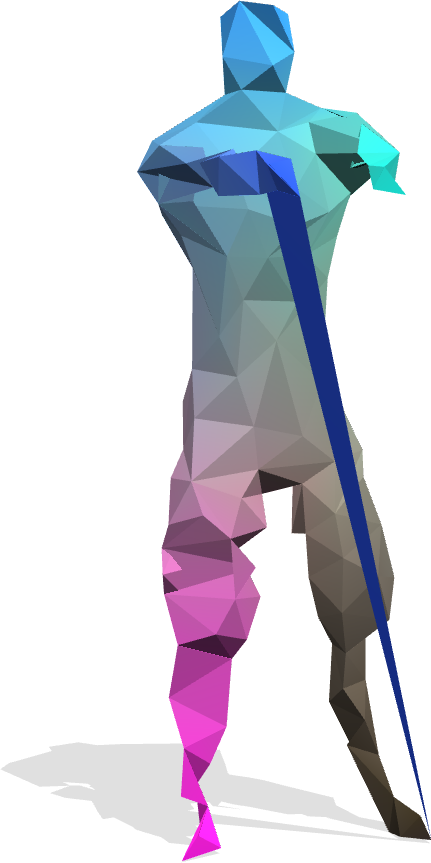}&
    \hspace{\hspaceCols}
    \includegraphics[height=\heightQ, width=\widthQ]{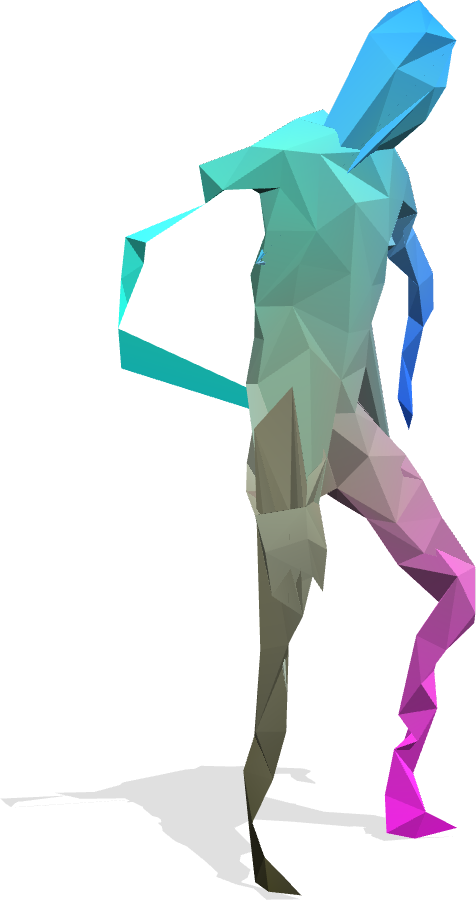}&
    \hspace{\hspaceCols}
    \includegraphics[height=\heightQ, width=\widthQ]{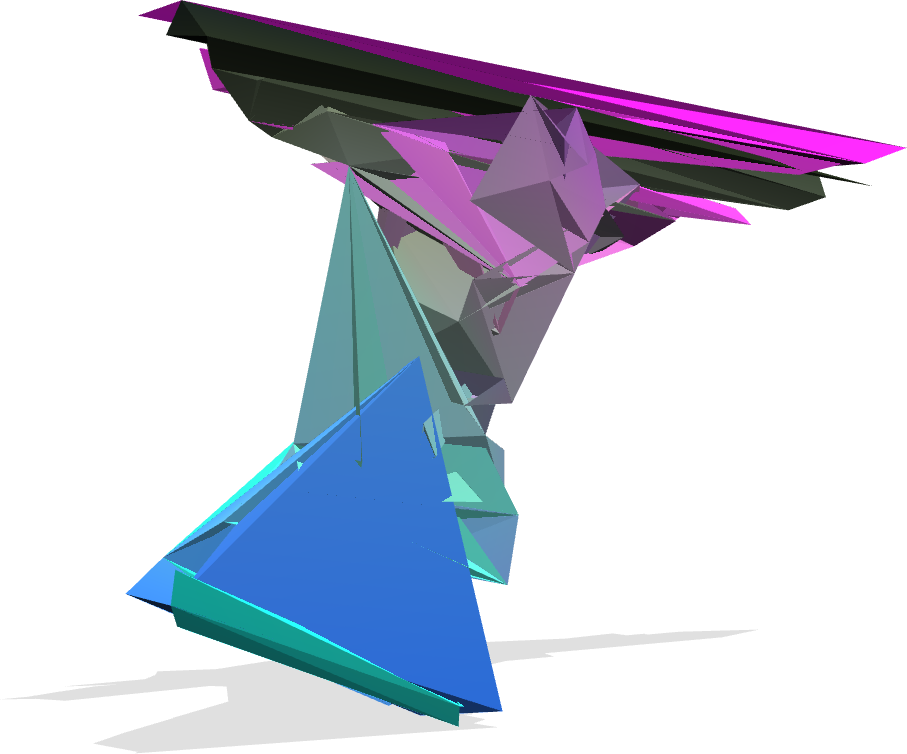}&
    \hspace{\hspaceCols}
    \includegraphics[height=\heightQ, width=\widthQ]{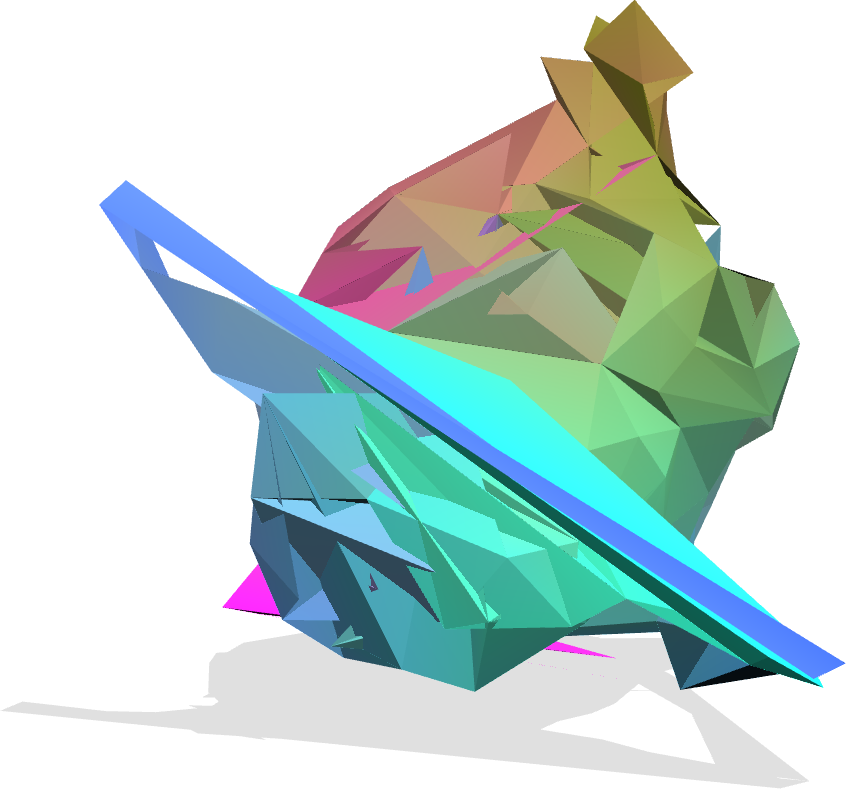}&
    \hspace{\hspaceCols}
    \includegraphics[height=\heightQ, width=\widthQ]{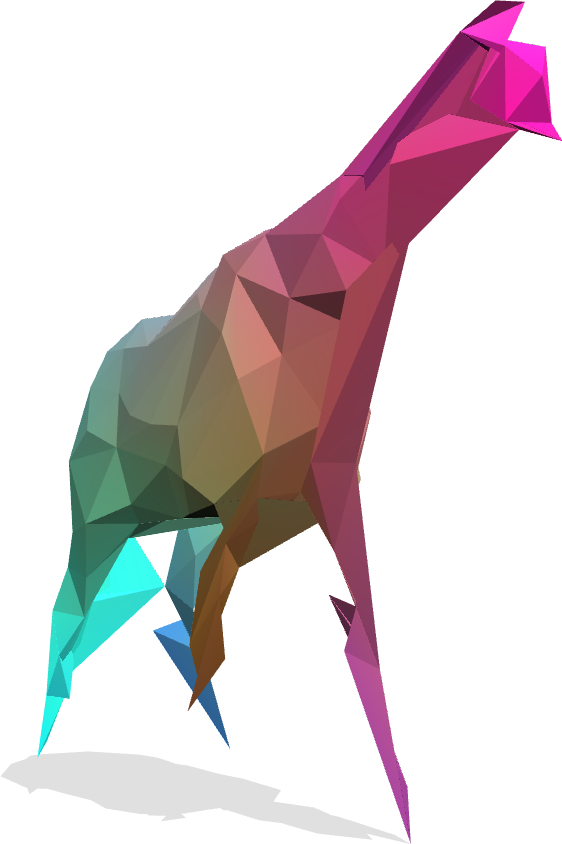}&
    \hspace{\hspaceCols}
    \includegraphics[height=\heightQ, width=\widthQ]{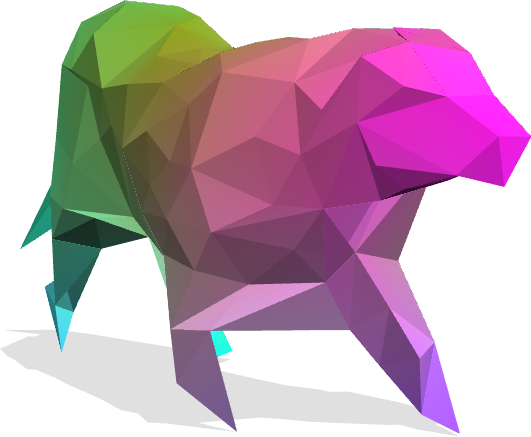}&
    \hspace{\hspaceCols}
    \includegraphics[height=\heightQ, width=\widthQ]{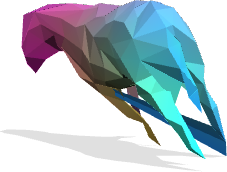}&
    \hspace{\hspaceCols}
    \includegraphics[height=\heightQ, width=\widthQ]{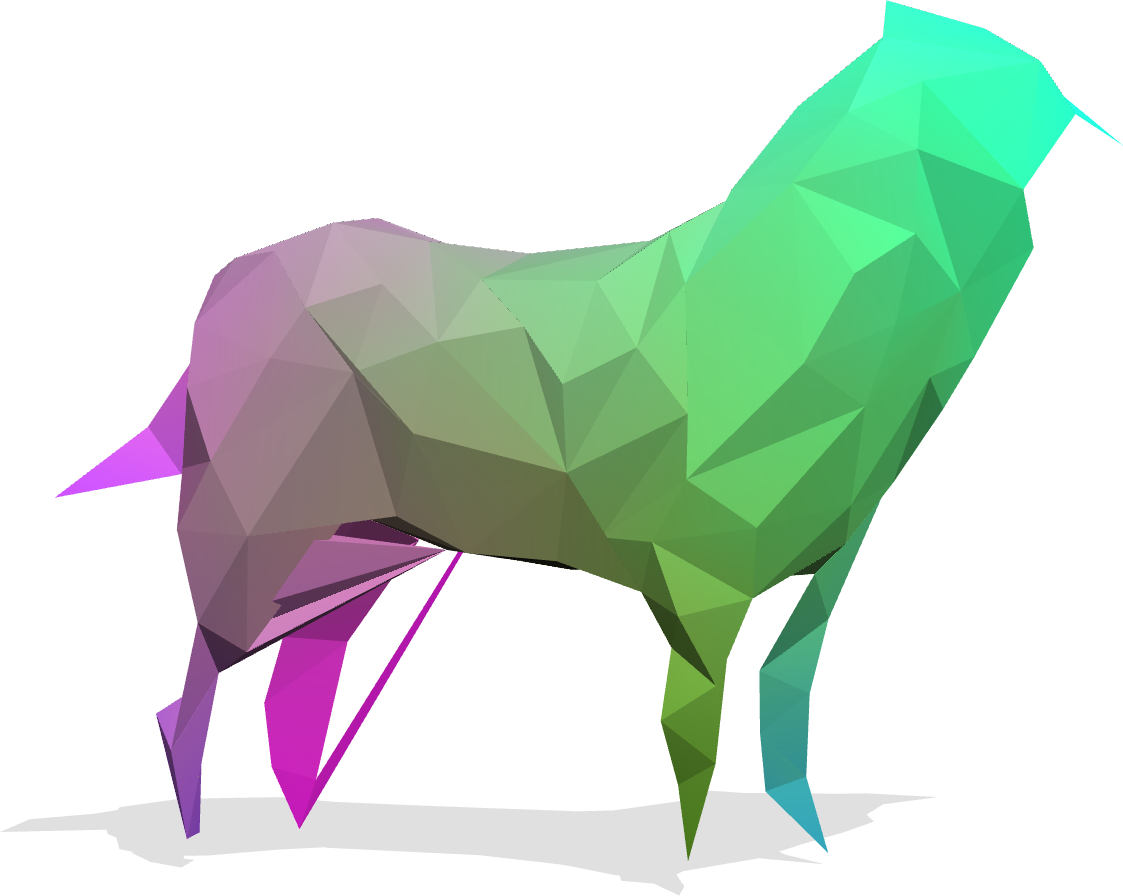}&
    \hspace{\hspaceCols}
    \includegraphics[height=\heightQ, width=\widthQ]{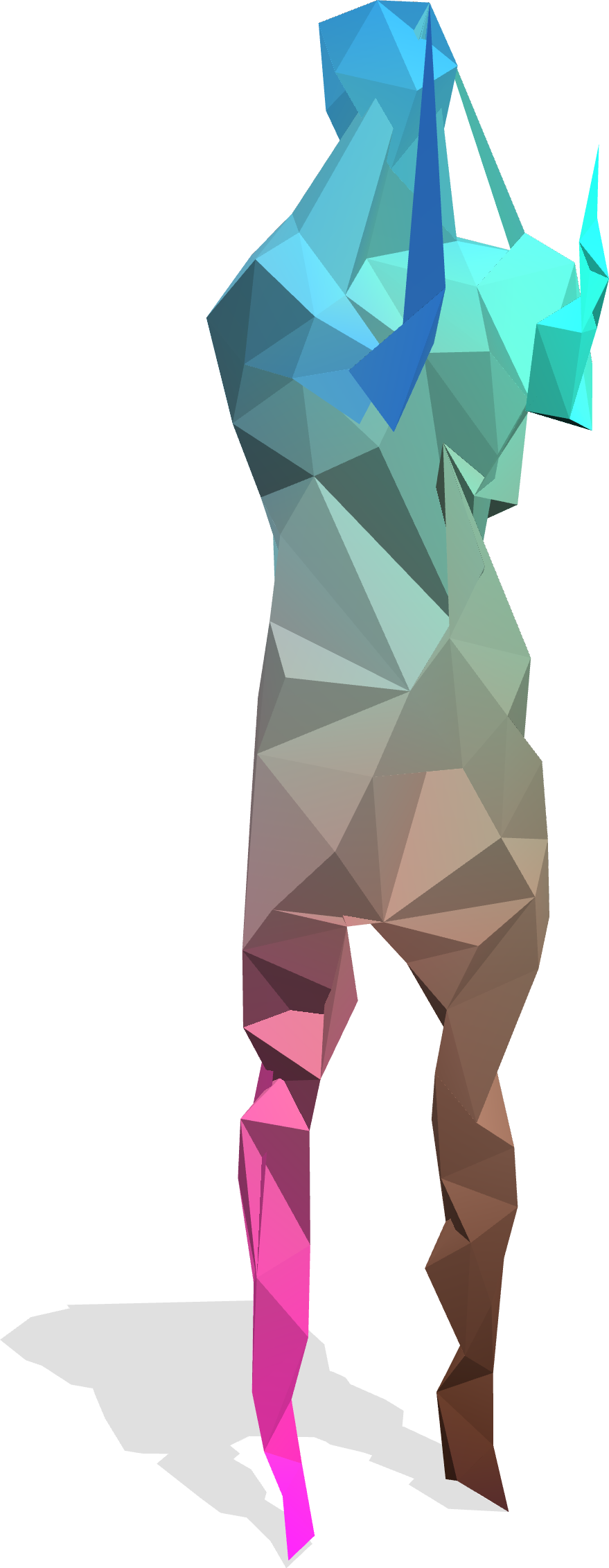}&
    \hspace{\hspaceCols}
    \includegraphics[height=\heightQ, width=\widthQ]{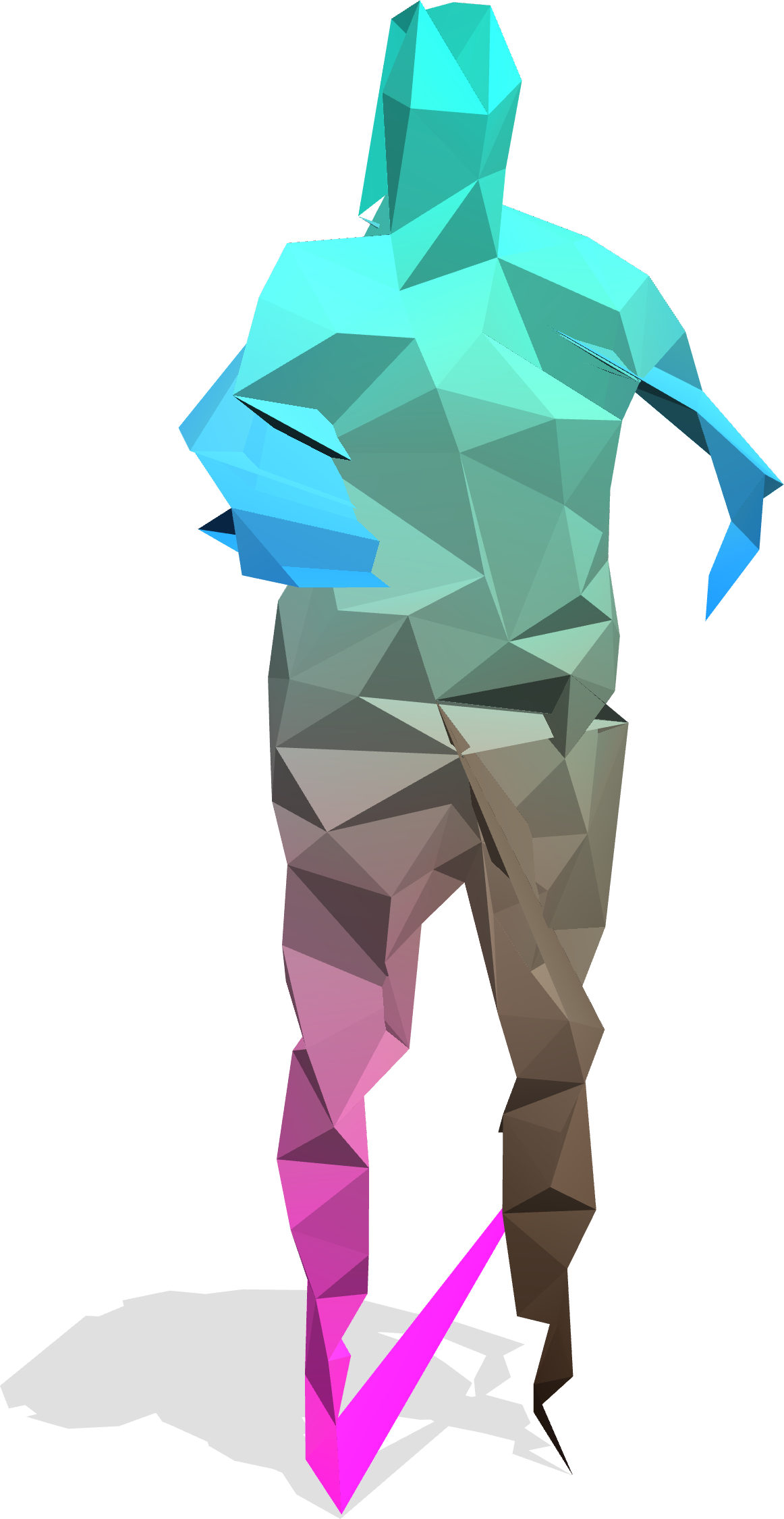}\\
    &\geoerrorstyle{0.112}&\geoerrorstyle{0.043}&\geoerrorstyle{0.058}&\geoerrorstyle{0.219}
    &\geoerrorstyle{0.144}&\geoerrorstyle{-}&\geoerrorstyle{-}&\geoerrorstyle{0.057}&
    \geoerrorstyle{0.046}&\geoerrorstyle{0.062}&\geoerrorstyle{0.079}\\
    \rotatedCentering{90}{\heightQ}{\ours}&
    \hspace{\hspaceCols}
    \includegraphics[height=\heightQ, width=\widthQ]{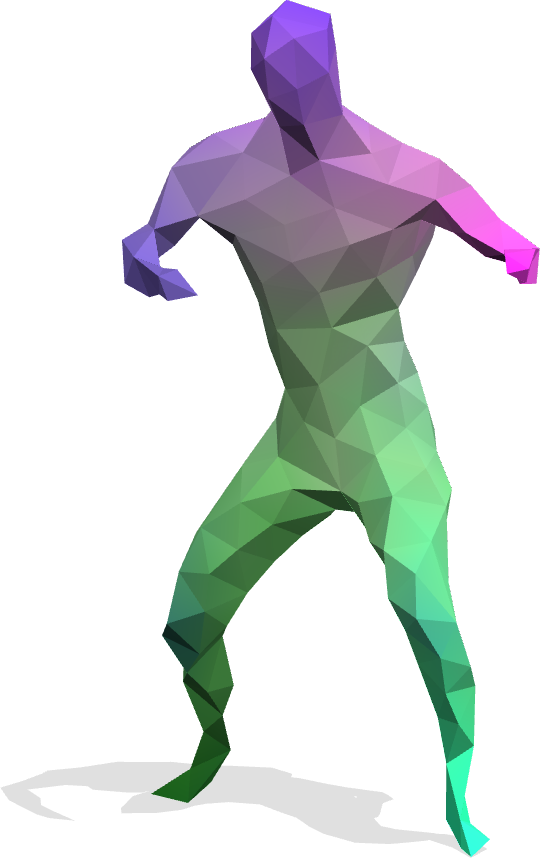}&
    \hspace{\hspaceCols}
    \includegraphics[height=\heightQ, width=\widthQ]{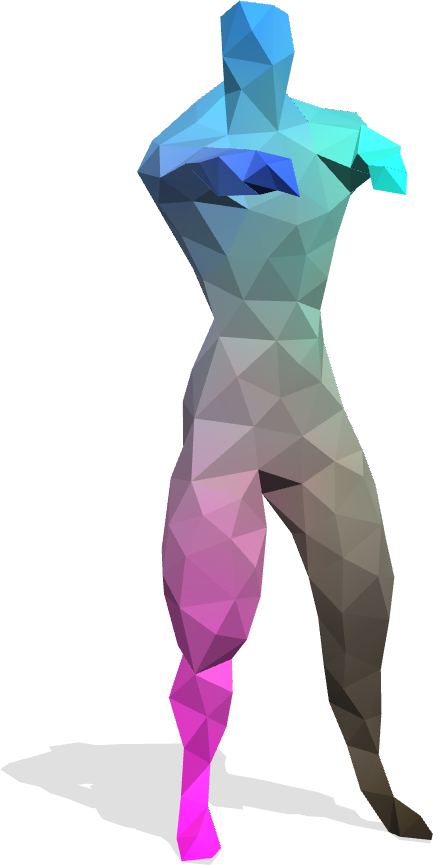}&
    \hspace{\hspaceCols}
    \includegraphics[height=\heightQ, width=\widthQ]{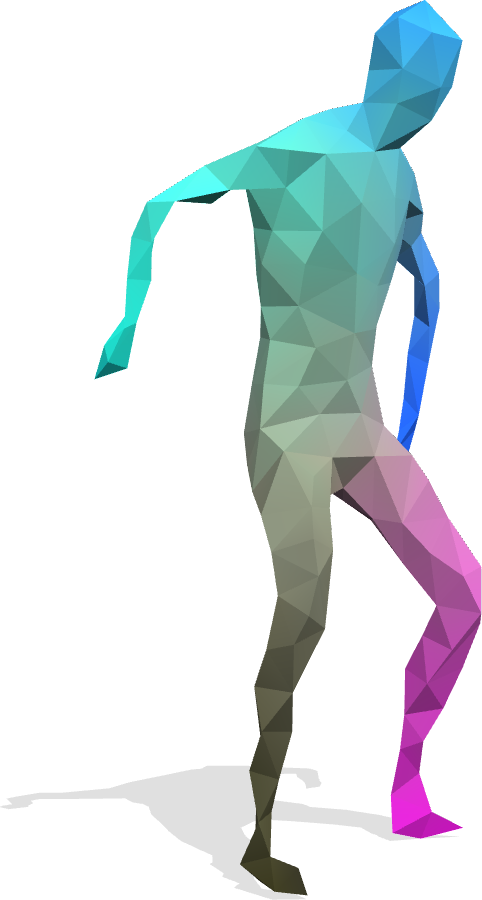}&
    \hspace{\hspaceCols}
    \includegraphics[height=\heightQ, width=\widthQ]{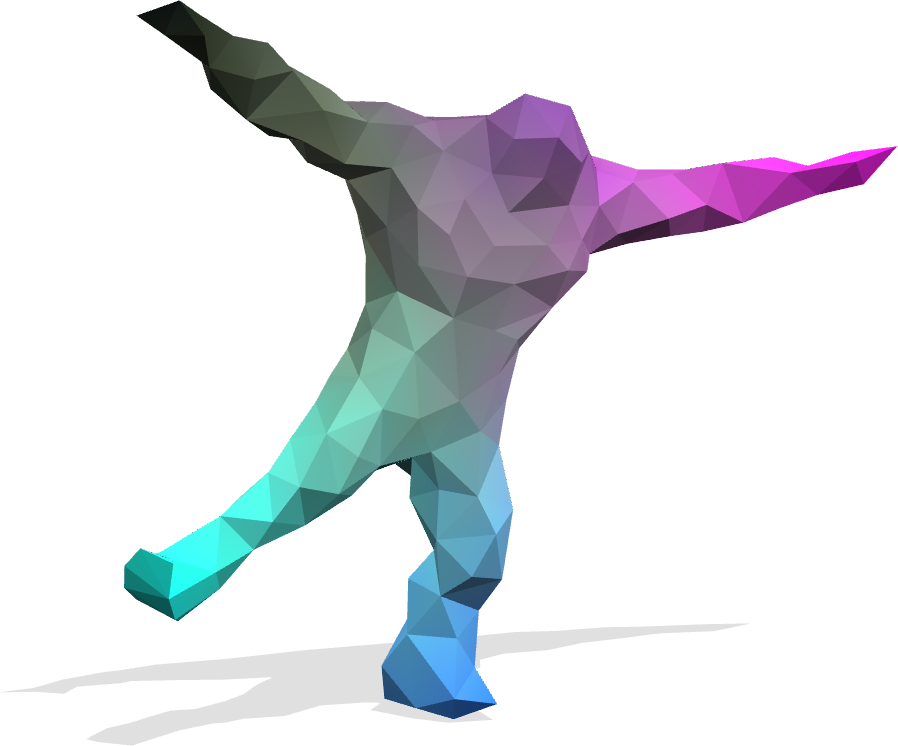}&
    \hspace{\hspaceCols}
    \includegraphics[height=\heightQ, width=\widthQ]{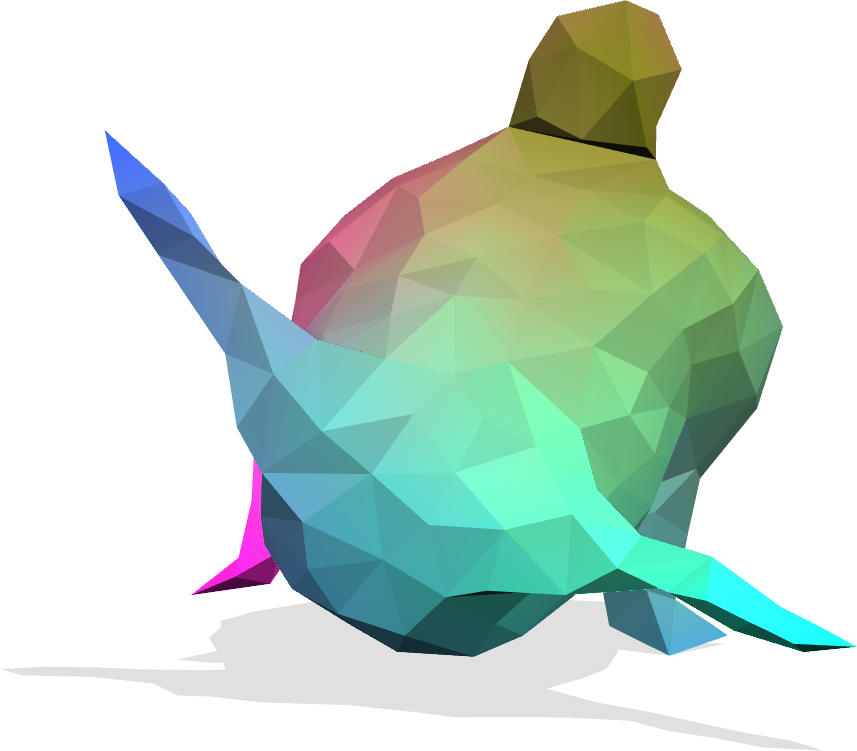}&
    \hspace{\hspaceCols}
    \includegraphics[height=\heightQ, width=\widthQ]{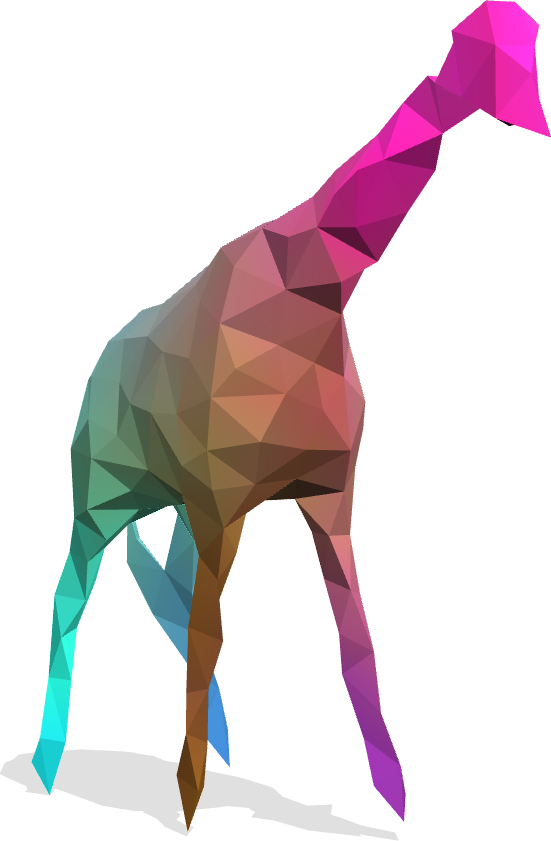}&
    \hspace{\hspaceCols}
    \includegraphics[height=\heightQ, width=\widthQ]{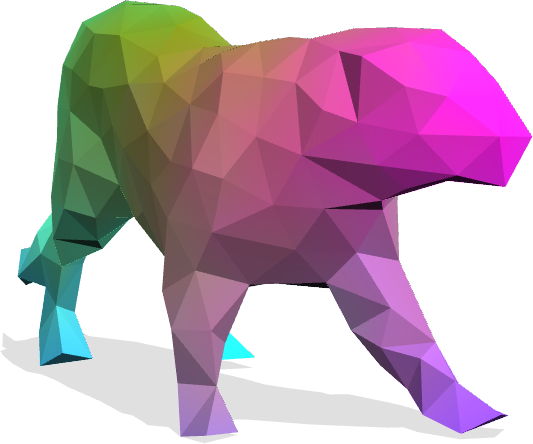}&
    \hspace{\hspaceCols}
    \includegraphics[height=\heightQ, width=\widthQ]{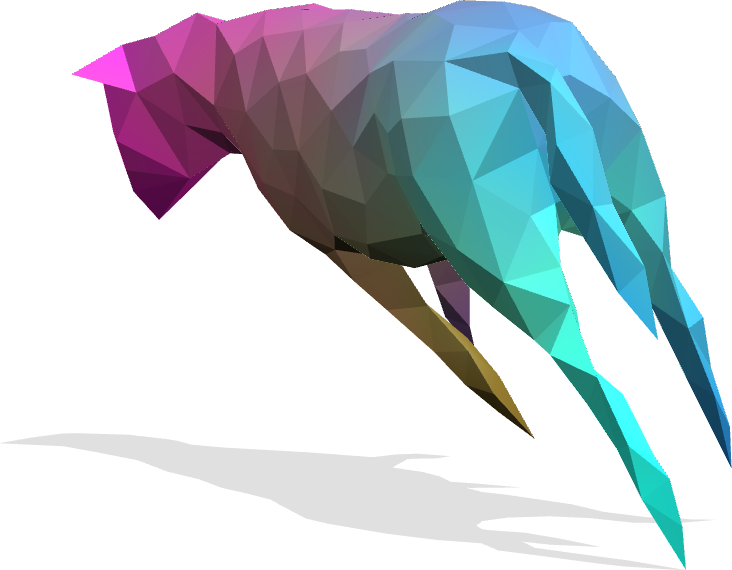}&
    \hspace{\hspaceCols}
    \includegraphics[height=\heightQ, width=\widthQ]{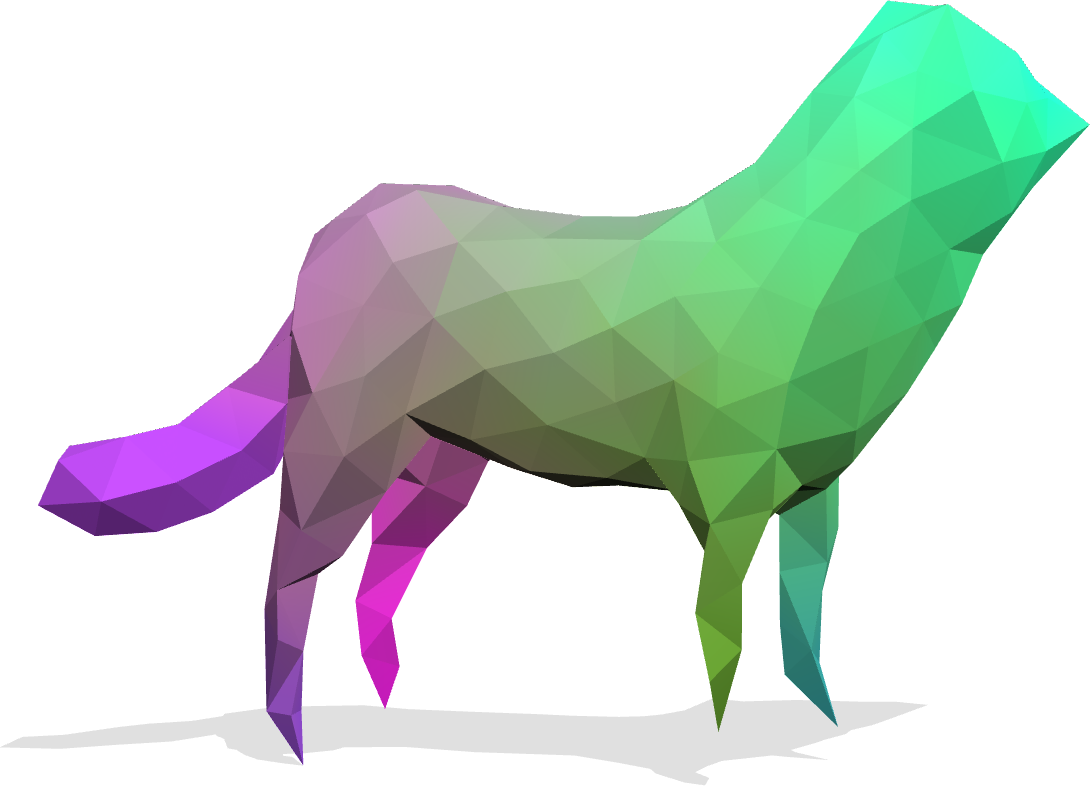}&
    \hspace{\hspaceCols}
    \includegraphics[height=\heightQ, width=\widthQ]{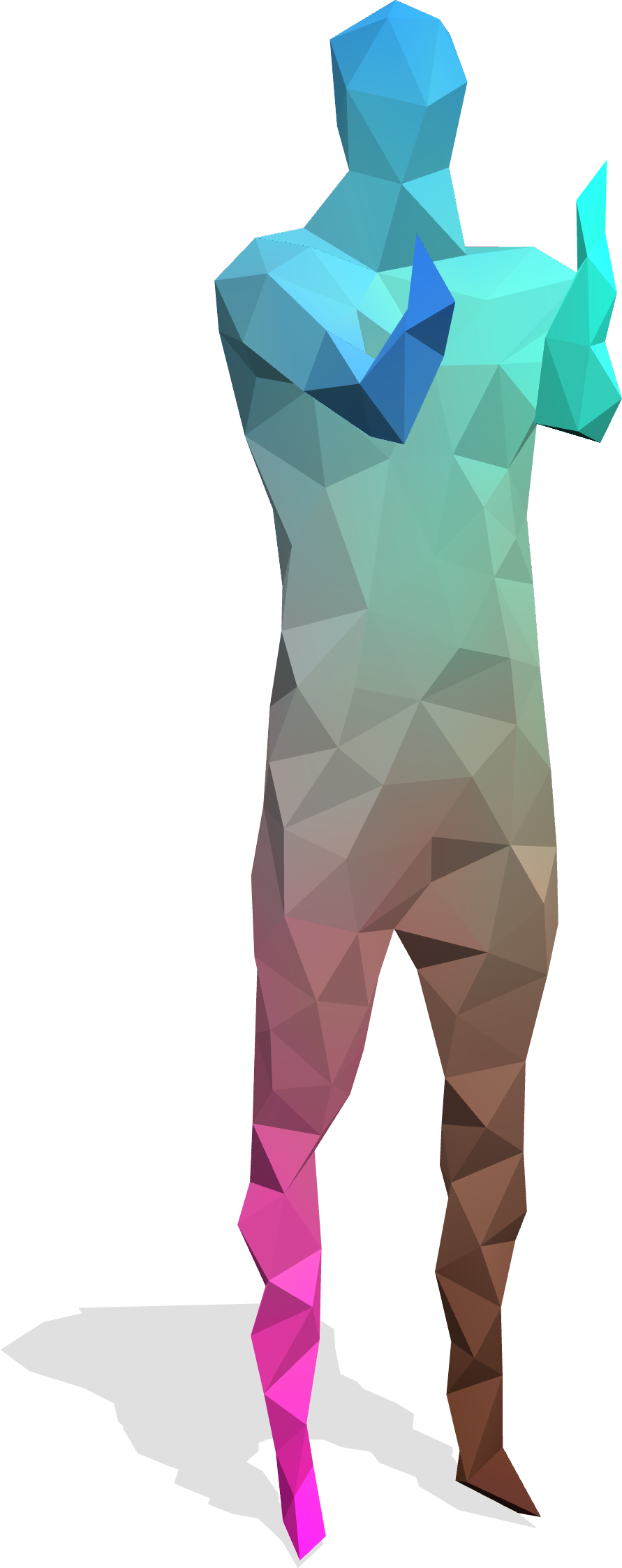}&
    \hspace{\hspaceCols}
    \includegraphics[height=\heightQ, width=\widthQ]{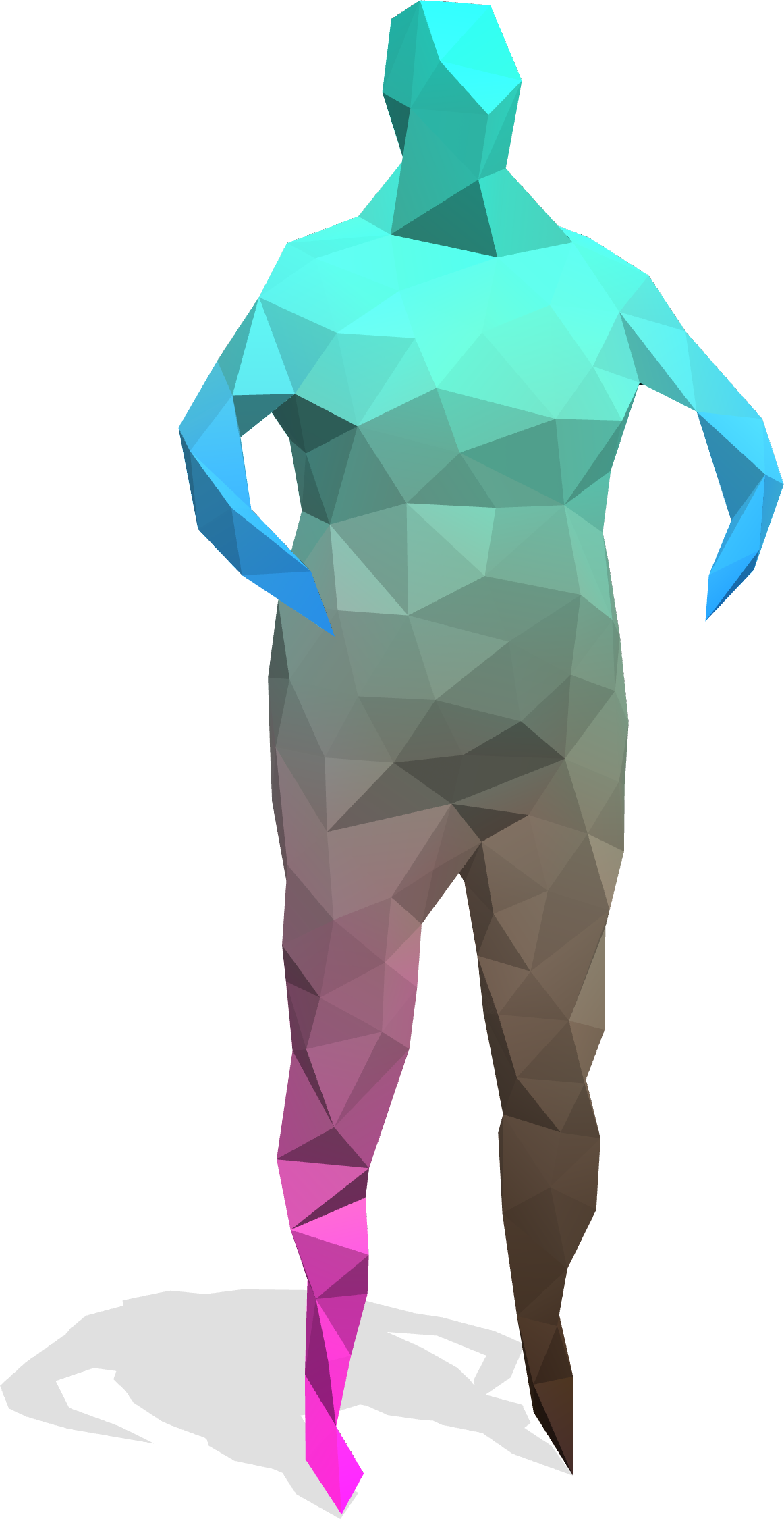}\\
    &\geoerrorstyle{0.073}&\geoerrorstyle{0.047}&\geoerrorstyle{0.039}&\geoerrorstyle{0.044}
    &\geoerrorstyle{0.060}&\geoerrorstyle{-}&\geoerrorstyle{-}&\geoerrorstyle{0.060}&
    \geoerrorstyle{0.048}&\geoerrorstyle{0.037}&\geoerrorstyle{0.040}
\end{tabular}
    \caption{{Qualitative results} on DT4D-Inter (\scircled{1},\scircled{2}), DT4D-Intra (\scircled{3}-\scircled{5}), SHREC'20 (\scircled{6},\scircled{7}), SMAL (\scircled{8},\scircled{9}) and FAUST (\scircled{10},\scircled{11}). We transfer colour and triangulation from source to target shape via computed matching. Numbers are mean geodesic errors for respective instances. Note that SHREC'20 dataset does not come with dense ground truth correspondences. Consequently,  we cannot evaluate geodesic errors for these pairs. Furthermore, we observe in columns \scircled{2}, \scircled{8} and \scircled{9} that mean geodesic errors are smaller for~\cite{cao2023unsupervised} even though matchings contain obvious geometric inconsistencies (cf. also~\cref{sec:appendix-dataset-analysis}).}
    \label{fig:appendix-qualitative}
\end{figure*}

In \cref{fig:appendix-qualitative}, we show more qualitative results on the datasets as well as individual mean geodesic errors for respective pairs.
In addition to our solutions being smooth, we can also observe that low mean geodesic error is not necessarily an indicator for geometrically consistent matchings. 
Some results computed with the method~\cite{cao2023unsupervised} yield lower error scores than ours despite visually obvious geometric inconsistencies.
This is due to discretisation artifacts stemming from the \textit{triangle-based} formalism (cf. \cref{fig:corres-types}) that we use to compute matchings.
In the datasets that we consider, ground-truth correspondences are given as point-wise maps and thus we need to convert triangle-based matchings into point-wise matchings (e.g.,~triangle-based matching $\big((m_1, n_1)$, $(m_2, n_1)$, $(m_3, n_1)\big)$ converts to point-wise matchings $(m_1, n_1)$, $(m_2, n_1)$, $(m_3, n_1)$, where $m_v \in V_\mathcal{M}$ and  $n_v \in V_\mathcal{N}$).
Given the ground-truth point-wise matching $(m_1, n_1)$, our approach does not only capture this point-wise matching, but also includes matchings $(m_2, n_1), (m_3, n_1)$ which leads to geodesic errors (because $(m_2, n_1), (m_3, n_1)$ are not part of ground-truth matchings).
Consequently, our approach can result in larger mean-geodesic errors due to local discretisation artifacts.
Nevertheless, the formulation we use~\eqref{eq:ilp-sm} is the only one which guarantees \textit{geometrically consistent} solutions.

Furthermore, we investigate geometric consistency by comparing a version of Dirichlet energies~\cite[Eq. 10]{cao2023unsupervised} of the respective matchings.
For that, we first rigidly align source and target shape with ground-truth matching.
Then we compute the matching deformation field $D \in \mathbb{R}^{|V_\mathcal{M}|\times 3}$ as $D = V_\mathcal{M} - \Pi_{\mathcal{M}\mathcal{N}}V_\mathcal{N}$ where $ \Pi_{\mathcal{M}\mathcal{N}}\in \{0,1\}^{|V_\mathcal{M}|\times |V_\mathcal{N}|}$ is the point-wise map, i.e.,~the computed matching.
Finally, we compute the respective Dirichlet energy $E_\text{Dirichlet}$ for a matching deformation field $D$ as follows
\begin{equation}
    E_\text{Dirichlet} = \text{trace}(D^T L_\mathcal{M} D)
\end{equation}
where $L_\mathcal{M}$ is the Laplacian of shape $\mathcal{M}$.
In \cref{fig:dirichlet} we show commulative Dirichlet energies on datasets FAUST, SMAL and DT4D (intra and inter).
Considering the shown Dirichlet energies, we can see that \smcomb, \fastdog\ and \ours\ (i.e.,~all methods solving problem~\ref{eq:ilp-sm} and thus optimising for a geometrically consistent matching) lead to less distorted matchings, i.e.,~matchings with less Dirichlet energy compared to the non-geometrically consistent method \caoetal.

\begin{figure*}[h]
    \centering
    \setlength{\tabcolsep}{-9pt}
    \begin{tabular}{cc}
    \hspace{-0.6cm}
         \newcommand{\pckLineWidth}{3pt}
\newcommand{\plotWidth}{0.55\columnwidth}
\newcommand{\plotHeight}{0.45\columnwidth}
\newcommand{\pckTitle}{FAUST: Dirichlet Energy}

\pgfplotsset{%
    label style = {font=\normalfont},
    tick label style = {font=\normalfont},
    title style =  {font=\normalfont},
    legend style={  fill= gray!10,
                    fill opacity=0.6, 
                    font=\normalfont,
                    draw=gray!20, %
                    text opacity=1}
}
\begin{tikzpicture}[scale=1.0, transform shape]
	\begin{axis}[
		width=\plotWidth,
		height=\plotHeight,
		grid=major,
		title=\pckTitle,
		legend style={
			at={(0.97,0.03)},
			anchor=south east,
			legend columns=1},
		legend cell align={left},
        title style={yshift=-0.2cm},
	ylabel={{$\%$ Instances w/ Smaller Energy}},
        xlabel={Dirichlet Energy Threshold},
        xmin=0,
        xmax=15,
        ylabel near ticks,
        xtick={0, 5, 10, 15},
        xticklabels={$0$, $5$, $10$, $15$},
        ymin=0,
        ymax=1,
        ytick={0, 0.20, 0.40, 0.60, 0.80, 1},
        yticklabels={$0$, $20$, $40$, $60$, $80$, $100$},
	]
    \addplot [color=cPLOT2, smooth, line width=\pckLineWidth]
    table[row sep=crcr]{%
0	0\\
0.517241379310345	0\\
1.03448275862069	0\\
1.55172413793103	0\\
2.06896551724138	0\\
2.58620689655172	0.0105263157894737\\
3.10344827586207	0.0368421052631579\\
3.62068965517241	0.0947368421052632\\
4.13793103448276	0.152631578947368\\
4.6551724137931	0.226315789473684\\
5.17241379310345	0.310526315789474\\
5.68965517241379	0.368421052631579\\
6.20689655172414	0.442105263157895\\
6.72413793103448	0.542105263157895\\
7.24137931034483	0.594736842105263\\
7.75862068965517	0.621052631578947\\
8.27586206896552	0.7\\
8.79310344827586	0.747368421052632\\
9.31034482758621	0.784210526315789\\
9.82758620689655	0.8\\
10.3448275862069	0.852631578947368\\
10.8620689655172	0.9\\
11.3793103448276	0.942105263157895\\
11.8965517241379	0.947368421052632\\
12.4137931034483	0.947368421052632\\
12.9310344827586	0.963157894736842\\
13.448275862069	0.968421052631579\\
13.9655172413793	0.973684210526316\\
14.4827586206897	0.973684210526316\\
15	0.973684210526316\\
    };
    \addlegendentry{\textcolor{black}{\caoetal: 7.15}}
    \addplot [color=cPLOT0, smooth, line width=\pckLineWidth]
    table[row sep=crcr]{%
0	-0.0029008465109188\\
0.517241379310345	0\\
1.03448275862069	0.178036485971705\\
1.55172413793103	0.753797099819674\\
2.06896551724138	0.987062178925572\\
2.58620689655172	0.991085336906896\\
3.10344827586207	1\\
3.62068965517241	1\\
13.9655172413793	1\\
14.4827586206897	1\\
15	1\\
    };
    \addlegendentry{\textcolor{black}{\smcomb: 1.37}}
    \addplot [color=cPLOT3, dotted, smooth, line width=\pckLineWidth]
    table[row sep=crcr]{%
0	-0.000645747808046444\\
0.517241379310345	0\\
1.03448275862069	0.17944139114988\\
1.55172413793103	0.755923760998398\\
2.06896551724138	0.989199317105142\\
2.58620689655172	0.993326092195403\\
3.10344827586207	1\\
3.62068965517241	1\\
14.4827586206897	1\\
15	1\\
    };
    \addlegendentry{\textcolor{black}{\fastdog: \textbf{1.35}}}
    \addplot [color=cPLOT5, dashed, smooth, line width=\pckLineWidth]
    table[row sep=crcr]{%
0	0\\
0.517241379310345	0\\
1.03448275862069	0.17989417989418\\
1.55172413793103	0.756613756613757\\
2.06896551724138	0.989417989417989\\
2.58620689655172	0.994708994708995\\
3.10344827586207	1\\
3.62068965517241	1\\
14.4827586206897	1\\
15	1\\
    };
    \addlegendentry{\textcolor{black}{\ours: \textbf{1.35}}}
        
	\end{axis}
\end{tikzpicture}& 
         \newcommand{\pckLineWidth}{3pt}
\newcommand{\plotWidth}{0.55\columnwidth}
\newcommand{\plotHeight}{0.45\columnwidth}
\newcommand{\pckTitle}{SMAL: Dirichlet Energy}

\pgfplotsset{%
    label style = {font=\normalfont},
    tick label style = {font=\normalfont},
    title style =  {font=\normalfont},
    legend style={  fill= gray!10,
                    fill opacity=0.6, 
                    font=\normalfont,
                    draw=gray!20, %
                    text opacity=1}
}
\begin{tikzpicture}[scale=1.0, transform shape]
	\begin{axis}[
		width=\plotWidth,
		height=\plotHeight,
		grid=major,
		title=\pckTitle,
		legend style={
			at={(0.97,0.03)},
			anchor=south east,
			legend columns=1},
		legend cell align={left},
        title style={yshift=-0.2cm},
        xlabel={Dirichlet Energy Threshold},
        xmin=0,
        xmax=10,
        ylabel near ticks,
        xtick={0, 2.5, 5, 7.5, 10},
        xticklabels={$0$, $2.5$, $5$, $7.5$, $10$},
        ymin=0,
        ymax=1,
        ytick={0, 0.20, 0.40, 0.60, 0.80, 1},
        yticklabels={$0$, $20$, $40$, $60$, $80$, $100$},
	]
    \addplot [color=cPLOT2, smooth, line width=\pckLineWidth]
    table[row sep=crcr]{%
0	0\\
0.344827586206897	0\\
0.689655172413793	0\\
1.03448275862069	0\\
1.37931034482759	0.176470588235294\\
1.72413793103448	0.507352941176471\\
2.06896551724138	0.661764705882353\\
2.41379310344828	0.786764705882353\\
2.75862068965517	0.838235294117647\\
3.10344827586207	0.897058823529412\\
3.44827586206897	0.919117647058823\\
3.79310344827586	0.941176470588235\\
4.13793103448276	0.955882352941177\\
4.48275862068965	0.970588235294118\\
4.82758620689655	0.977941176470588\\
5.17241379310345	0.985294117647059\\
5.51724137931035	0.992647058823529\\
5.86206896551724	0.992647058823529\\
6.20689655172414	0.992647058823529\\
6.55172413793103	0.992647058823529\\
6.89655172413793	1\\
7.24137931034483	1\\
9.31034482758621	1\\
9.6551724137931	1\\
10	1\\
    };
    \addlegendentry{\textcolor{black}{\caoetal: 2.03}}
    \addplot [color=cPLOT0, smooth, line width=\pckLineWidth]
    table[row sep=crcr]{%
0	0\\
0.344827586206897	0.00342413731581821\\
0.689655172413793	0.0066860754884076\\
1.03448275862069	0.0778067535077708\\
1.37931034482759	0.614301739234863\\
1.72413793103448	0.908729529384691\\
2.06896551724138	0.968284450562116\\
2.41379310344828	0.982189366742858\\
2.75862068965517	0.989624695395832\\
3.10344827586207	0.989273951606445\\
3.44827586206897	1\\
9.31034482758621	1\\
9.6551724137931	1\\
10	1\\
    };
    \addlegendentry{\textcolor{black}{\smcomb: \textbf{1.37}}}
    \addplot [color=cPLOT3, dotted, smooth, line width=\pckLineWidth]
    table[row sep=crcr]{%
0	-0.0015258881052338\\
0.344827586206897	0\\
0.689655172413793	0.000501443813822455\\
1.03448275862069	0.0791800768546712\\
1.37931034482759	0.617080105438744\\
1.72413793103448	0.911478154887639\\
2.06896551724138	0.969130705836472\\
2.41379310344828	0.984213787174341\\
2.75862068965517	0.991546117638772\\
3.10344827586207	0.991058391418956\\
3.44827586206897	1\\
3.79310344827586	1\\
9.31034482758621	1\\
9.6551724137931	1\\
10	1\\
    };
    \addlegendentry{\textcolor{black}{\fastdog: \textbf{1.37}}}
    \addplot [color=cPLOT5, dashed, smooth, line width=\pckLineWidth]
    table[row sep=crcr]{%
0	0\\
0.344827586206897	0\\
0.689655172413793	0\\
1.03448275862069	0.0808823529411765\\
1.37931034482759	0.617647058823529\\
1.72413793103448	0.911764705882353\\
2.06896551724138	0.970588235294118\\
2.41379310344828	0.985294117647059\\
2.75862068965517	0.992647058823529\\
3.10344827586207	0.992647058823529\\
3.44827586206897	1\\
3.79310344827586	1\\
9.6551724137931	1\\
10	1\\
    };
    \addlegendentry{\textcolor{black}{\ours: \textbf{1.37}}}
        
	\end{axis}
\end{tikzpicture} \\ 
         \newcommand{\pckLineWidth}{3pt}
\newcommand{\plotWidth}{0.55\columnwidth}
\newcommand{\plotHeight}{0.45\columnwidth}
\newcommand{\pckTitle}{DT4D-Intra: Dirichlet Energy}

\pgfplotsset{%
    label style = {font=\normalfont},
    tick label style = {font=\normalfont},
    title style =  {font=\normalfont},
    legend style={  fill= gray!10,
                    fill opacity=0.6, 
                    font=\normalfont,
                    draw=gray!20, %
                    text opacity=1}
}
\begin{tikzpicture}[scale=1.0, transform shape]
	\begin{axis}[
		width=\plotWidth,
		height=\plotHeight,
		grid=major,
		title=\pckTitle,
		legend style={
			at={(0.97,0.03)},
			anchor=south east,
			legend columns=1},
		legend cell align={left},
        title style={yshift=-0.2cm},
        xlabel={Dirichlet Energy Threshold},
        xmin=0,
        xmax=25,
        ylabel near ticks,
        xtick={0, 5, 10, 15, 20, 25},
        xticklabels={$0$, $5$, $10$, $15$, $20$, $25$},
        ymin=0,
        ymax=1,
        ytick={0, 0.20, 0.40, 0.60, 0.80, 1},
        yticklabels={$0$, $20$, $40$, $60$, $80$, $100$},
	]
    \addplot [color=cPLOT2, smooth, line width=\pckLineWidth]
    table[row sep=crcr]{%
0	0\\
0.862068965517241	0\\
1.72413793103448	0\\
2.58620689655172	0.04\\
3.44827586206897	0.07\\
4.31034482758621	0.13\\
5.17241379310345	0.27\\
6.03448275862069	0.38\\
6.89655172413793	0.48\\
7.75862068965517	0.62\\
8.62068965517241	0.7\\
9.48275862068965	0.73\\
10.3448275862069	0.78\\
11.2068965517241	0.8\\
12.0689655172414	0.81\\
12.9310344827586	0.81\\
13.7931034482759	0.82\\
14.6551724137931	0.83\\
15.5172413793103	0.86\\
16.3793103448276	0.88\\
17.2413793103448	0.91\\
18.1034482758621	0.91\\
18.9655172413793	0.91\\
19.8275862068966	0.93\\
20.6896551724138	0.95\\
21.551724137931	0.95\\
22.4137931034483	0.95\\
23.2758620689655	0.96\\
24.1379310344828	0.96\\
25	0.97\\
    };
    \addlegendentry{\textcolor{black}{\caoetal: 9.23}}
    \addplot [color=cPLOT0, smooth, line width=\pckLineWidth]
    table[row sep=crcr]{%
0	0\\
0.862068965517241	0.116825702754881\\
1.72413793103448	0.777854940261842\\
2.58620689655172	0.938196496514435\\
3.44827586206897	0.977784789613994\\
4.31034482758621	0.977006240050957\\
5.17241379310345	1\\
6.03448275862069	1\\
6.89655172413793	1\\
7.75862068965517	1\\
8.62068965517241	1\\
9.48275862068965	1\\
25	1\\
    };
    \addlegendentry{\textcolor{black}{\smcomb: {1.43}}}
    \addplot [color=cPLOT3, dotted, smooth, line width=\pckLineWidth]
    table[row sep=crcr]{%
0	0\\
0.862068965517241	0.118960174653581\\
1.72413793103448	0.778870500682544\\
2.58620689655172	0.938654287813493\\
3.44827586206897	0.979354180295526\\
4.31034482758621	0.978202355677659\\
5.17241379310345	1\\
6.03448275862069	1\\
6.89655172413793	1\\
7.75862068965517	1\\
8.62068965517241	1\\
25	1\\
    };
    \addlegendentry{\textcolor{black}{\fastdog: \textbf{1.42}}}
    \addplot [color=cPLOT5, dashed, smooth, line width=\pckLineWidth]
    table[row sep=crcr]{%
0	0\\
0.862068965517241	0.12\\
1.72413793103448	0.78\\
2.58620689655172	0.94\\
3.44827586206897	0.98\\
4.31034482758621	0.98\\
5.17241379310345	1\\
6.03448275862069	1\\
6.89655172413793	1\\
7.75862068965517	1\\
8.62068965517241	1\\
9.48275862068965	1\\
25	1\\
    };
    \addlegendentry{\textcolor{black}{\ours: \textbf{1.42}}}
        
	\end{axis}
\end{tikzpicture}& 
         \newcommand{\pckLineWidth}{3pt}
\newcommand{\plotWidth}{0.55\columnwidth}
\newcommand{\plotHeight}{0.45\columnwidth}
\newcommand{\pckTitle}{DT4D-Inter: Dirichlet Energy}

\pgfplotsset{%
    label style = {font=\normalfont},
    tick label style = {font=\normalfont},
    title style =  {font=\normalfont},
    legend style={  fill= gray!10,
                    fill opacity=0.6, 
                    font=\normalfont,
                    draw=gray!20, %
                    text opacity=1}
}
\begin{tikzpicture}[scale=1.0, transform shape]
	\begin{axis}[
		width=\plotWidth,
		height=\plotHeight,
		grid=major,
		title=\pckTitle,
		legend style={
			at={(0.97,0.03)},
			anchor=south east,
			legend columns=1},
		legend cell align={left},
        title style={yshift=-0.2cm},
        xlabel={Dirichlet Energy Threshold},
        xmin=0,
        xmax=25,
        ylabel near ticks,
        xtick={0, 5, 10, 15, 20, 25},
        xticklabels={$0$, $5$, $10$, $15$, $20$, $25$},
        ymin=0,
        ymax=1,
        ytick={0, 0.20, 0.40, 0.60, 0.80, 1},
        yticklabels={$0$, $20$, $40$, $60$, $80$, $100$},
	]
    \addplot [color=cPLOT2, smooth, line width=\pckLineWidth]
    table[row sep=crcr]{%
0	0\\
0.862068965517241	0\\
1.72413793103448	0.310344827586207\\
2.58620689655172	0.482758620689655\\
3.44827586206897	0.603448275862069\\
4.31034482758621	0.655172413793103\\
5.17241379310345	0.724137931034483\\
6.03448275862069	0.775862068965517\\
6.89655172413793	0.810344827586207\\
7.75862068965517	0.827586206896552\\
8.62068965517241	0.862068965517241\\
9.48275862068965	0.879310344827586\\
10.3448275862069	0.913793103448276\\
11.2068965517241	0.913793103448276\\
12.0689655172414	0.913793103448276\\
12.9310344827586	0.931034482758621\\
13.7931034482759	0.948275862068966\\
14.6551724137931	0.96551724137931\\
15.5172413793103	0.96551724137931\\
16.3793103448276	0.96551724137931\\
17.2413793103448	0.96551724137931\\
18.1034482758621	0.96551724137931\\
18.9655172413793	0.96551724137931\\
19.8275862068966	0.96551724137931\\
20.6896551724138	0.96551724137931\\
21.551724137931	0.96551724137931\\
22.4137931034483	0.96551724137931\\
23.2758620689655	0.96551724137931\\
24.1379310344828	0.96551724137931\\
25	0.96551724137931\\
    };
    \addlegendentry{\textcolor{black}{\caoetal: 5.39}}
    \addplot [color=cPLOT0, smooth, line width=\pckLineWidth]
    table[row sep=crcr]{%
0	0\\
0.862068965517241	0.00417044549396296\\
1.72413793103448	0.638183699689321\\
2.58620689655172	0.745132320990872\\
3.44827586206897	0.832847638070759\\
4.31034482758621	0.887869698490728\\
5.17241379310345	0.900122470836841\\
6.03448275862069	0.898758415894389\\
6.89655172413793	0.900228540386437\\
7.75862068965517	0.923533500848693\\
8.62068965517241	0.920781311141448\\
9.48275862068965	0.934175961881193\\
10.3448275862069	0.943434635750424\\
11.2068965517241	0.963939282710693\\
12.0689655172414	0.966061305511144\\
12.9310344827586	0.96407106501296\\
13.7931034482759	0.964460895459748\\
14.6551724137931	0.985854223498241\\
15.5172413793103	0.984543027515065\\
16.3793103448276	1\\
17.2413793103448	1\\
18.1034482758621	1\\
18.9655172413793	1\\
19.8275862068966	1\\
25	1\\
    };
    \addlegendentry{\textcolor{black}{\smcomb: {2.67}}}
    \addplot [color=cPLOT3, dotted, smooth, line width=\pckLineWidth]
    table[row sep=crcr]{%
0	0\\
0.862068965517241	0.000955514110875206\\
1.72413793103448	0.640067466499979\\
2.58620689655172	0.74842192669897\\
3.44827586206897	0.834674325263295\\
4.31034482758621	0.889570711391701\\
5.17241379310345	0.901213721654609\\
6.03448275862069	0.900177421829087\\
6.89655172413793	0.901851501023329\\
7.75862068965517	0.923767923225605\\
8.62068965517241	0.923710850552706\\
9.48275862068965	0.934389252731661\\
10.3448275862069	0.944874046104338\\
11.2068965517241	0.966667160082535\\
12.0689655172414	0.967363269969113\\
12.9310344827586	0.966095532122587\\
13.7931034482759	0.967162409885519\\
14.6551724137931	0.988778799127147\\
15.5172413793103	0.987615695885773\\
16.3793103448276	1\\
17.2413793103448	1\\
18.1034482758621	1\\
18.9655172413793	1\\
25	1\\
    };
    \addlegendentry{\textcolor{black}{\fastdog: \textbf{2.66}}}
    \addplot [color=cPLOT5, dashed, smooth, line width=\pckLineWidth]
    table[row sep=crcr]{%
0	0\\
0.862068965517241	0\\
1.72413793103448	0.641304347826087\\
2.58620689655172	0.75\\
3.44827586206897	0.83695652173913\\
4.31034482758621	0.891304347826087\\
5.17241379310345	0.902173913043478\\
6.03448275862069	0.902173913043478\\
6.89655172413793	0.902173913043478\\
7.75862068965517	0.923913043478261\\
8.62068965517241	0.923913043478261\\
9.48275862068965	0.934782608695652\\
10.3448275862069	0.945652173913043\\
11.2068965517241	0.967391304347826\\
12.0689655172414	0.967391304347826\\
12.9310344827586	0.967391304347826\\
13.7931034482759	0.967391304347826\\
14.6551724137931	0.989130434782609\\
15.5172413793103	0.989130434782609\\
16.3793103448276	1\\
17.2413793103448	1\\
18.1034482758621	1\\
18.9655172413793	1\\
25	1\\
    };
    \addlegendentry{\textcolor{black}{\ours: \textbf{2.66}}}
        
	\end{axis}
\end{tikzpicture}
    \end{tabular}%
    \caption{{Geometric consistency evaluation} by comparing Dirichlet energies on FAUST, SMAL, DT4D-Intra and DT4D-Inter.
    Values in legends are mean Dirichlet energies across all instances.
    Methods solving problem~\ref{eq:ilp-sm} (i.e.,~\smcomb, \fastdog\ and \ours) and thus providing geometric consistency guarantees lead to smaller Dirichlet energies compared to non-geometrically consistent method by \caoetal\ across all datasets.
    }\label{fig:dirichlet}
\end{figure*}

\subsection{Constraint Splitting}\label{sec:bdd-splitting}
A BDD is a directed acyclic graph $G=(V,A^0,A_1)$ that represents Boolean functions $f : \{0,1\}^n \rightarrow \{0,1\}$ as follows:
\begin{itemize}
    \item Each regular node $v$ has an associated variable $\var(v)$. We call all nodes that belong to a variable a layer and $\idx(v) = l$ if $v$ is the $l$-th node of layer.
    \item Additionally, there are two terminal nodes, the true terminal $\top$ and the false terminal $\bot$.
    \item Each regular node $v$ has two outgoing arcs: a zero-arc $v,v^0 \in A^0$ and a one-arc $v,v^1 \in A^1$. They correspond to assigning the associated variable the respective value.
    \item Each arc leads either from variable $x_i$ to $x_{i+1}$ or to a terminal.
    \item There exists exactly one root node associated with $x_1$.
    \item All paths from the unique node associated with $x_1$ to the true terminal $\top$ correspond to variable assignments $x$ with $f(x) = 1$, all paths to terminal $\bot$ to $f(x) = 0$ (this is actually a small departure from the original BDD definition which allows skipping arcs).
\end{itemize}
Additionally, BDDs are required to be reduced, i.e.,~no isomorphic subgraphs are present.
For a good introduction to BDDs we refer to~\cite{taocp4a}. 
See the top-part of Figure~\ref{fig:BDD} for a BDD of the constraint $\sum_{i=1}^8 x_i = 2$.

The constraint splitting is desribed in Algorithm~\ref{alg:BDD-splitting}.
It proceeds as follows: Let a variable $x_i$ be specified after which to split the constraint involving variables $x_1,\ldots,x_n$.
Let $k$ be the number of nodes associated with variable $x_{i+1}$.
Then auxiliary variables $y_1,\ldots,y_k$ are introduced.
The BDD is split into two sub-BDDs, which we call left and right sub-BDD.
The left contains variables $x_1,\ldots,x_i,y_1,\ldots,y_k$, the right one $y_1,\ldots,y_k,x_{k+1},x_n$.
The arcs are arranged so that $y_j = 1$ $\Leftrightarrow$ the $j$-th node associated to variable $x_{i+1}$ is used.

\begin{algorithm}[htb!]
\newcommand\mycommfont[1]{\footnotesize\ttfamily\textcolor{cBLUE}{#1}}
      \SetCommentSty{mycommfont}
      \DontPrintSemicolon
      \KwInput{
      BDD $G=(V,A^0,A^1)$\\
      split variables $x_i$
      }
      \KwResult{
      left BDD $G_l=(V_l,A_l^0,A_l^1)$,\\
      right BDD $G_r = (V_r,A_r^0,A_r^1)$
      }
      \tcp{Determine number of auxiliary variables}
      $k := \max\{\idx(v) : v \in V, \var(v) = i+1\}$\;
      \tcp{Initialize left and right sub-BDD with left and right part of the nodes and arcs of the original BDD}
      $V_l = \{v \in V : \var(v) \leq i\}$\;
      $A_l^{\beta} = \{vw \in A^{\beta} : \var(w) \leq i\}$, $\beta\ \in \{0,1\}$\;
      $V_r = \{v \in V : \var(v) > i\}$\;
      $A_r^{\beta} = \{vw \in A^{\beta} : \var(v) > i\}$, $\beta \in \{0,1\}$\;
      \tcp{Add auxiliary nodes for left BDD}
      $V_l = V_l \cup \{aux_{1,1},\ldots,aux_{1,k}\}$\;
      $\var(aux_{1,l}) := y_j$ $\forall l$\;
      \For{$j=2,\ldots,k$}
      {
      $V_l = V_l \cup \{aux_{j,1},\ldots, aux_{j,k-j+2}\}$ \;
      $\var(aux_{j,l}) := y_j$ $\forall l$\;
      }
      \tcp{connect left BDD part to auxiliary nodes}
      $A_l^{\beta} = \cup \{v\, aux_{1,\idx(w)} : vw \in A^{\beta}, \var(v) = i\}$, $\beta \in \{0,1\}$\;
      \tcp{connect left BDD auxiliary nodes}
      \For{$j=1,\ldots,k-1$}
      {
      \For{$l=1\ldots,k-j+1$}
      {
      \If{$l = k-j$}{
      $A^1_l = A^l_l \cup \{ aux_{j,k-j} aux_{j+1,k-j}\}$\;
      } \ElseIf{$l = k-j+1$} {
      $A^0_l = A^0_l \cup \{ aux_{j,k-j+1} aux_{j+1,k-j}\}$\;
      } \Else {
      $A^0_l = A^0_l \cup \{ aux_{j,l} aux_{j+1,l} \}$\;
      }
      }
      }
      $A^1_l = A^1_l \cup \{aux_{k,1} \top\}$\;
      $A^0_l = A^0_l \cup \{aux_{k,2} \top\}$\;
      \tcp{Add auxiliary nodes for right BDD}
      \For{$j=1,\ldots,k$}
      {
      $V_r = V_r \cup \{aux'_{j,1},\ldots,aux'_{j,j}\}$\;
      $\var(aux'_{jl}) := y_j$\;
      }
      \tcp{Connect right BDD auxiliary nodes}
      \For{$j=1,\ldots,k-1$}
      {
      $A^1_r = A^1_r \cup \{aux'_{j,1} aux'_{j+1,2}\}$\;
      $A^0_r = A^0_r \cup \{aux'_{j,l} aux'_{j+1,l+1}\}$, $l \in \{2,j\}$\;
      }
      \tcp{Connect auxiliary nodes to right BDD part}
      $A^1_r = A^1_r \cup \{aux'_{k,1} w: w \in V, \var(w) = i+1, \idx(w) = 1\}$\;
      \For{l=2,\ldots,k}
      {
      $A^0_r = A^0_r \cup \{aux'_{k,1} w: w \in V, \var(w) = i+1, \idx(w) = l\}$\;
      }

      Add all missing arcs such that endpoint goes to $\bot$.
      \caption{BDD Splitting}
      \label{alg:BDD-splitting}
\end{algorithm}

In Figure~\ref{fig:BDD} we illustrate the BDD splitting for a simple cardinality constraint. 

\begin{figure*}[htb!]
\begin{center}
    \begin{tikzpicture}[scale=0.3]
      \tikzset{
      vertex/.style={draw, circle},
      terminal/.style={rectangle,draw},
      wariable/.style={draw},
      one/.style={solid,->},
      zero/.style={dashed,->}
      }
      
      \fill[cBLUE!100,nearly transparent] (-0.9,-0.9) -- (6.9,-0.9) -- (6.9,4.9) -- (-0.9,4.9) -- cycle;
      \fill[cPINK!100,nearly transparent] (7.1,-0.9) -- (14.9,-0.9) -- (14.9,4.9) -- (7.1,4.9) -- cycle;
      \node [style=vertex] (0) at (0, 0) {};
      \node [style=vertex] (1) at (2, 0) {};
      \node [style=vertex] (2) at (2, 2) {};
      \node [style=vertex] (3) at (4, 4) {};
      \node [style=vertex] (4) at (4, 2) {};
      \node [style=vertex] (5) at (4, 0) {};
      \node [style=vertex] (6) at (6, 0) {};
      \node [style=vertex] (7) at (6, 2) {};
      \node [style=vertex] (8) at (6, 4) {};
      \node [style=vertex] (9) at (8, 4) {};
      \node [style=vertex] (10) at (8, 2) {};
      \node [style=vertex] (11) at (8, 0) {};
      \node [style=vertex] (12) at (10, 0) {};
      \node [style=vertex] (13) at (10, 2) {};
      \node [style=vertex] (14) at (10, 4) {};
      \node [style=vertex] (15) at (12, 4) {};
      \node [style=vertex] (16) at (12, 2) {};
      \node [style=vertex] (17) at (12, 0) {};
      \node [style=vertex] (18) at (14, 4) {};
      \node [style=vertex] (19) at (14, 2) {};
      \node [style=terminal] (21) at (24, 3) {$\top$};
      \draw[one] (0) to (2);
      \draw[one] (2) to (3);
      \draw[one] (1) to (4);
      \draw (4) to (8);
      \draw (5) to (7);
      \draw (7) to (9);
      \draw (6) to (10);
      \draw (10) to (14);
      \draw (11) to (13);
      \draw (13) to (15);
      \draw (12) to (16);
      \draw (16) to (18);
      \draw (17) to (19);
      \draw[zero] (3) to (8);
      \draw[zero] (8) to (9);
      \draw[zero] (9) to (14);
      \draw[zero] (14) to (15);
      \draw[zero] (15) to (18);
      \draw[zero] (2) to (4);
      \draw[zero] (4) to (7);
      \draw[zero] (7) to (10);
      \draw[zero] (10) to (13);
      \draw[zero] (13) to (16);
      \draw[zero] (16) to (19);
      \draw[zero] (0) to (1);
      \draw[zero] (1) to (5);
      \draw[zero] (5) to (6);
      \draw[zero] (6) to (11);
      \draw[zero] (11) to (12);
      \draw[zero] (12) to (17);
      \draw[one] (18) to (21);
      \draw[one] (19) to (21);
      
      \begin{scope}[shift={(0,-8)}]
      \fill[cBLUE!100,nearly transparent] (-0.9,-0.9) -- (6.9,-0.9) -- (6.9,4.9) -- (-0.9,4.9) -- cycle;
      \fill[cGREEN!100, nearly transparent] (15.1,-0.9) -- (20.9,-0.9) -- (20.9,4.9) -- (15.1,4.9) -- cycle;
      \node [style=vertex] (0) at (0, 0) {};
      \node [style=vertex] (1) at (2, 0) {};
      \node [style=vertex] (2) at (2, 2) {};
      \node [style=vertex] (3) at (4, 4) {};
      \node [style=vertex] (4) at (4, 2) {};
      \node [style=vertex] (5) at (4, 0) {};
      \node [style=vertex] (6) at (6, 0) {};
      \node [style=vertex] (7) at (6, 2) {};
      \node [style=vertex] (8) at (6, 4) {};
      \node [style=vertex] (aux10) at (16,0) {};
      \node [style=vertex] (aux11) at (16,2) {};
      \node [style=vertex] (aux12) at (16,4) {};
      \node [style=vertex] (aux20) at (18,0) {};
      \node [style=vertex] (aux21) at (18,2) {};
      \node [style=vertex] (aux22) at (18,4) {};
      \node [style=vertex] (aux30) at (20,0) {};
      \node [style=vertex] (aux31) at (20,2) {};
      \node [style=terminal] (top) at (24,1) {$\top$};
      \draw[one] (0) to (2);
      \draw[one] (2) to (3);
      \draw[one] (1) to (4);
      \draw (4) to (8);
      \draw (5) to (7);
      \draw[zero] (3) to (8);
      \draw[zero] (2) to (4);
      \draw[zero] (4) to (7);
      \draw[zero] (0) to (1);
      \draw[zero] (1) to (5);
      \draw[zero] (5) to (6);
      \draw[zero] (8) to (aux12);
      \draw[one] (7) to (aux12);
      \draw[zero] (7) to (aux11);
      \draw[one] (6) to (aux11);
      \draw[zero] (6) to (aux10);
      \draw[one] (aux12) to (aux22);
      \draw[zero] (aux11) to (aux21);
      \draw[zero] (aux10) to (aux20);
      \draw[zero] (aux22) to (aux31);
      \draw[one] (aux21) to (aux31);
      \draw[zero] (aux20) to (aux30);
      \draw[zero] (aux31) to (top);
      \draw[one] (aux30) to (top);
      \end{scope}
      
      \begin{scope}[shift={(0,-16)}]
      \fill[cPINK!100,nearly transparent] (7.1,-0.9) -- (14.9,-0.9) -- (14.9,4.9) -- (7.1,4.9) -- cycle;
      \fill[cGREEN!100, nearly transparent] (-6.9,-0.9) -- (-1.1,-0.9) -- (-1.1,4.9) -- (-6.9,4.9) -- cycle;
        \node [style=vertex] (aux10) at (-6,0) {};
        \node [style=vertex] (aux20) at (-4,0) {};
        \node [style=vertex] (aux21) at (-4,2) {};
        \node [style=vertex] (aux30) at (-2,0) {};
        \node [style=vertex] (aux31) at (-2,2) {};
        \node [style=vertex] (aux32) at (-2,4) {};
        \node [style=vertex] (9) at (8, 4) {};
        \node [style=vertex] (10) at (8, 2) {};
        \node [style=vertex] (11) at (8, 0) {};
        \node [style=vertex] (12) at (10, 0) {};
        \node [style=vertex] (13) at (10, 2) {};
        \node [style=vertex] (14) at (10, 4) {};
        \node [style=vertex] (15) at (12, 4) {};
        \node [style=vertex] (16) at (12, 2) {};
        \node [style=vertex] (17) at (12, 0) {};
        \node [style=vertex] (18) at (14, 4) {};
        \node [style=vertex] (19) at (14, 2) {};
        \node [style=terminal] (term) at (24,3) {$\top$};
        \draw[one] (aux10) to (aux21);
        \draw[zero] (aux10) to (aux20);
        \draw[zero] (aux21) to (aux32);
        \draw[one] (aux20) to (aux31);
        \draw[zero] (aux20) to (aux30);
        \draw[one] (aux30) to (11);
        \draw[zero] (aux31) to (10);
        \draw[zero] (aux32) to (9);
      \draw (10) to (14);
      \draw (11) to (13);
      \draw (13) to (15);
      \draw (12) to (16);
      \draw (16) to (18);
      \draw (17) to (19);
      \draw[zero] (9) to (14);
      \draw[zero] (14) to (15);
      \draw[zero] (15) to (18);
      \draw[zero] (10) to (13);
      \draw[zero] (13) to (16);
      \draw[zero] (16) to (19);
      \draw[zero] (11) to (12);
      \draw[zero] (12) to (17);
      \draw[one] (19) to (term);
      \draw[zero] (18) to (term);
      \end{scope}
\begin{scope}[shift={(0,-19)}]
      \node[wariable] (x1) at (0,0) {$x_1$};
      \node[wariable] (x2) at (2,0) {$x_2$};
      \node[wariable] (x3) at (4,0) {$x_3$};
      \node[wariable] (x4) at (6,0) {$x_4$};
      \node[wariable] (x5) at (8,0) {$x_5$};
      \node[wariable] (x6) at (10,0) {$x_6$};
      \node[wariable] (x7) at (12,0) {$x_7$};
      \node[wariable] (x8) at (14,0) {$x_8$};
      \node[wariable] (y1) at (16,0) {$y_1$};
      \node[wariable] (y2) at (18,0) {$y_2$};
      \node[wariable] (y3) at (20,0) {$y_3$};
      \node[wariable] (y1') at (-6,0) {$y_1$};
      \node[wariable] (y2') at (-4,0) {$y_2$};
      \node[wariable] (y3') at (-2,0) {$y_3$};
\end{scope}
      \end{tikzpicture}
    \end{center}
    \caption{BDD splitting for constraint $x_1 + x_2 + x_3 + x_4 + x_5 + x_6 + x_7 + x_8 = 2$ (top) with chunk size 4, resulting in two sub-BDDs (middle and bottom). Three additional variables $y_1,y_2,y_3$ are introduced.
    $\top$ denotes true terminal node, solid arcs denote variable assignment to $1$ and dashed ones to $0$.
    The false terminal node $\bot$ is not shown for readability.
    Non-shown arcs all lead to the false terminal $\bot$.
    \rebuttalfix{We note that constraint splitting works for arbitrary linear constraints.}
    }
    \label{fig:BDD}
\end{figure*}

In \cref{fig:sparse-pattern}, we show the sparsity pattern of the constraint matrix $A$ of the optimisation problem~\eqref{eq:ilp-sm}.
It visualises the amount of variables involved in individual constraints and especially that the equalities $A^\mathcal{M} x = \mathbf{1}$ as well as  $A^\mathcal{N} x = \mathbf{1}$ involve many variables.
The latter motivates our improved parallelism schemes.
\begin{figure}[htb!]
    \centering
    \hspace{-0.3cm}
    \input{figs/tikz/sparse_pattern}
    \caption{{Sparsity pattern} of constraint matrix $A$ for a pair of shapes with $4$ triangles each. We can see that the $A^\partial$ block only couples few variables (non-zeros per row $\approx20$), while the $A^\mathcal{M}$ and $A^\mathcal{N}$ blocks couple many variables (non-zeros per row $\approx20\cdot|T_\mathcal{N}|$ or rather $\approx20\cdot|T_\mathcal{M}|$) which motivates the constraint splitting technique for enhanced parallelism.}
    \label{fig:sparse-pattern}
\end{figure}

\myparagraph{Evaluation}
\label{sec:splitting_comparison}
In Figure~\ref{fig:constraint_splitting_convergence} we evaluate performance of our solver (with quasi-Newton updates) with or without the BDD splitting strategy. We observe that constraint splitting gives runtime improvement on top of our contribution of quasi-Newton updates.

\begin{figure}[htb!]
    \centering
    \includegraphics[width=0.5\textwidth]{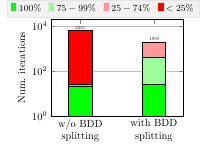}
    \caption{
    {Comparison of number of iterations and GPU utilisation} in Algorithm~\ref{alg:lbfgs} with and without BDD splitting on a shape matching problem with $500$ triangles. The y-axis denotes number of iterations required for traversing all BDDs (in log-scale) and bar colors indicate GPU utilisation. Due to BDD splitting the number of iterations is reduced and GPU utilisation is increased. 
    }
    \label{fig:bdd_splitting_layout_comparison}
\end{figure}

In Figure~\ref{fig:bdd_splitting_layout_comparison} we report GPU utilisation and number of (sequential) iterations required for a complete BDD traversal in Alg~\ref{alg:lbfgs}. Due to BDD splitting the number of iterations is reduced and GPU utilisation is increased. 

\begin{figure}[htb!]
    \centering
    \begin{tabular}{cc}
        \hspace{-1cm}
        \newcommand{\drawsquare}[1]{%
\begin{tikzpicture}[#1]%
\node[fill=cPLOT5, inner sep=0pt,minimum size=7pt] {};
\end{tikzpicture}%
}

\newcommand{\drawplus}[1]{%
\begin{tikzpicture}[#1]%
\node[mark size=4pt, color=cPLOT5, line width=1.5pt] at (0, -1) {\pgfuseplotmark{+}};
\end{tikzpicture}%
}

\newcommand{\drawcircle}[1]{%
\begin{tikzpicture}[#1]%
\node[mark size=3pt, color=cPLOT5, line width=1.5pt] at (0, -1) {\pgfuseplotmark{o}};
\end{tikzpicture}%
}

\newcommand{\drawtriangle}[1]{%
\begin{tikzpicture}[#1]%
\node[mark size=5pt, color=cPLOT5, line width=1.5pt] at (0, -1) {\pgfuseplotmark{triangle}};
\end{tikzpicture}%
}

\pgfplotsset{%
	label style = {font=\normalfont},
	tick label style = {font=\normalfont},
	title style =  {font=\normalfont},
	legend style={  fill= gray!10,
		fill opacity=1, 
		font=\normalfont,
		draw=gray!20, %
		text opacity=1}
}
\newcommand{\runtimeLineWidth}{3pt}
\newcommand{\rtCplotWidth}{0.8\textwidth}
\newcommand{\rtCplotHeight}{0.6\textwidth}
\pgfplotscreateplotcyclelist{plot_colors}{
smooth,line width=\runtimeLineWidth,color=cPLOT5,dashed,dash pattern=on 3pt off 3pt\\%
smooth,line width=\runtimeLineWidth,color=cPLOT5\\%
}
\begin{tikzpicture}[scale=0.5, transform shape]
	\begin{axis}[%
            cycle list name=plot_colors,
		width=\rtCplotWidth,
		height=\rtCplotHeight,
        title style={align=center},
		title={FAUST: Constraint Splitting comparison},
		title style={yshift=-0.25cm},
		scale only axis,
		grid=major,
		legend style={
			at={(0.98,0.98)},
			anchor=north east,
			legend columns=1,
			legend cell align={left}},
             xlabel={Time [s]},
             ylabel shift = -0.1cm,
             ylabel={$\leftarrow$ Relative gaps $\leftarrow$},
		xmin=-20,
		xmax=250,
		ymin=0.000001,
		ymax=1,
            ymode=log,
		ytick={0.000001, 0.00001, 0.0001, 0.001, 0.01, 0.1, 1},
		ylabel near ticks,
		]

    \addplot table [x index = 0, y index = 1, col sep=comma] {figs/tikz/logs/faust_lbfgs_final_lb_results_500_upwards.csv};
    \addlegendentry{Without constraint splitting\, \drawtriangle{}};
    
    \addplot table [x index = 0, y index = 1, col sep=comma] {figs/tikz/logs/faust_v2_lbfgs_split_final_lb_results_500_upwards.csv};
    \addlegendentry{With constraint splitting\, \drawcircle{}};
    
    \node[color=cPLOT5, inner sep=0pt, scale=2.5, minimum size=7pt] at (153.7339791666667, 0.0009921594239426948) {\pgfuseplotmark{triangle}};

    \node[color=cPLOT5, inner sep=1pt, scale=1.8, line width=1pt] at (120.79632727272725, 0.002610783941259561) {\pgfuseplotmark{o}};

    \end{axis}
\end{tikzpicture}&
        \hspace{-1.1cm}
        \newcommand{\drawplus}[1]{%
\begin{tikzpicture}[#1]%
\node[mark size=4pt, color=cPLOT5, line width=1.5pt] at (0, -1) {\pgfuseplotmark{+}};
\end{tikzpicture}%
}

\newcommand{\drawcircle}[1]{%
\begin{tikzpicture}[#1]%
\node[mark size=3pt, color=cPLOT5, line width=1.5pt] at (0, -1) {\pgfuseplotmark{o}};
\end{tikzpicture}%
}

\newcommand{\drawtriangle}[1]{%
\begin{tikzpicture}[#1]%
\node[mark size=4pt, color=cPLOT5, line width=1pt] at (0, -1) {\pgfuseplotmark{triangle}};
\end{tikzpicture}%
}

\pgfplotsset{%
	label style = {font=\normalfont},
	tick label style = {font=\normalfont},
	title style =  {font=\normalfont},
	legend style={  fill= gray!10,
		fill opacity=1, 
		font=\normalfont,
		draw=gray!20, %
		text opacity=1}
}
\newcommand{\runtimeLineWidth}{3pt}
\newcommand{\rtCplotWidth}{0.8\textwidth}
\newcommand{\rtCplotHeight}{0.6\textwidth}
\pgfplotscreateplotcyclelist{plot_colors}{
smooth,line width=\runtimeLineWidth,color=cPLOT5,dashed,dash pattern=on 3pt off 3pt\\%
smooth,line width=\runtimeLineWidth,color=cPLOT5\\%
}
\begin{tikzpicture}[scale=0.5, transform shape]
	\begin{axis}[%
            cycle list name=plot_colors,
		width=\rtCplotWidth,
		height=\rtCplotHeight,
        title style={align=center},
		title={DT4D-Intra: Constraint Splitting comparison},
		title style={yshift=-0.25cm},
		scale only axis,
		grid=major,
		legend style={
			at={(0.98,0.98)},
			anchor=north east,
			legend columns=1,
			legend cell align={left}},
             xlabel={Time [s]},
		xmin=-10,
		xmax=199,
		ymin=0.000001,
		ymax=1,
            ymode=log,
		ytick={0.000001, 0.00001, 0.0001, 0.001, 0.01, 0.1, 1},
		ylabel near ticks,
		]

    \addplot table [x index = 0, y index = 1, col sep=comma] {figs/tikz/logs/dt4d_lbfgs_final_lb_results_500_upwards.csv};
    \addlegendentry{Without constraint splitting\, \drawtriangle{}};
    
    \addplot table [x index = 0, y index = 1, col sep=comma] {figs/tikz/logs/dtd_lbfgs_split_final_lb_results_500_upwards.csv};
    \addlegendentry{With constraint splitting\, \drawcircle{}};

    \node[color=cPLOT5, inner sep=0pt, scale=2.5, minimum size=7pt] at (99.81567272727273, 0.0012258442488621958) {\pgfuseplotmark{triangle}};
    \node[color=cPLOT5, inner sep=1pt, scale=1.8, line width=1pt] at (75.4741636363636, 0.0011593452992476594) {\pgfuseplotmark{o}};

    \end{axis}
\end{tikzpicture}
    \end{tabular}%
    \caption{\textbf{Convergence} plots with and without constraint splitting. The results are averaged over 5 instances with different shape resolutions from $\{500, 550, \cdots, 1000\}$. Curves depict relative dual gaps and markers denote quality of primal solutions through primal-dual gaps. Both solvers use our quasi-Newton updates. 
    Our contribution of constraint splitting Sec~\ref{sec:faster_optimisation} yields improvement in convergence although with slightly worse dual objectives. 
    }
    \label{fig:constraint_splitting_convergence}
\end{figure}

\subsection{Quasi-Newton Updates}
\label{sec:quasi_newton_details}

Here we provide implementation details of our Algorithm~\ref{alg:lbfgs}. For quasi-Newton updates we directly use L-BFGS~\cite{liu1989lbfgs} without modifications. For parallel min-marginal averaging we use the dual optimisation algorithm of~\cite{abbas2022fastdog}. For completeness we describe our procedure for quasi-Newton updates below by introducing the following notation
\begin{equation}  
    \begin{split}
    s_k &= \lambda_{k+1} - \lambda_k, \\
    y_k &= g(\lambda_{k+1}) - g(\lambda_k), \\
    \rho_k &= \frac{1}{\langle s_k, y_k \rangle}, \\
    \end{split}
\end{equation}
where $g(\lambda_k)$ is a (super-)gradient of~\eqref{eq:dual-problem} at $\lambda_k$. 

The L-BFGS algorithm maintains an approximation of the Hessian inverse by storing information of at most $m$ past iterates. The Algorithm~\ref{alg:lbfgs_hessian_update_expanded} describes the procedure for storing such iterates (used in Alg.~\ref{alg:lbfgs} line~\ref{alg:lbfgs_update_hessian}). Given this information about previous iterates Algorithm~\ref{alg:lbfgs_direction_expanded} computes ascent direction for maximising~\eqref{eq:dual-problem} which is used in Alg.~\ref{alg:lbfgs} line~\ref{alg:lbfgs_update_direction}. 

By denoting the dual objective~\eqref{eq:dual-problem} as $E(\lambda)$ in Algorithm~\ref{alg:lbfgs_find_step_size_expanded} we provide our strategy to find a step size with sufficient improvement. In detail, we increase the step size by a factor $\overline{\alpha} > 1$ if we find non-negative improvement in the objective which is below the threshold $\Delta_{min}$ and decrease the step size by $0 < \underline{\alpha} < 1$ if we find non-improvement in the objective. The step size search is done for at most $K$ many iterations for efficiency reasons. If the final step size does not yield an improvement in the objective we do not perform the LBFGS update in the current iteration. For ensuring sufficient ascent we set $\Delta_{min} = 10^{-6} (E(\lambda^1) - E(\lambda^0))$ where $\lambda^0$ denotes Lagrange variables at the start of dual optimisation and $\lambda^1$ after first invocation of Alg.~\ref{alg:lbfgs}. Rest of the parameters are set as $\epsilon = 10^{-8}, \overline{\alpha} = 1.1, \underline{\alpha} = 0.8$ and $K = 5$ for all experiments. 
\begin{algorithm}[htb!]
\footnotesize
\newcommand\mycommfont[1]{\footnotesize\ttfamily\textcolor{cBLUE}{#1}}
\SetCommentSty{mycommfont}
\DontPrintSemicolon
\KwInput
{
current Lagrange variables and subgradient $\lambda_i, g_i$,\\
Previous diff.\ of Lagrange vars.\ 
$S = (s_{k - 1}, \ldots, s_{k - m})$, \\
Previous diff.\ of subgradients 
$Y = (y_{k - 1}, \cdots, y_{k - m})$
}

\tcp{Check curvature condition}
\uIf{$s_i^\top y_i \geq \epsilon$}{
    $S \leftarrow (s_i, s_{k - 1}, \cdots, s_{k - m - 1})$\;
    $Y \leftarrow (y_i, y_{k - 1}, \cdots, y_{k - m - 1})$\;
}
\Return{$S, Y$}
\caption{Inverse Hessian Update (Alg.~\ref{alg:lbfgs} line~\ref{alg:lbfgs_update_hessian})}
\label{alg:lbfgs_hessian_update_expanded}
\end{algorithm}

\begin{algorithm}[htb!]
\footnotesize
\newcommand\mycommfont[1]{\footnotesize\ttfamily\textcolor{cBLUE}{#1}}
\SetCommentSty{mycommfont}
\DontPrintSemicolon
\KwInput
{
current subgradient $g$, \\
Diff.\ of Lagrange variables 
$S = (s_{k}, \cdots, s_{k - m + 1})$, \\
Diff.\ of subgradients 
$Y = (y_{k}, \cdots, y_{k - m + 1})$, \\
}
\For{$j=k,k-1,\ldots,k-m+1$}
{
    $ \alpha_j = \rho_j s_j^\top g$\;
    $ g = g - \alpha_j y_j$\;
}
$d = (s_k^\top y_k / y_k^\top y_k) g$\;
\For{$j=k-m+1,k-m+2,\ldots,k$}
{
    $ \beta = \rho_j y_j^\top r$\;
    $ d = d + s_j(\alpha_j - \beta)$\;
}
\Return{$d$}
\caption{L-BFGS Direction (Alg.~\ref{alg:lbfgs} line~\ref{alg:lbfgs_update_direction})}
\label{alg:lbfgs_direction_expanded}
\end{algorithm}
\begin{algorithm}[htb!]
\footnotesize
\newcommand\mycommfont[1]{\footnotesize\ttfamily\textcolor{cBLUE}{#1}}
\SetCommentSty{mycommfont}
\SetKw{Break}{break}
\DontPrintSemicolon
\KwInput
{
    Lagrange variables $\lambda$,
    Dual feasible ascent direction $d$,
    Previous step size $\gamma$
}
$\gamma_m = \gamma$\;
$E_{init} = E(\lambda + \gamma d)$\;
\For{$t=1,\ldots,K$}
{
    \uIf{$E(\lambda + \gamma d) \leq E_{init}$}
    {
        $\gamma = \underline{\alpha}\gamma$ \tcp{Decrease step size}
    }
    \uElse
    {
        $\gamma = \overline{\alpha}\gamma$ \tcp{Increase step size}
    }
    \uIf{$E(\lambda + \gamma d) \geq E(\lambda + \gamma_m d)$}
    {
        $\gamma_m = \gamma$\;
    }
    \uIf{$E(\lambda + \gamma d) - E_{init} \geq \Delta_{min}$}
    {
        \Break\; 
    }
}
\Return{$\gamma_m$}
\caption{FindStepSize (Alg.~\ref{alg:lbfgs} line~\ref{alg:find_step_size})}
\label{alg:lbfgs_find_step_size_expanded}
\end{algorithm}

Lastly we observe that the Algorithms~\ref{alg:lbfgs_direction_expanded} and~\ref{alg:lbfgs_find_step_size_expanded} contain trivially parallelisable operations. We perform these operations on GPU and thereby maintain the massively parallel nature of the base solver~\cite{abbas2022fastdog}.

\subsection{Coarse-to-fine solving strategy}
Although we use the primal heuristic of~\cite{abbas2022fastdog} for solving our full shape matching problems, it does not produce feasible solutions in many cases of pruned ILPs. 
To circumvent this we aim to fix a large amount of confident variables to reduce problem sizes and solve the rest by Gurobi~\cite{gurobi}. 
To this end, given dual variables $\lambda$ after convergence of~\eqref{eq:dual-problem} we compute min-marginal differences~\cite{abbas2022fastdog} defined for all variables $i$ and subproblems $j$ as

\begin{equation}
    M_{i}^j = \left[\min_{s \in \mathcal{S}_j, s_i = 1} s^\top \lambda^{j}\right] - \left[\min_{s \in \mathcal{S}_j, s_i = 0} s^\top \lambda^{j}\right].
\end{equation}
Note that $M_{i}^j > 0$ implies the corresponding variable preferring an assignment of $0$ and viceversa. Moreover if all subproblems $j$ have min-marginal differences with the same sign the corresponding variable \textit{agrees} on that  assignment. For our purposes we compute $\sum_j M_{i}^j$ of all agreeing variables and fix top scoring $90\%$ of them to their preferable values. 
The resulting smaller problem is then solved by Gurobi~\cite{gurobi}.
Although such strategy can still fail, empirically it provides a solution in most cases.

\end{document}